\documentclass[12pt]{article}
\usepackage{amsmath,amsthm,amssymb,bm,mathrsfs}
\usepackage{graphicx}
\usepackage{xcolor}
\usepackage[ruled,linesnumbered]{algorithm2e}

\usepackage{caption,enumerate,listings}
\usepackage{setspace}
\usepackage[OT1]{fontenc}
\usepackage[colorlinks,linkcolor=red,anchorcolor=black,citecolor=black]{hyperref}
\usepackage{pifont}
\usepackage[margin=1in]{geometry}
\usepackage[protrusion=false,expansion=true]{microtype}
\usepackage[title]{appendix}
\usepackage{bbm}
\usepackage[authoryear,round]{natbib}
\makeatletter
\renewcommand\section{\@startsection{section}{1}{\z@}%
  {-1.5ex \@plus -0.5ex \@minus -.2ex}
  {0.3ex \@plus 0.2ex \@minus .2ex}
  {\normalfont\Large\bfseries}}
\makeatother

\usepackage{subcaption}
\usepackage{comment}
\usepackage{enumitem}
\usepackage{multirow}
\def\spacingset#1{\renewcommand{\baselinestretch}%
{#1}\small\normalsize} \spacingset{1}
\newcommand{\argmin}{\mathop{\mathrm{argmin}}}

\newtheorem{lemma}{{\bf Lemma}}
\newtheorem{corollary}{{\bf Corollary}}
\newtheorem{theorem}{{\bf Theorem}}
\newtheorem{assumption}{{\bf Assumption}}

\newtheorem{remark}{{\bf Remark}}
\newtheorem{example}{{\bf Example}}
\usepackage{setspace}
\usepackage{booktabs}
\theoremstyle{plain}
\usepackage{algorithm}
\usepackage{algorithmic}
\usepackage{titlesec}
\titleformat{\section}[block]{\normalfont\large\bfseries}{\thesection}{1em}{}

\floatname{algorithm}{Fresh Data Augmentation Case}

\title{\Large \bf A Probabilistic Perspective on Model Collapse}
\date{ }
\author{}

\begin{document}

\date{ }
 \author{
 Shirong Xu$^\dag$ \quad \footnote{The first two authors contribute equally to this work}
 Hengzhi He$^\dag$ \quad 
 Guang Cheng$^\dag$ \\
  $^\dag$Department of Statistics and Data Science\\
      University of California, Los Angeles
 }
\maketitle

\begin{abstract}
In recent years, model collapse has become a critical issue in language model training, making it essential to understand the underlying mechanisms driving this phenomenon. In this paper, we investigate recursive parametric model training from a probabilistic perspective, aiming to characterize the conditions under which model collapse occurs and, crucially, how it can be mitigated. We conceptualize the recursive training process as a random walk of the model estimate, highlighting how the sample size influences the step size and how the estimation procedure determines the direction and potential bias of the random walk. Under mild conditions, we rigorously show that progressively increasing the sample size at each training step is necessary to prevent model collapse. In particular, when the estimation is unbiased, the required growth rate follows a superlinear pattern. This rate needs to be accelerated even further in the presence of substantial estimation bias. Building on this probabilistic framework, we also investigate the probability that recursive training on synthetic data yields models that outperform those trained solely on real data. Moreover, we extend these results to general parametric model family in an asymptotic regime. Finally, we validate our theoretical results through extensive simulations and a real-world dataset.

\end{abstract}

\spacingset{1.7} 
\section{Introduction}

In recent years, the use of synthetic data to train large-scale models has become increasingly widespread, driven by the rising demand for data as generative models continue to scale up. As large language models \citep{achiam2023gpt,touvron2023llama,team2023gemini} grow, scaling laws call for ever-larger training datasets to fully realize their capacity. Yet the supply of high-quality, human-generated data has not kept pace \citep{villalobos2024position}, prompting researchers and practitioners to turn to synthetic data to bridge the gap. Meanwhile, synthetic data generated by these models is increasingly disseminated online, where it becomes indistinguishable from human-generated material and is often inevitably incorporated into future training datasets. Recent studies have highlighted the risks of this growing reliance on synthetic data, including performance degradation and divergence from the real data distribution \citep{xu2023utility,shumailov2024ai}.

\begin{figure}[h]
    \centering
    \includegraphics[scale=0.3]{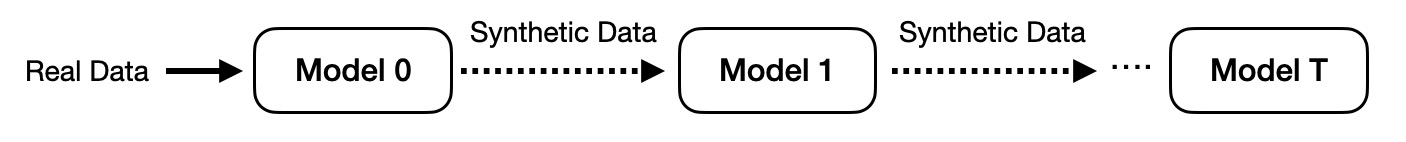}
    \caption{Model Collapse in Recursive Training Framework \citep{shumailov2024ai}.}
    \label{Fig:RST}
\end{figure}

The growing trend of using synthetic data has raised a broader concern known as model collapse, which refers to the gradual degradation in model performance when generative models are iteratively trained on their own synthetic outputs, as illustrated in Figure \ref{Fig:RST}.

\citet{shumailov2024ai} show that when a Gaussian model is iteratively trained on datasets of {\em equal sample size} generated by the previous Gaussian, its estimated mean can diverge arbitrarily from the true distribution, while its covariance matrix degenerates toward zero. This outcome signals severe model collapse, marked by vanishing variability and the progressive loss of meaningful representation of the original data distribution. To our knowledge, model collapse is primarily caused by the progressive loss of information about the real data distribution as the recursive training process advances (i.e., as $T$ increases in Figure~\ref{Fig:RST}). Particularly, if each training step utilizes the same size of training data, a consistent and irreversible loss of information accumulates over successive generations, leading to degraded model performance.

To mitigate model collapse, existing research has focused on developing strategies such as continuously accumulating historical data \citep{gerstgrasser2024is,kazdan2024collapse,dey2024universality} or regularly incorporating real data to counterbalance synthetic data in each iteration \citep{alemohammad2024selfconsuming,he2025golden}. For instance, \citet{gerstgrasser2024is} demonstrate that accumulating historical data can effectively control test error in linear regression, a result that \citet{dey2024universality} further generalize to exponential family models. Moreover, \citet{gerstgrasser2024is} show that regularly incorporating new real data during recursive training effectively prevents model collapse, while \citet{he2025golden} provide a rigorous theoretical analysis of how to optimally balance real and synthetic data to train better models. The essence of existing methods lies in preserving as much information about the true data distribution as possible by feeding more real data to counteract information loss. This naturally raises a question: 
$$
\textit{How much data do we really need for training to prevent model collapse?}
$$

This question is critical because increasing the size of synthetic data leads to higher computational costs. Understanding the boundary of the size of training data that prevents model collapse is key to achieving a balance between computational efficiency and model utility. To address this question, \citet{shumailov2023curse} study this phenomenon using a Gaussian toy model, demonstrating that adopting a superlinear sample-size growth schedule can effectively prevent unbounded risk of parameter estimation. However, whether this result extends to a broader class of recursive training problems remains an open question.

To address this question, we explore the impact of sample size schedule on model collapse during recursive training. In a novel approach, we conceptualize recursive training---where model collapse occurs---as a type of random walk of the model parameters. Specifically, the estimated parameter, derived from samples drawn from a parametric model, can be viewed as a drift from the true value to its estimate. Within this framework, several factors influence the trajectory of the random walk, including the sample size of the training dataset (which determines the step size) and the estimation procedure (which determines the direction). 

Under mild conditions, we derive several key insights. First, if the estimation procedure is unbiased or has negligible bias, the sample-size growth schedule required to prevent model collapse follows a superlinear pattern, i.e., \( O(t^{1+s}) \) for some \( s > 0 \), where \( t \) denotes the \( t \)-th training step. This generalizes the result of \citet{shumailov2023curse} as a special case. However, a more interesting scenario is when the estimation bias is substantial. In this case, an even faster sample-size growth schedule is required to prevent model collapse. In other words, biased estimation accelerates model collapse during recursive training. 

Second, we conceptualize recursive training as a type of random walk to address a critical question: \textit{Can recursive training with synthetic data lead to an improved model compared with the initial one trained purely on real data?} To explore this, we decompose the estimation error after multiple recursive steps into a sum of incremental updates, allowing us to analyze the cumulative effect of training. In the Gaussian setting—where the recursion corresponds to the Gaussian random walk—we derive a closed-form expression for the probability that recursive training leads to improvement within a finite number of steps. We then extend this analysis to general parametric models in the asymptotic regime, under the assumption that recursive estimators based on synthetic data are asymptotically normal. Our results offer a principled framework for quantifying the probability of model improvement through recursive training with synthetic data in a broad class of parametric estimation problems.

Overall, the contributions of this paper are summarized as follows:
\begin{itemize}
\item We rigorously demonstrate that, under mild conditions, progressively increasing the sample size at each generation effectively mitigates model collapse across a broad class of parametric estimation problems. Specifically, we characterize the threshold on the synthetic sample size required to prevent model collapse. Furthermore, we show that large estimation bias can significantly accelerate the onset of model collapse.

\item We systematically characterize how the required sample size growth rate depends on the properties of different estimators, clearly distinguishing between the regimes applicable to unbiased and biased estimators.

\item We provide insights into the probability that estimators trained solely on synthetic data from the previous generation can outperform those trained exclusively on real data in terms of parameter estimation. Notably, this result extends naturally to any estimator that satisfies asymptotic normality. Moreover, we show that synthetic data expansion can further increase this probability. This finding addresses a deeper question in the use of synthetic data: \textit{If synthetic data can be beneficial, what is the probability that it actually leads to a better model?}
\end{itemize}

The remainder of the paper is structured as follows. After introducing some necessary notation, Section~\ref{Sec:PS} provides the background on recursive training and presents the model collapse phenomenon from a probabilistic perspective. Section~\ref{Sec:RW} frames the recursive training process as a type of random walk, highlighting the role of training sample size in influencing model collapse throughout the process. Section~\ref{Sec:DE} analyzes how the proposed synthetic data expansion schedule mitigates model collapse, shedding light on the limits of using synthetic data depending on the properties of the estimators employed by generative models. Section~\ref{Sec:MLE_Study_Case} investigates whether recursive training can yield a better model than one trained on real data. Section~\ref{Sec:Exp} validates our theoretical findings through simulations and a real-world dataset. All proofs of examples and theoretical results are provided in the Appendix.

\textbf{Notation.} In this paper, we adopt bold notation to represent multivariate quantities and unbolded notation for univariate quantities. For example, $\bm{\theta}^\star$ denotes the multivariate ground-truth parameter, whereas $\theta^\star$ denotes the univariate counterpart. For any positive integer \( M \), we define the set \( [M] = \{1, 2, \ldots, M\} \). For instance, \( \bm{x} \in \mathbb{R}^p \) denotes a \( p \)-dimensional vector, while \( x \) represents a scalar. For a vector \( \bm{x} \in \mathbb{R}^p \), its \( l_2 \)-norm is given by \( \Vert \bm{x} \Vert_2 = \left( \sum_{i=1}^p |x_i|^2 \right)^{1/2} \). For a multivariate continuous random variable \( \bm{X} \), \( p_{\bm{\theta}}(\bm{x}) \) denotes its probability density function at \( \bm{x} \), and \( \mathbb{P}_{\bm{\theta}} \) refers to the associated probability measure, where \( \bm{\theta} \) is the underlying parameter. The expectation with respect to the randomness of \( \bm{X} \) is denoted by \( \mathbb{E}_{\bm{X}} \). For two sets $A = \{a_1, \ldots, a_n\}$ and $B = \{b_1, \ldots, b_m\}$, we define their Cartesian product as $A \times B = \{(a_i, b_j) : a_i \in A,\, b_j \in B\}$.

\section{Preliminaries}
\label{Sec:PS}
In this section, we first provide an overview of recursive training using synthetic data generated from the previous iteration of the generative model, as well as the model collapse phenomenon, which is discussed in Section \ref{subsec:MCS}. Subsequently, in Section \ref{subsec:PMC}, we offer a new probabilistic perspective on the model collapse phenomenon.

\subsection{Model Collapse Setup}
\label{subsec:MCS}
Consider a family of generative models parameterized by $ \bm{\theta} \in \bm{\Theta} $, denoted as $ \mathcal{P} = \{\mathbb{P}_{\bm{\theta}} : \bm{\theta} \in \bm{\Theta}\} $, where $ \mathbb{P}_{\bm{\theta}}(\bm{x}) $ defines a probability distribution supported on $ \mathcal{X} $. Let $ \bm{\theta}^\star $ be the ground truth parameter. The objective is to analyze the efficiency of $ \widehat{\bm{\theta}}_T $ in estimating $ \bm{\theta}^\star $ over successive training steps, particularly as $ T $ approaches infinity.

\begin{figure}[h]
    \centering
    \includegraphics[width=0.95\linewidth]{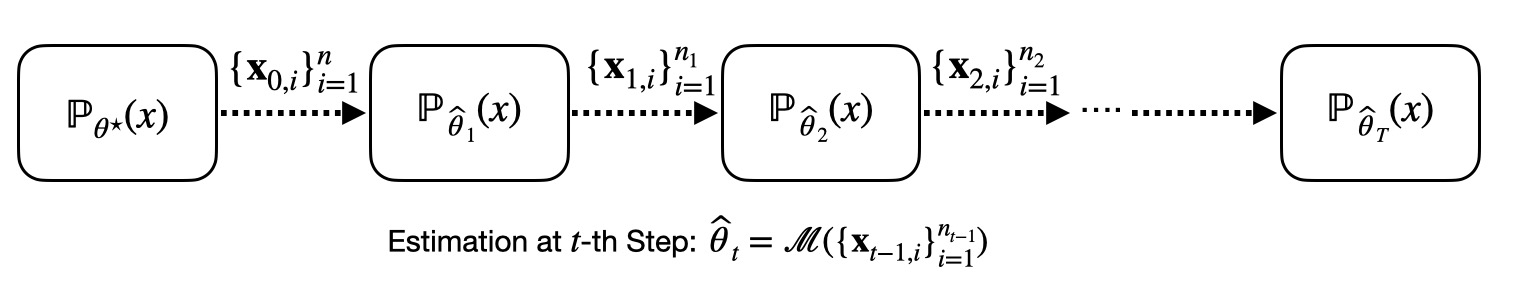}
    \caption{A General Framework for Recursive Training with Fully Synthetic Data}
    \label{fig:FS_framework}
\end{figure}

As illustrated in Figure \ref{fig:FS_framework}, a real dataset $\mathcal{D}_0 = \{\bm{x}_{0,i}\}_{i=1}^n$, drawn from $\mathbb{P}_{\bm{\theta}^\star}$, is available for training the next generative model $\mathbb{P}_{\widehat{\bm{\theta}}_1}$. Specifically, at the $t$-th training step, $\mathbb{P}_{\widehat{\bm{\theta}}_t}$ is trained on the dataset $\mathcal{D}_{t-1} = \{\bm{x}_{t-1,i}\}_{i=1}^{n_{t-1}}$ generated from last generative model. Throughout the recursive training process depicted in Figure \ref{fig:FS_framework}, an estimation scheme $\mathcal{M}: \mathcal{X}^{\mathbb{N}} \to \bm{\Theta}$ is consistently applied at each estimation step, ensuring that $\widehat{\bm{\theta}}_{t+1} = \mathcal{M}(\mathcal{D}_{t})$ for all $t \geq 0$. The framework depicted in Figure \ref{fig:FS_framework} is referred to as fully synthetic training in the existing literature \citep{shumailov2024ai}. In this paper, we refer to $\widehat{\bm{\theta}}_1$ as the real-data estimate, and to $\widehat{\bm{\theta}}_t$ for $t \geq 2$ as synthetic-data estimates.

Existing literature demonstrates that when \( n_t = n \) for any \( t \geq 1 \), the recursive estimation causes \( \widehat{\bm{\theta}}_T \) to fail, as the estimation error of \( \widehat{\bm{\theta}}_T \) in estimating \( \bm{\theta}^\star \) diverges \citep{he2025golden, shumailov2024ai}. From an information-theoretic viewpoint, the information contained in $\mathcal{D}_0$ about the real distribution \( \mathbb{P}_{\bm{\theta}^\star} \) gradually diminishes as the repeated training-and-generation process progresses. To illustrate this point, we use the following example of Gaussian mean estimation (Example \ref{Exam:prop_recursive_variance}). In this example, the goal is to estimate the true mean of a standard normal distribution. Clearly, as the recursive training progresses, the estimation error of the final estimator will diverge to infinity if $n_t = n$ for $t\geq 1$, as confirmed in the existing literature \citep{kazdan2024collapse,he2025golden}. Therefore, this phenomenon reveals that the main culprit behind model collapse is the new randomness introduced by the recursive training, which causes the estimation error to increase over time. 

In the following, we present the example of recursive Gaussian estimation to shed light on how the sample schedule $\{n_t\}_{t=1}^{\infty}$ influences the estimator after multi-step recursive training.

\begin{example}
    \label{Exam:prop_recursive_variance}
Consider a recursive Gaussian mean estimation process defined by $\widehat{\theta}_{t+1} = \mathcal{M}(\mathcal{D}_{t})=\frac{1}{n_{t}}\sum_{i=1}^{n_t}x_{t,i}$ with $\mathcal{D}_{t}=\{ x_{t,i}\}_{i=1}^{n_{t}} \sim N(\widehat{\theta}_{t},1)$ for $t \geq 1$ and $\mathcal{D}_{0} = \{x_{0,i}\}_{i=1}^n \sim N(\theta^\star,1)$. Then, for all \( t \geq 1 \), we have $\mathbb{E}[(\widehat{\theta}_{t+1} -\theta)^2]=
    \mathrm{Var}(\widehat{\theta}_{t+1}) = \frac{1}{n_{t}}+\mathrm{Var}(\widehat{\theta}_{t})$. This then implies that $\mathrm{Var}(\widehat{\theta}_{T})=n^{-1}+\sum_{t=1}^{T-1}n_t^{-1}$.
\end{example}

Example \ref{Exam:prop_recursive_variance} illustrates that the uncertainty of \(\widehat{\theta}_{T}\) increases as \(T\) grows. However, if \(\{n_t\}_{t=1}^{T-1}\) are sufficiently large, the degradation in estimation accuracy can be mitigated. This is because generating a sufficient amount of synthetic data for the next round of generative model training helps preserve more information from the previous generative model, which slows down the speed of model collapse. This result highlights that a key factor affecting the model collapse phenomenon is the sample size schedule \(\{n_t\}_{t=1}^{\infty}\). Without loss of generality, we assume that the sample size schedule follows the form:  
$$
\text{Sample Size Schedule: } n_{t} = c_t \times n \text{ for } t \geq 1.
$$
\begin{remark}
The estimation error does not necessarily diverge to infinity in the fully synthetic framework illustrated in Figure \ref{fig:FS_framework}. Consider the case where \( T = \infty \) under the sample size schedule \(\{c_t\}_{t=1}^{\infty}\). Then, the estimation error is given by  
\begin{align*}
 \mathbb{E}[(\widehat{\theta}_{\infty} -\theta^\star)^2] \triangleq
\lim_{T\rightarrow \infty}
    \mathbb{E}[(\widehat{\theta}_{T}-\theta^\star)^2] = \frac{1}{n} + \frac{1}{n} \sum_{t=1}^{\infty} \frac{1}{c_t}.
\end{align*}  
If \(\{c_t\}_{t=1}^{\infty}\) is chosen such that \(\sum_{t=1}^{\infty}c_t^{-1} < \infty\), then we still have $\lim\limits_{n\rightarrow \infty} \lim\limits_{T\rightarrow \infty}\mathbb{E}[(\widehat{\theta}_{T}-\theta^\star)^2] = 0$.
\end{remark}

From the above analysis, a simple trick to prevent exploding population risk is through an appropriate synthetic data expansion scheme. For example, if \( c_t = 2^{t} \), the estimation error of Gaussian mean estimation (Example \ref{Exam:prop_recursive_variance}) is given by \(\mathbb{E}[(\widehat{\theta}_{\infty} -\theta^\star)^2] = 2/n\). Clearly, statistical consistency is still maintained, i.e., \(\lim_{n\rightarrow \infty} \mathbb{E}[(\widehat{\theta}_{\infty} -\theta^\star)^2] = 0\). The underlying reason is that adaptively expanding the sample size mitigates the information loss caused by recursive training. Notably, if \( c_t = \infty \), the estimation error reduces to \( 1/n \), which corresponds to the case of using only real data. Nevertheless, from a practical perspective, using too much data for training inevitably increases computational cost. Therefore, preventing model collapse may require balancing the tradeoff between computational cost and statistical consistency if model collapse is defined as diverging estimation error \citep{kazdan2024collapse,he2025golden}.

\subsection{Probabilistic Model Collapse}
\label{subsec:PMC}
In this section, we provide an alternative probabilistic perspective to gain deeper insights into the behavior of synthetic-data estimates. This perspective lays the foundation for understanding model collapse through a random walk framework.

As noted in \citet{schaeffer2025position}, the existing literature presents diverse definitions of model collapse. Consequently, relying solely on the population risk of estimators during the recursive training process may not always capture model collapse effectively. This is because model collapse occurs in the context of large language model (LLM) training, which may involve only a limited number of recursive training processes for generative models. In contrast, population risk evaluates statistical consistency, inherently reflecting the averaged behavior of generative model estimation over an infinite number of recursive training processes. To support this claim, we introduce the results of recursive Gaussian estimation in Theorem \ref{Exam:Probvari}. In this example, we assume that the mean $\mu$ remains known throughout the process, and the primary objective is to estimate the variance of a normal distribution using a synthetic dataset generated from the preceding normal distribution. The goal of this case is to investigate how the diversity of synthetic data evolves throughout the recursive training process.

\begin{theorem}
\label{Exam:Probvari}
Consider the following recursive estimation process for the variance of a Gaussian distribution with a known mean \(\mu\). At the \(t\)-th step, a dataset \(\mathcal{D}_{t} = \{x_{t-1,i}\}_{i=1}^n\) of fixed size \(n\) is sampled from \(N(\mu, \widehat{\sigma}_{t-1}^2)\). The variance is updated as $\widehat{\sigma}_{t}^2 = \frac{1}{n} \sum_{i=1}^n (x_{t-1,i} - \mu)^2$.
\begin{align*}
    N(\mu,\sigma^2) \xrightarrow{\mathcal{D}_0} N(\mu,\widehat{\sigma}_1^2)
     \xrightarrow{\mathcal{D}_1}  \cdots \xrightarrow{\mathcal{D}_{T-1}} N(\mu,\widehat{\sigma}_T^2)
\end{align*}
For $n \geq 2$ and $\epsilon>0$, it holds that
    \begin{align*}
    \textnormal{\textbf{Diverging   Population  Risk:} }&
        \lim_{T\rightarrow \infty}\mathbb{E}[(\widehat{\sigma}_T^2-\sigma^2)^2]=\lim_{T\rightarrow \infty}\left[\left(1+\frac{2}{n}\right)^{T}-1\right]\sigma^4=\infty, \\
    \textnormal{\textbf{Vanishing Diversity:} } & \lim_{T\rightarrow \infty} \mathbb{P}(\widehat{\sigma}_T^2 \leq \epsilon) = 1.
    \end{align*}
\end{theorem}

Theorem \ref{Exam:Probvari} demonstrates that the population risk 
\( \mathbb{E}\left[(\widehat{\sigma}_T^2 - \sigma^2)^2\right] \) 
diverges to infinity as \( T \) increases to infinity. This result indicates that the estimator \( \widehat{\sigma}_T^2 \) fails to consistently estimate \( \sigma^2 \) as \( T \to \infty \), regardless of the sample size \( n \), highlighting the phenomenon of diverging population risk. Interestingly, as \( T \) approaches infinity, \( \widehat{\sigma}_T^2 \) tends to be close to zero with probability approaching one. This suggests that, in a recursive training process, the final normal distribution is highly likely to degenerate into a singular Gaussian distribution. When viewed as a generative model, this final Gaussian distribution would only generate homogeneous synthetic data. This phenomenon is referred to as the loss of diversity in the existing literature about model collapse \citep{bertrand2024on, alemohammad2024selfconsuming}.

It is worth noting that the vanishing diversity and the diverging population risk in Theorem \ref{Exam:Gaus} may appear contradictory. The former suggests that \( \widehat{\sigma}_T^2 \) is highly likely close to zero, while the latter asserts that the expected squared difference between \( \widehat{\sigma}_T^2 \) and \( \sigma^2 \) diverges to infinity as \( T \) approaches infinity. However, these two phenomena can indeed co-exist. The key reason is that while most recursive training processes yield an estimate \( \widehat{\sigma}_T^2 \approx 0 \), there remain instances where \( \widehat{\sigma}_T^2 \) diverges, ultimately leading to an unbounded risk. To support above claim, we conduct the following experiment on Gaussian estimation in Theorem \ref{Exam:Probvari}.


\begin{figure}[h!]
    \centering
    \begin{subfigure}[b]{0.323\textwidth}
        \centering
        \includegraphics[width=\textwidth]{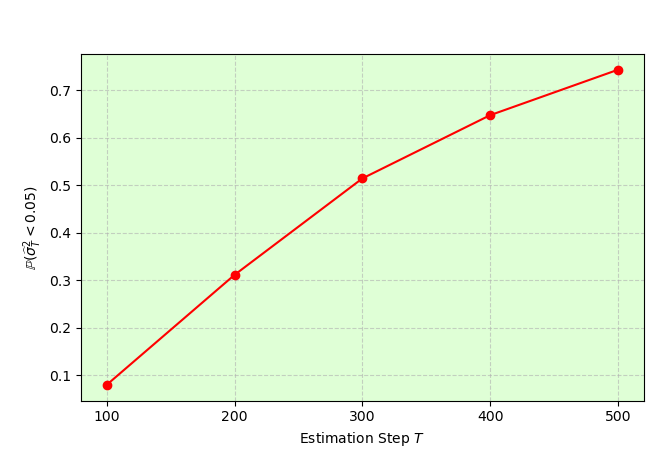}
        \caption{$\mathbb{P}(\widehat{\sigma}_T^2<0.05)$}
        \label{fig:sub1}
    \end{subfigure}
    \hfill
    \begin{subfigure}[b]{0.323\textwidth}
        \centering
        \includegraphics[width=\textwidth]{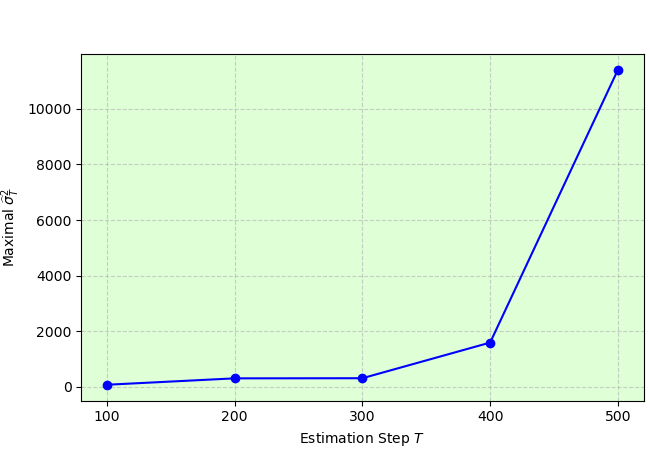}
        \caption{Maximal $\widehat{\sigma}_T^2$}
        \label{fig:sub2}
    \end{subfigure}
    \hfill
    \begin{subfigure}[b]{0.323\textwidth}
        \centering
        \includegraphics[width=\textwidth]{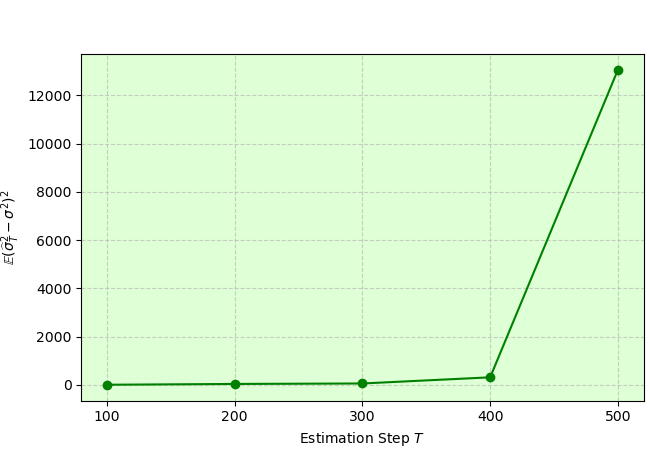}
        \caption{$\mathbb{E}(\widehat{\sigma}_T^2-\sigma^2)^2$}
        \label{fig:sub3}
    \end{subfigure}
    \caption{\textbf{Experimental Setup for Recursive Gaussian Estimation:} We fix parameters \( (n, \mu, \sigma^2) = (100, 0, 1) \) and vary \( T \in \{100, 200, 300, 400, 500\} \). For each \( T \), we conduct \( 10^4 \) replications, recording the estimate \( \widehat{\sigma}_{T,i}^2 \) for each replication. We then report the percentage of replications with \( \widehat{\sigma}_{T,i}^2 \leq 0.05 \), the maximum value \( \max_{i} \widehat{\sigma}_{T,i}^2 \), and the estimated population risk across all replications.}
    \label{fig:Ex1}
\end{figure}

As shown in Figure \ref{fig:sub1}, as \( T \) increases, the proportion of \( \widehat{\sigma}_T^2 \) values falling below 0.05 rises from approximately 10\% at \( T = 100 \) to over 70\% at \( T = 500 \) among 10,000 replications. This result suggests that, with high probability, the generative model loses diversity after recursive training. In contrast, Figure \ref{fig:sub2} shows that the maximum \( \widehat{\sigma}_T^2 \) also diverges, exceeding 10,000 when \( T = 500 \). This indicates that, while rare, there exist events where the final generative model exhibits greater diversity than the real distribution ($\widehat{\sigma}_T^2 \gg 
 \sigma^2$). However, the probability of such occurrences vanishes as \( T \) approaches infinity. Furthermore, as shown in Figure \ref{fig:sub3}, the population risk \( \mathbb{E}[(\widehat{\sigma}_T^2 - \sigma^2)^2] \) exceeds 12,000 when \( T = 500 \), highlighting the statistical inconsistency of \( \widehat{\sigma}_T^2 \).

The experimental result in Figure \ref{fig:Ex1} raises a key question: \textit{How to understand or define model collapse}? As highlighted in the existing literature \citep{schaeffer2025position}, the definitions of model collapse do not reach concensus at this moment. In this paper, we argue that model collapse should be understood from a probabilistic perspective. For example, the experimental result demonstrates that the diversity of the final generative model will vanish with probability one. This form of model collapse provides more meaningful insights for the training of practical generative models, especially large language models \citep{shumailov2024ai}. The reason for this is that, in large language model training, researchers typically focus on the outcome of a single recursive training process, rather than the averaged result of many training processes, which is computationally unfeasible in practice. In this context, understanding model collapse through population risk provides limited guidance for recursive training. Therefore, in this paper, we define \textbf{model collapse} in parametric models as follows:
\[
\text{Model Collapse: } \lim_{n \rightarrow \infty} \lim_{T \rightarrow \infty} \mathbb{P} \left( \|\widehat{\bm{\theta}}_{T} - \bm{\theta}^\star\|_2 \geq \delta \right) = 1,  \text{ for some } \delta > 0.
\]
This definition captures the phenomenon where, regardless of the initial sample size \( n \), the synthetic-data estimate \( \widehat{\bm{\theta}}_T \) inevitably drifts away from the true parameter \( \bm{\theta}^\star \) as the number of recursive training steps \( T \) grows.

\section{Recursive Training: A Random Walk Perspective}
\label{Sec:RW}
To understand the probabilistic nature of recursive training, which provides valuable insights into the model collapse phenomenon, we aim to formalize the recursive training process as a form of a random walk.

\begin{figure}[h]
    \centering
    \includegraphics[scale=0.2]{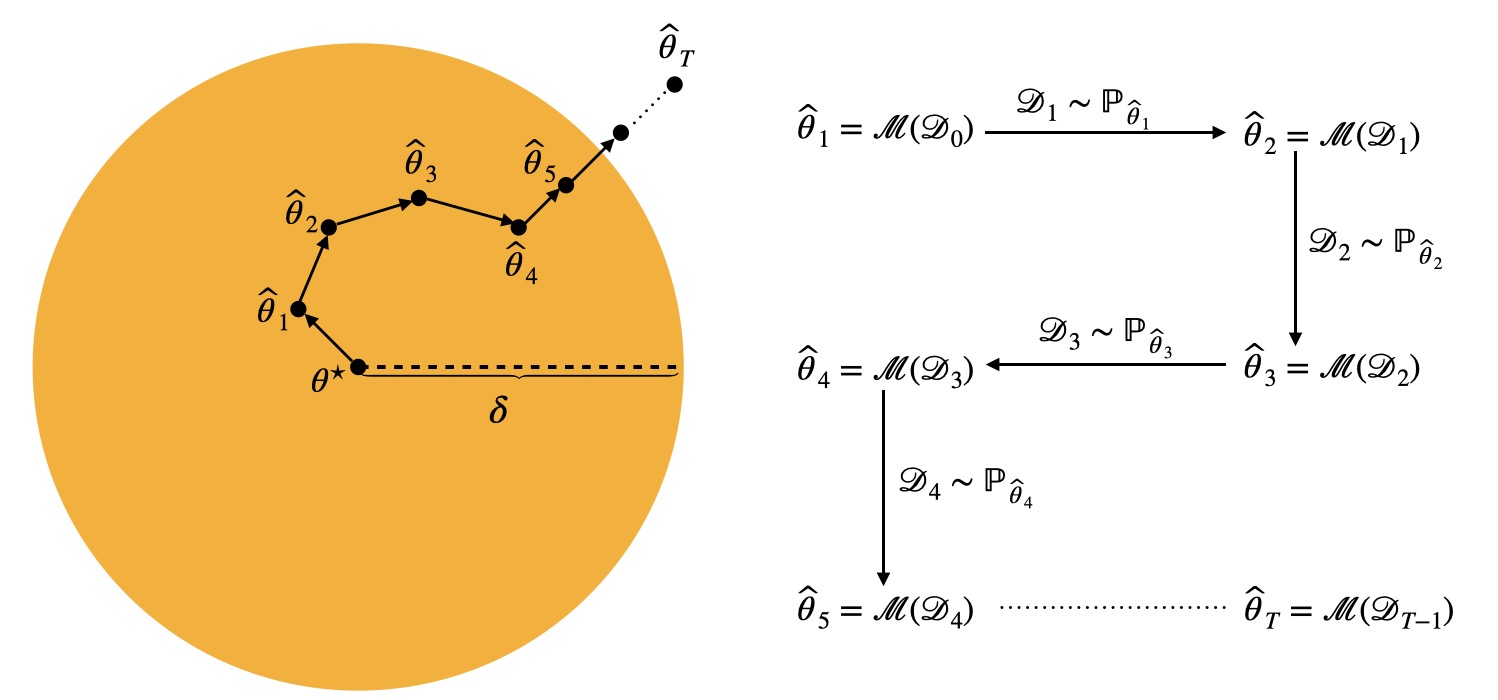}
    \caption{An illustration of recursive training represented as a random walk.}
    \label{fig:BallCollapse}
\end{figure}

Within the framework of Figure \ref{fig:FS_framework}, the recursive training process of parametric generative models can be equivalently represented in Figure \ref{fig:BallCollapse}. At the \((t+1)\)-th training step, the parameter \(\widehat{\bm{\theta}}_{t+1}\) is estimated based on the dataset \(\mathcal{D}_{t} = \{\bm{x}_{t,i}\}_{i=1}^{n_t} \overset{\text{i.i.d.}}{\sim} \mathbb{P}_{\widehat{\bm{\theta}}_t}\). Consequently, \(\widehat{\bm{\theta}}_{t+1}\) can be viewed as following a random walk originating from \(\widehat{\bm{\theta}}_t\), where the randomness arises from \(\mathcal{D}_t\). Thus, the recursive training process can be interpreted as the estimated parameter undergoing a stochastic trajectory in the parameter space.
At the \( t \)-th step, the estimated parameter moves from \( \widehat{\bm{\theta}}_{t-1} \) to \( \widehat{\bm{\theta}}_{t} \), where the direction and step size at the \( t \)-th step primarily depend on the generative model \( \mathbb{P}_{\widehat{\bm{\theta}}_{t-1}} \) and the size of \( \mathcal{D}_{t-1} \), respectively. For example, if \( \mathcal{D}_{t-1} = \{\bm{x}_{t-1,i}\}_{i=1}^{n_{t-1}} \) is generated from \( N(\widehat{\bm{\theta}}_{t-1},\bm{I}) \) and \( \widehat{\bm{\theta}}_t = \frac{1}{n_{t-1}} \sum_{i=1}^{n_{t-1}} \bm{x}_{t-1,i} \), then the direction of \( \widehat{\bm{\theta}}_{t} - \widehat{\bm{\theta}}_{t-1} \) is completely random due to the covariance structure, and the step size \( \Vert \widehat{\bm{\theta}}_{t} - \widehat{\bm{\theta}}_{t-1} \Vert_2 \) depends on \( n_{t-1} \). Particularly, if \( n_{t-1}=\infty \), then we have \( \widehat{\bm{\theta}}_{t} = \widehat{\bm{\theta}}_{t-1} \), indicating that the step size is zero. Within this framework, two questions for understanding recursive training naturally arise:  
\begin{enumerate}  
    \item[\textbf{Q1}] Considering \( B(\delta) = \{ \bm{\theta} : \|\bm{\theta} - \bm{\theta}^\star\|_2 < \delta \} \) as the desirable region for the parameters of generative models, what is the probability that \( \widehat{\bm{\theta}}_T \) remains within \( B(\delta) \) when $T\rightarrow \infty$ under a specific sample schedule $\{c_t\}_{t=1}^{\infty}$?  
    \item[\textbf{Q2}] Since \( \widehat{\bm{\theta}}_1 \) is trained on real data, while \( \widehat{\bm{\theta}}_t \) for \( t \geq 2 \) are trained on synthetic data, an important question is: Is it possible to obtain a generative model trained on synthetic data that surpasses \( \mathbb{P}_{\widehat{\bm{\theta}}_1} \)? Specifically, we are interested in understanding the probability that $\|\widehat{\bm{\theta}}_T - \bm{\theta}^\star\|_2 < \|\widehat{\bm{\theta}}_1 - \bm{\theta}^\star\|_2
$ under a specific sample schedule $\{c_t\}_{t=1}^{\infty}$.
\end{enumerate}

Theoretically, the recursive training process reduces the estimation efficiency of $\widehat{\bm{\theta}}_T$, causing its deviation from $\bm{\theta}^\star$ to increase as $T$ grows. As a result, the probability of either $\widehat{\bm{\theta}}_T$ remaining within a radius $\delta$ around $\bm{\theta}^\star$ or outperforming $\widehat{\bm{\theta}}_1$ becomes increasingly difficult as $T$ increases. Intuitively, for a fixed $n$, the answer to \textbf{Q1} depends on the estimation procedure $\mathcal{M}$, the dimension of the parameter \( \bm{\theta} \) (denoted by \( p \)), the number of recursive training iterations \( T \), and most importantly, the sample schedule \( c_t \). In general, when \( T \) is fixed and finite, the probability that \( \{\|\widehat{\bm{\theta}}_{T} - \bm{\theta}^\star\|_2 < \delta\} \) approaches one as \( n \rightarrow \infty \), reflecting standard estimation consistency. However, the central question is whether this statistical consistency persists even as \( T \rightarrow \infty \). Intuitively, by carefully choosing an appropriate sample size schedule \( c_t \), we can ensure that \( \widehat{\bm{\theta}}_T \) remains within a \( \delta \)-neighborhood of \( \bm{\theta}^\star \) with high probability even when $T\rightarrow \infty$. In other words, we aim to identify a suitable sampling scheme \( \{c_t\}_{t=1}^{\infty} \) such that,
\[
\text{Expanding sample schedule} \; (c_{t+1} \geq c_t): \quad
\lim_{n\rightarrow \infty}\lim_{T\rightarrow\infty} \mathbb{P} \left( \|\widehat{\bm{\theta}}_{T} - \bm{\theta}^\star\|_2 \geq \delta \right) = 0,
\text{ for any } \delta > 0
\]

To answer \textbf{Q2}, it is necessary to understand the probabilistic behavior of $\widehat{\bm{\theta}}_T$ and its connection to the estimation procedure $\mathcal{M}$, which determines the walking direction and step size. Additionally, the sample schedule $c_t$ governs the step size at each stage of the random walk, since with infinite samples used for training, the estimated parameters would exactly match those from the previous round.

\section{Synthetic Data Expansion Prevents Model Collapse}
\label{Sec:DE}
In this section, we aim to explore \textbf{Q1} specified in Section \ref{Sec:RW}, investigating the conditions under which $\widehat{\bm{\theta}}_T$ remains close to $\bm{\theta}^\star$ as $T \to \infty$. In other words, we seek a proper sample scheme $\{c_t\}_{t=1}^{\infty}$ to guarantee that
\begin{align}
\label{ModelCollapse}
\lim_{n\rightarrow \infty}\lim_{T\rightarrow\infty} \mathbb{P} \left( \|\widehat{\bm{\theta}}_{T} - \bm{\theta}^\star\|_2 \geq \delta \right) = 0, \text{ for any }\delta>0.
\end{align}
Additionally, using a faster expanding scheme $c_t$ will certainly ensure that (\ref{ModelCollapse}) holds true; however, it also results in higher computational cost. Therefore, understanding the dividing line of the sequence $\{c_t\}_{t=1}^{\infty}$ that determines whether (\ref{ModelCollapse}) holds or fails is important for balancing computational cost and model utility.

\subsection{Trivial Union Bound}
\begin{figure}[h]
    \centering
    \includegraphics[scale=0.2]{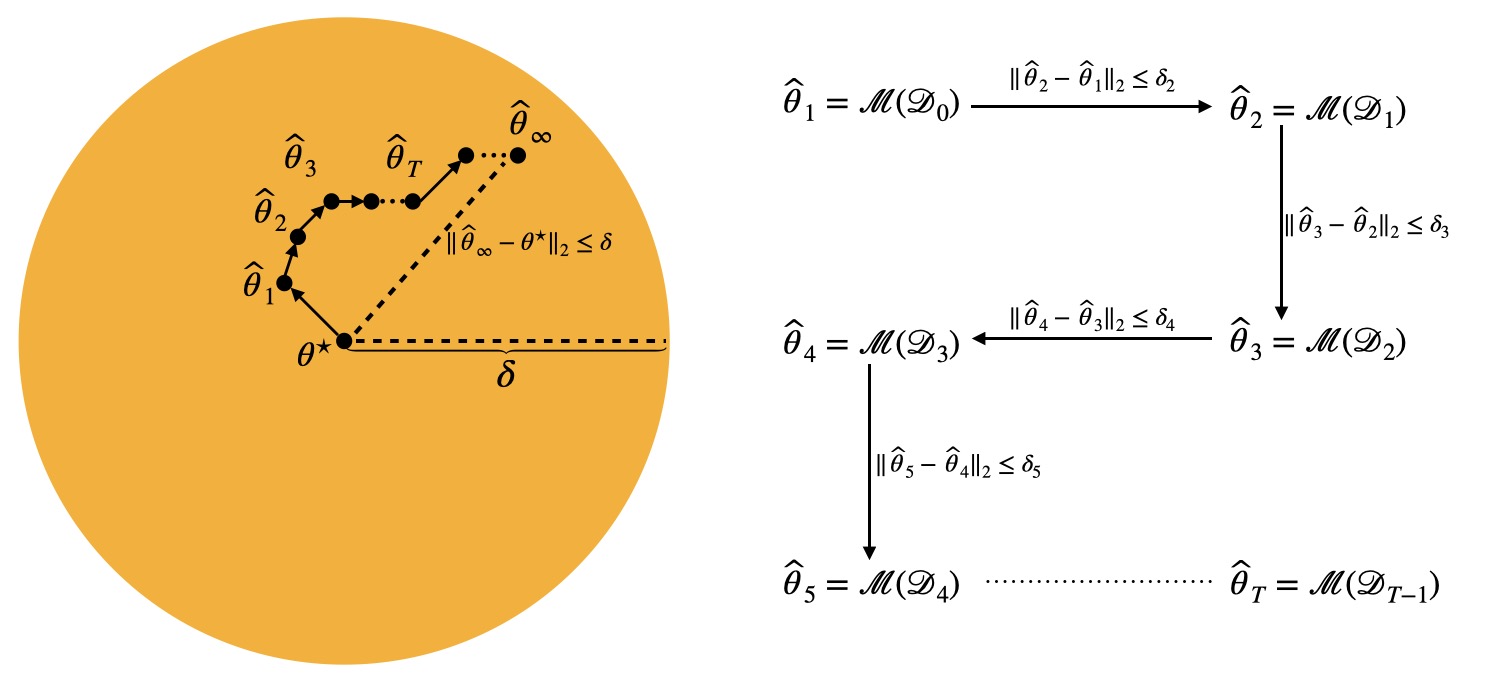}
    \caption{An illustration of recursive training represented as a random walk.}
    \label{fig:BallNoCollapse}
\end{figure}

As discussed in Section~\ref{Sec:RW}, the behavior of $\widehat{\bm{\theta}}_{\infty}$ can be interpreted as an infinite-step random walk, where $\Vert \widehat{\bm{\theta}}_{t} - \widehat{\bm{\theta}}_{t-1} \Vert_2$ represents the step size at the $t$-th iteration. If the cumulative movement remains sufficiently small, we can ensure that the final estimate $\widehat{\bm{\theta}}_{\infty}$ stays close to the true parameter $\bm{\theta}^\star$. This basic idea is illustrated in Figure~\ref{fig:BallNoCollapse}.

We introduce the following assumption (Assumption~\ref{Ass:Uniform}), which states that the statistical consistency of the estimation procedure $\mathcal{M}(\cdot)$ is uniform across the underlying models $\mathbb{P}_{\bm{\theta}}$. In other words, the convergence behavior of $\mathcal{M}$ does not depend on a specific model, but instead follows a unified pattern that can be leveraged for quantifying its performance.

\begin{assumption}
\label{Ass:Uniform}
For a class of parametric generative models $\mathcal{P} = \{\mathbb{P}_{\bm{\theta}}:\bm{\theta} \in \bm{\Theta}\}$, let \( \mathcal{D} = \{\bm{x}_i\}_{i=1}^n \) be a dataset generated from \( \mathbb{P}_{\bm{\theta}} \) for some \( \bm{\theta} \in \bm{\Theta} \). Suppose that \( \widehat{\bm{\theta}} = \mathcal{M}(\mathcal{D}) \) is an estimate of \( \bm{\theta} \) obtained under the estimation scheme \( \mathcal{M} \). Assume there exist constants $C_1 , C_2, \gamma>0$ and a positive diverging sequence \( r(n) \) such that for all \( n \geq 1 \) and any \( \delta > 0 \), we have  
\begin{equation}
\sup_{\bm{\theta} \in \bm{\Theta}} \mathbb{P}\left( \| \widehat{\bm{\theta}} - \bm{\theta}\|_2 \geq \delta \right) \leq C_1 \exp(-C_2 r(n) \delta^\gamma),
\end{equation}  
where the probability is taken over the randomness of $\mathcal{D}$ generated i.i.d. from $\mathbb{P}_{\bm{\theta}}(\bm{x})$.
\end{assumption}

In Assumption \ref{Ass:Uniform}, we use $r(n)$ and $\gamma$ to characterize the statistical efficiency of the estimation procedure $\mathcal{M}$. A more efficient $\mathcal{M}$ typically exhibits a faster-diverging $r(n)$ and a smaller $\gamma$. Moreover, the forms of $r(n)$ and $\gamma$ are problem-dependent and vary according to the nature of the estimation task. To illustrate this, we provide several examples below.

\begin{example}[Gaussian Estimation]
\label{Exam:Gaus}
  If $\mathcal{P} = \{\mathbb{P}_{\bm{\theta}}=N(\bm{\theta},\bm{I}) \mid \bm{\theta} \in \mathbb{R}^p\}$ and the estimation scheme is $\widehat{\bm{\theta}}=\mathcal{M}(\mathcal{D})=\frac{1}{n}\sum_{i=1}^n \bm{x}_i$ for any given dataset $\mathcal{D} = \{\bm{x}_i\}_{i=1}^n \sim N(\bm{\theta},\bm{I})$, then it holds true for any $\delta>0$ that
  \begin{align}
  \label{GaussEq}
  \sup_{\bm{\theta} \in \mathbb{R}^p}
      \mathbb{P}\left(\Vert \widehat{\bm{\theta}}-\bm{\theta}\Vert_2>\delta\right)  \leq \exp(p/2) \exp\left(-\frac{n\delta^2}{4}\right).
  \end{align}
  In this example, Assumption \ref{Ass:Uniform} holds with $(C_1,C_2,\gamma,r(n))=(\exp(p/2),1/4,2,n)$.
\end{example}

Example \ref{Exam:Gaus} presents a Gaussian case, where $r(n) = n$ and $\gamma = 2$. The tail bound in (\ref{GaussEq}) holds for any ground truth parameter $\bm{\theta}$. It is important to emphasize that the rapid convergence rate depends on the efficiency of mean estimation. In contrast, under the same estimation problem, an inefficient estimation procedure $\mathcal{M}$ may lead to a different tail bound. To further illustrate this, we present an additional example (Example \ref{Exam:Gauss_subopt}) within the same Gaussian estimation framework.
\begin{example}[Inappropriate Estimator]
\label{Exam:Gauss_subopt}
 If $\mathcal{P} = \{\mathbb{P}_{\theta}=N(\theta,1) \mid \theta \in \mathbb{R}\}$ and the estimation scheme is $\widehat{\theta}=\mathcal{M}(\mathcal{D})=\sum_{i=1}^n w_i x_i$ with $w_i = \frac{1/i}{\sum_{i=1}^n 1/i}$, then it holds true for any $\delta>0$ that
 \color{black}
  \begin{align*}
\sup_{\theta \in \mathbb{R}}\mathbb{P}\left(
|\widehat{\theta}-\theta|>\delta
    \right) \leq 2\exp\left(-\frac{3 (\log n)^2\delta^2}{\pi^2}\right).
  \end{align*}
  \color{black}
  In this example, Assumption \ref{Ass:Uniform} holds with \color{black}$(C_1,C_2,\gamma,r(n))=(2,3/\pi^2,2,(\log n)^2)$.\color{black}
\end{example}

Example \ref{Exam:Gauss_subopt} presents a univariate case of Example \ref{Exam:Gaus}, but with a suboptimal estimation procedure $\mathcal{M}$. Using the poorly weighted mean $\widehat{\theta} = \mathcal{M}(\mathcal{D}) = \sum_{i=1}^n w_i x_i$, the form of $r(n)$ becomes $r(n) = (\log n)^2$, which grows more slowly than the $r(n) = n$ in Example \ref{Exam:Gaus}.

\begin{example}[Uniform Distribution Estimation]
\label{Exam:Unif}
  If $\mathcal{P} = \{\mathbb{P}_{\theta}=\mathrm{Unif}(0,\theta) \mid \theta \in [1,M]\}$ and the estimation scheme is $\widehat{\theta}=\mathcal{M}(\mathcal{D})=\max_{i \in [n]} x_{i}$ for any given dataset $\mathcal{D} = \{x_i\}_{i=1}^n \sim \mathrm{Unif}(0,\theta)$, then it holds true that
  \begin{align*}
\sup_{\theta \in [1, M]} \mathbb{P}\left( |\widehat{\theta} - \theta| > \delta \right) \leq \exp\left( -\frac{n\delta}{M} \right).
  \end{align*}
  In this example, Assumption \ref{Ass:Uniform} holds with $(C_1,C_2,\gamma,r(n))=(1,M^{-1},1,n)$.
\end{example}

Example \ref{Exam:Unif} illustrates the estimation of a uniform distribution (specifically, estimating the upper range). In this case, $\gamma = 1$ suggests that the estimation problem is relatively easier, resulting in a faster convergence rate in the tail bound.

\begin{theorem}
\label{Thm:Main}
Denote $D(\delta) = \{\Delta=(\delta_t)_{t=1}^{\infty}:\sum_{t=1}^{\infty} \delta_t \leq \delta\}$. Under Assumption \ref{Ass:Uniform}, for any $\delta > 0$ and $n \geq 1$, it holds that
\color{black}
\begin{align}
\label{MainEqn_Th2}
\lim_{T \rightarrow \infty}
    \mathbb{P}\left(\Vert \widehat{\bm{\theta}}_{T} - \bm{\theta}^\star\Vert_2 > \delta \right)
\leq \inf_{\Delta \in D(\delta)}C_1 \sum_{t=1}^{\infty} \exp(-C_2 r(c_t n )\delta_t^\gamma),
\end{align}
\color{black}
where $C_1$ and $C_2$ are as defined in Assumption \ref{Ass:Uniform}. 
\end{theorem}

Based on Assumption~\ref{Ass:Uniform}, we derive an upper bound for $\lim\limits_{T \rightarrow \infty}\mathbb{P}\left(\Vert \widehat{\bm{\theta}}_{T} - \bm{\theta}^\star\Vert_2 > \delta \right)$ using a straightforward union bound argument. The result in (\ref{MainEqn_Th2}) demonstrates that, as long as there exists a sampling pattern $\{c_{t}\}_{t=1}^{\infty}$ such that $\sum_{t=1}^{\infty} \exp(-C_2 r(c_t n )\delta_t^\gamma)$ is finite for any fixed $n$ and vanishes as $n \to \infty$, the recursive training procedure guarantees that $\Vert \widehat{\bm{\theta}}_{\infty} - \bm{\theta}^\star\Vert_2 \leq \delta$ almost surely. This result is generally true for any estimation problem satisfying Assumption \ref{Ass:Uniform}. Under Theorem \ref{Ass:Uniform}, we provide sufficient conditions on $\{c_{t}\}_{t=1}^{\infty}$ to ensure that $\widehat{\bm{\theta}}_{T}$ remains close to $\bm{\theta}^\star$ almost surely for a specific estimation problem.

\begin{corollary}
\label{corollary_main_thm}
Under Assumption \ref{Ass:Uniform} with $r(n) = n^{\kappa}$ for some $\kappa > 0$, and define  
$c_t \asymp  t^{\frac{\gamma(1+s)}{\kappa}}$ for $t \geq 1$ and $s> 0$. Then, for any $\delta > 0$, we have $\lim\limits_{n\rightarrow \infty}\lim\limits_{T\rightarrow\infty} \mathbb{P} \left( \|\widehat{\bm{\theta}}_{T} - \bm{\theta}^\star\|_2 \geq \delta \right) = 0$.
\end{corollary}

Corollary \ref{corollary_main_thm} provides a sufficient condition on the sequence $\{c_t\}_{t=1}^{\infty}$ to ensure that $\widehat{\bm{\theta}}_\infty$ remains bounded around $\bm{\theta}^\star$ with probability approaching one as $n$ tends to infinity. Specifically, as long as $c_t$ grows at the rate of $t^{\frac{\gamma(1+s)}{\kappa}}$ for some $s > 0$, the estimated parameter $\widehat{\bm{\theta}}_\infty$ will stay close to the true parameter $\bm{\theta}^\star$, thereby preventing model collapse. It is worth noting that this growth pattern inherently depends on $\gamma$ and $\kappa$, which jointly characterize the difficulty of the estimation problem. For more challenging estimation tasks (i.e., larger values of $\gamma$) or under a less effective estimation scheme (i.e., smaller values of $\kappa$), a more rapidly increasing sequence ${c_t}$ is required to prevent model collapse.

Furthermore, we would like to emphasize that Corollary \ref{corollary_main_thm} provides only a sufficient condition, derived via the union bound, for a broad class of estimation problems that satisfy Assumption \ref{Ass:Uniform}. However, for certain specific estimation problems, the conditions imposed by Corollary \ref{corollary_main_thm} can be relaxed. In particular, a slower growth rate of the sequence $\{c_t\}_{t=1}^\infty$ may still suffice to prevent model collapse. To illustrate this, we present a Gaussian example (Example \ref{Exam:Gaussian}) demonstrating that a more refined pattern of $\{c_t\}_{t=1}^{\infty}$ can be derived through a sharper analysis of random walk behavior.

\begin{example}
\label{Exam:Gaussian}
   Consider a recursive Gaussian mean estimation process: $\widehat{\bm{\theta}}_t = \mathcal{M}(\mathcal{D}_{t-1}) = \frac{1}{n_{t-1}} \sum_{i=1}^{n_{t-1}} \bm{x}_{t-1,i}$, where $\bm{x}_{t-1,i} \sim N(\widehat{\bm{\theta}}_{t-1}, \bm{I}_p)$ and $\bm{x}_{0,i} \sim N(\bm{\theta}^\star, \bm{I}_p)$. In this case, Corollary \ref{corollary_main_thm} holds true with $(\kappa, \gamma) = (1,2)$, and Corollary \ref{corollary_main_thm} shows that using $c_t \asymp t^{2 + s}$ with $s > 0$ ensures
\[
    \lim_{n\rightarrow \infty}\lim_{T\rightarrow \infty}
    \mathbb{P}\left(
    \Vert \widehat{\bm{\theta}}_T - \bm{\theta}^\star \Vert_2 \geq \delta
    \right)  = 0,
\]
for any $\delta > 0$. However, a sharper analysis in this example reveals that
\[
    \mathbb{P}\left(
    \Vert \widehat{\bm{\theta}}_T - \bm{\theta}^\star \Vert_2 \geq \delta
    \right) \leq \exp(p)\exp\left(-\frac{n\delta^2}{\sum_{t=0}^{T-1}\frac{1}{c_t}}\right).
\]
Clearly, if $c_t = t^{1 + s}$ for any $s > 0$, the series $\sum_{t=1}^{\infty} c_t^{-1}$ converges, which implies
\[
    \lim_{n\rightarrow \infty}\lim_{T\rightarrow \infty}
    \mathbb{P}\left(
    \Vert \widehat{\bm{\theta}}_T - \bm{\theta}^\star \Vert_2 \geq \delta
    \right)  = 0,
\]
under this sampling pattern. 
\end{example}

Example \ref{Exam:Gaussian} illustrates the behavior of a Gaussian random walk. Interestingly, in this case, the synthetic data expansion pattern relaxes from $c_t \asymp t^{2+s}$ (as suggested by Corollary \ref{corollary_main_thm}) to $t^{1+s}$. The key reason is that, during the recursive training process, the estimated parameter moves back and forth in the parameter space, rather than proceeding along a deterministic and steadily progressing trajectory, as implicitly assumed in Corollary \ref{corollary_main_thm}. In other words, Corollary \ref{corollary_main_thm} provides a sharp condition for preventing model collapse when $\mathcal{M}(\cdot)$ exhibit a large estimation bias. To illustrate this, we present Example \ref{Exam:Gaussian_unbound} below.

\begin{example}
\label{Exam:Gaussian_unbound}
   Consider a recursive biased Gaussian mean estimation process: $$
   \textnormal{Biased Mean Estimation: }
   \widehat{\bm{\theta}}_t = \mathcal{M}(\mathcal{D}_{t-1}) = \frac{1}{n_{t-1}} \sum_{i=1}^{n_{t-1}} \bm{x}_{t-1,i}+\frac{1}{\sqrt{n_{t-1}}} \cdot \bm{1}_p,
   $$
   where $\bm{x}_{t-1,i} \sim N(\widehat{\bm{\theta}}_{t-1}, \bm{I}_p)$ and $\bm{x}_{0,i} \sim N(\bm{\theta}^\star, \bm{I}_p)$. In this setting, the estimation bias is $\Vert \mathbb{E}(\widehat{\bm{\theta}}_t|\widehat{\bm{\theta}}_{t-1})-\widehat{\bm{\theta}}_{t-1} \Vert_2\asymp n^{-\frac{1}{2}}$. Corollary \ref{corollary_main_thm} holds with $(\kappa, \gamma) = (1, 2)$. Specifically, if $c_t \asymp t^{a}$, then we have 
   \begin{align*} \lim_{n \rightarrow \infty} \lim_{T \rightarrow \infty} \mathbb{P}\left(\Vert\widehat{\bm{\theta}}_{T} - \bm{\theta}^\star\Vert_2 \geq \delta\right) = \begin{cases} 0, & \text{if } c_t \asymp t^{a} \text{ with } a > 2, \\ 1, & \text{if } c_t \asymp t^{a} \text{ with } 1 < a \leq 2. \end{cases} \end{align*} This result demonstrates that the result given in Corollary \ref{corollary_main_thm} is sharp.
\end{example}

In Example~\ref{Exam:Gaussian_unbound}, we consider a recursive training process for biased Gaussian mean estimation. At each estimation step~$t$, the estimator is adjusted by an additional term $\frac{1}{\sqrt{n_{t-1}}} \cdot \bm{1}_p$. This procedure implies that the resulting estimator consistently tends to drift in the direction of $\bm{1}_p = (1, \ldots, 1)$. In Example~\ref{Exam:Gaussian_unbound}, the sufficient condition specified in Corollary~\ref{corollary_main_thm} is sharp, demonstrating that the corollary provides a tight characterization in certain cases. Furthermore, Example~\ref{Exam:Gaussian_unbound} illustrates a phase transition in the synthetic sampling schedule, where model collapse arises when \( c_t \asymp t^a \) for \( a \leq 2 \), and is alleviated when \( a > 2 \).

Clearly, Examples~\ref{Exam:Gaussian} and~\ref{Exam:Gaussian_unbound} differ in their estimation procedures and exhibit distinct convergence conditions for the sequence $\{c_t\}_{t=1}^\infty$. This observation suggests that the presence of bias in the estimator accelerates model collapse during recursive training, as the random walk consistently drifts in a specific direction. In particular, when estimation bias is present, the condition for avoiding model collapse specified in Corollary~\ref{corollary_main_thm} appears to be sharp.

\subsection{Martingale Processes in Random Walks}
In this section, we investigate the behavior of recursive training with the goal of deriving sharper synthetic data expansion schemes, which are not covered by Corollary \ref{corollary_main_thm}. The key motivation stems from the contrast between Examples \ref{Exam:Gaussian} and \ref{Exam:Gaussian_unbound}. The possibility of achieving a more refined analysis of synthetic data expansion schemes is primarily attributed to the unbiasedness or some ignorable estimation bias.

Consider a general $T$-step recursive training procedure, where the difference $\widehat{\bm{\theta}}_T - \bm{\theta}^\star$ can be decomposed as  
\begin{align}
\label{Decomposition}
    \widehat{\bm{\theta}}_T - \bm{\theta}^\star 
    = \left(\widehat{\bm{\theta}}_T - \widehat{\bm{\theta}}_{T-1}\right) 
    + \left(\widehat{\bm{\theta}}_{T-1} - \widehat{\bm{\theta}}_{T-2}\right) 
    + \cdots + \left(\widehat{\bm{\theta}}_1 - \bm{\theta}^\star\right) = \sum_{t=1}^T \bm{\xi}_t,
\end{align}
where $\bm{\xi}_t = \widehat{\bm{\theta}}_t - \widehat{\bm{\theta}}_{t-1}$ for $t \geq 2$, and $\bm{\xi}_1 = \widehat{\bm{\theta}}_1 - \bm{\theta}^\star$.

From the decomposition in (\ref{Decomposition}), the difference $\widehat{\bm{\theta}}_T - \bm{\theta}^\star$ can be expressed as the cumulative sum of the variables, that is $\widehat{\bm{\theta}}_T - \bm{\theta}^\star = \sum_{t=1}^T \bm{\xi}_t$. If the estimation procedure $\mathcal{M}(\cdot)$ is unbiased, then we can show that the sequence $\{\bm{\xi}_t\}_{t=1}^T$ forms a discrete-time martingale, that is,
\[
\mathbb{E}(\bm{\xi}_t \mid \bm{\xi}_1, \ldots, \bm{\xi}_{t-1}) = 0, \text{ for any } t \geq 1.
\]
In the following, we begin by considering the case of unbiased estimators. Specifically, we make the assumption (Assumption \ref{Ass:Unbiased}) that the estimation procedure is unbiased, which implies that $\{\bm{\xi}_{t}\}_{t=1}^{T}$ forms a discrete-time martingale.

\begin{assumption}
    \label{Ass:Unbiased}
Assume that the estimation procedure is unbiased, meaning that for any dataset $\mathcal{D} = \{\bm{x}_{i}\}_{i=1}^n$ drawn from $\mathbb{P}_{\bm{\theta}}$ for some $\bm{\theta}$, it holds that $\mathbb{E}(\widehat{\bm{\theta}}) = \mathbb{E}(\mathcal{M}(\mathcal{D})) = \bm{\theta}$, where the expectation is taken with respect to the randomness of $\mathcal{D}$.
\end{assumption}

\begin{theorem}
\label{Thm:Martingale}
Under Assumption~\ref{Ass:Uniform} with $r(n)=n^\kappa$ and $\kappa \geq \gamma / 2$ and Assumption \ref{Ass:Unbiased}, the following holds for any $\delta>0$,
\begin{align}
\label{Eqn_Martin}
    \lim_{n \to \infty} \lim_{T \to \infty} \mathbb{P}\left( \|\widehat{\bm{\theta}}_T - \bm{\theta}^\star\|_2 \geq \delta \right) = 0,
\end{align}
for a sequence $c_t = t^{1+s}$ with any $s > 0$.
\end{theorem}

In Theorem \ref{Thm:Martingale}, we show that as long as $\kappa \geq \gamma/2$, the optimal sample size schedule for ensuring that $\widehat{\bm{\theta}}_T$ is centered around $\bm{\theta}^\star$ when $T\rightarrow \infty$ is given by $c_t = t^{1+s}$ for some $s > 0$. It is worth noting that Theorem \ref{Thm:Martingale} is quite general, as it applies to all unbiased estimation problems that satisfy Assumption \ref{Ass:Uniform} with $\kappa \geq \gamma/2$. It is worth noting that $\kappa = \gamma/2$ essentially implies the $\sqrt{n}$-consistency of $\mathcal{M}$. Particularly, it includes the Gaussian mean estimation (Example \ref{Exam:Gaussian}) as a special case, where $(\kappa, \gamma) = (1, 2)$.

Nevertheless, the martingale property is not a necessary condition for achieving the optimal sample size schedule \( c_t = t^{1+s} \). Intuitively, if the estimation bias is sufficiently small, the requirement of unbiasedness can be relaxed. Motivated by this observation, we introduce Assumption~\ref{Ass:biased}, which relaxes the unbiasedness condition in Assumption~\ref{Ass:Unbiased} and allows the estimation procedure \( \mathcal{M} \) to exhibit a small amount of bias.

\begin{assumption}
\label{Ass:biased}
Assume that the estimation procedure $\mathcal{M}(\cdot)$ satisfies the following condition: for any dataset $\mathcal{D} = \{\bm{x}_{i}\}_{i=1}^n$ drawn from $\mathbb{P}_{\bm{\theta}}$ for some $\bm{\theta}$, it holds that $ |\mathbb{E}(\widehat{\theta}_i) - \theta_i| 
    \asymp \frac{v_i}{n^\rho}$ for some positive constants $\rho$ and $v_i$, and for any $i \in [p]$ and $\bm{\theta} \in \bm{\Theta}$.
\end{assumption}

In Assumption~\ref{Ass:biased}, we assume that for the \( i \)-th coordinate of the parameter vector, the estimation bias is bounded by \( \frac{v_i}{n^\rho} \), where \( \rho \) is a positive constant and \( v_i \) is a fixed constant depending only on the $i$-th dimension. A larger value of \( \rho \) corresponds to a smaller estimation bias. This form of bias commonly arises in practical estimation problems. To illustrate this, we present an example based on the exponential distribution below.

\begin{example}
    \label{Exam:Exponential}
    Let $\mathcal{D}=\{x_{i}\}_{i=1}^n$ be a dataset generated from a exponential distribution with density function $p_{\theta}(x)=\theta e^{-\theta x}$ for some $\theta \in [1,M]$. Let $\widehat{\theta}=\mathcal{M}(\mathcal{D})=\argmin_{\theta \in \mathbb{R}} -\sum_{i=1}^n\log p_{\theta}(x_i)$ be the maximum likelihood estimate (MLE). Then it holds that $\mathbb{E}(\widehat{\theta})-\theta = \frac{\theta}{n-1} \asymp \frac{2M}{n}$ for $n \geq 2$. Clearly, the estimation procedure $\mathcal{M}(\cdot)$ satisfies Assumption~\ref{Ass:biased} with $\rho=1$. 
\end{example}

In Example~\ref{Exam:Exponential}, we illustrate the bias of the MLE for the exponential distribution, where the estimation bias satisfies Assumption~\ref{Ass:biased} with $\rho=1$. This result can be readily verified for many other estimation problems involving MLEs \citep{firth1993bias}. In particular, Example~\ref{Exam:Gaussian_unbound} satisfies Assumption~\ref{Ass:biased} with \( \rho = 1/2 \).

\begin{lemma}
\label{Lemma:SpeedBound}
For a class of models \( \mathcal{P} = \{\mathbb{P}_{\bm{\theta}} : \bm{\theta} \in \bm{\Theta}\} \), let \( \mathcal{D} = \{\bm{x}_i\}_{i=1}^n \) denote a dataset generated from \( \mathbb{P}_{\bm{\theta}} \) for some \( \bm{\theta} \in \bm{\Theta} \). If the estimation procedure \( \widehat{\bm{\theta}} = \mathcal{M}(\mathcal{D}) \) satisfies Assumption \ref{Ass:Uniform} with \( r(n) = n^\kappa \) and Assumption \ref{Ass:biased} simultaneously, then it holds that  $\rho \geq \kappa/\gamma$.
\end{lemma}

A natural question arises: what determines the value of \( \rho \) in practice? We emphasize that \( \rho \) is fundamentally constrained by the rate of the estimation error. In particular, Lemma~\ref{Lemma:SpeedBound} shows that if the estimation procedure \( \mathcal{M}(\cdot) \) satisfies Assumption~\ref{Ass:Uniform} with \( r(n) = n^{\kappa} \), then \( \rho \) must be bounded below by \( \kappa / \gamma \). This relationship stems from the fact that, given a fixed convergence rate for the estimator, the bias cannot grow too rapidly. Moreover, this conclusion is consistent with the classical bias--variance tradeoff in estimation theory.

\begin{theorem}
   \label{Thm:positive_bias}
Under Assumption~\ref{Ass:Uniform} with $r(n)=n^\kappa$ and $\kappa \geq \gamma / 2$ and Assumption \ref{Ass:biased} with $\rho>0$, we have the following results:
\begin{itemize}
    \item[(1)] \textnormal{\textbf{(Small Bias)}} If $\rho \geq 1$, then
    $
    \lim_{n \to \infty} \lim_{T \to \infty} \mathbb{P}\left( \|\widehat{\bm{\theta}}_T - \bm{\theta}^\star\|_2 \geq \delta \right) = 
        0,
    $
    for any $c_t =  t^{1+s}$ with $s > 0$.
    \item[(2)] \textnormal{\textbf{(Large Bias)}} If $\frac{\kappa}{\gamma} \leq \rho <1$, we have
    \begin{align*}
        \lim_{n \to \infty} \lim_{T \to \infty} \mathbb{P}\left( \|\widehat{\bm{\theta}}_T - \bm{\theta}^\star\|_2 \geq \delta \right) = 
        \begin{cases}
            0, \text{ If }  c_t = t^{1+s} \text{ with } s>\frac{1}{\rho}-1, \\
            1, \text{ If }  c_t = t^{1+s} \text{ with } 0<s<\frac{1}{\rho}-1.\\
        \end{cases}
    \end{align*}
\end{itemize}
\end{theorem}

In Theorem \ref{Thm:positive_bias}, we demonstrate that if the estimation bias is small, i.e., $\rho \geq 1$, the sample size schedule is optimal, requiring only $c_t = t^{1+s}$ with $s > 0$ and the bias will eventually be compensated as \( n \to \infty \). However, when the estimation bias is large, i.e., $\frac{\kappa}{\gamma} \leq \rho < 1$, more synthetic samples are necessary to stabilize the recursive training process. In this case, the sample size schedule must follow $c_t = t^{1+s}$, with $s > \rho^{-1} - 1$. This result reveals an interesting phenomenon: biased estimation may accelerate the model collapse rate if the estimation bias is large, thereby necessitating more synthetic samples to mitigate this effect. This claim is supported by our simulation results (see \textbf{Scenario 2}). Additionally, it is worth noting that Theorem~\ref{Thm:positive_bias} considers the case where the estimation bias vanishes as the sample size \( n \) increases (Assumption \ref{Ass:biased}). If the estimation bias remains fixed regardless of \( n \), then no synthetic data schedule can prevent model collapse.

\section{Can Synthetic Data Produce Better Models?}
\label{Sec:MLE_Study_Case}
In this section, we address \textbf{Q2} as specified in Section \ref{Sec:RW}, aiming to explore whether recursive training produces a better model compared to training solely on the initial model learned from purely real data. In the context of parametric models, this question reduces to investigating the following probability:
\begin{align*}
\text{Model $\mathbb{P}_{\widehat{\bm{\theta}}_T}$ is Better than $\mathbb{P}_{\widehat{\bm{\theta}}_1}$: } P(T) \triangleq
        \mathbb{P}\big(\Vert \widehat{\bm{\theta}}_T - \bm{\theta}^\star \Vert_2 < \Vert \widehat{\bm{\theta}}_1 - \bm{\theta}^\star \Vert_2\big).
\end{align*}
Here, \( P(T) \) denotes the probability that, after a \( T \)-step recursive training process, the resulting generative model \( \mathbb{P}_{\widehat{\boldsymbol{\theta}}_T} \) is better than the initial model \( \mathbb{P}_{\widehat{\boldsymbol{\theta}}_1} \) in terms of parameter estimation.

To investigate the quantitative relationship between \( P(T) \), the total sample size \( n \), and the sample size schedule \( c_t \), we begin with the Gaussian mean setting as a warm-up.

\begin{theorem}
    \label{Thm:GaussianBetter}
    Consider a recursive multivariate Gaussian mean estimation process: $\widehat{\bm{\theta}}_t = \mathcal{M}(\mathcal{D}_{t-1}) = \frac{1}{n_{t-1}} \sum_{i=1}^{n_{t-1}} \bm{x}_{t-1,i}$, where $\bm{x}_{t-1,i} \sim N(\widehat{\bm{\theta}}_{t-1}, \bm{\Sigma})$ and $\bm{x}_{0,i} \sim N(\bm{\theta}^\star, \bm{\Sigma})$. For any $T\geq 2$, it holds that
    \begin{align}
    \label{Prob_Better}
        P(T)= \mathbb{E}_{\bm{X}}\left[\Phi\left(-\frac{\|\bm{X}\|_2}{2\Vert \bm{\Sigma}^{\frac{1}{2}} \widetilde{\bm{X}}_T \Vert_2}\right)\right],
    \end{align}
    where $\widetilde{\bm{X}}_T = \bm{X}/\Vert\bm{X}\Vert_2$ with $\bm{X} \sim N(\bm{0},\sum_{t=1}^{T-1}c_t^{-1}\bm{\Sigma})$ and $\Phi(x)= \int_{-\infty}^x \frac{1}{\sqrt{2\pi}} e^{-\frac{t^2}{2}} dt$ denotes the cumulative distribution function (CDF) of the standard normal distribution. Let $\lambda_{\min}$ and $\lambda_{\max}$ denote the minimum and maximum eigenvalues of $\bm{\Sigma}$, respectively. It then follows that
    \begin{align*}
  \mathbb{E}_{\bm{X}}\left[\Phi\left(-\frac{\|\bm{X}\|_2}{2\lambda_{\min}}\right)\right]  \leq 
        P(T)  \leq \mathbb{E}_{\bm{X}}\left[\Phi\left(-\frac{\|\bm{X}\|_2}{2\lambda_{\max}}\right)\right].
    \end{align*}
\end{theorem}

In Theorem \ref{Thm:GaussianBetter}, we show that in the problem of recursive Gaussian estimation, the probability of obtaining a better Gaussian model after $T$ steps of training is given by (\ref{Prob_Better}), from which several interesting results can be derived.
\begin{itemize}
    \item[(1)] \( P(T) \) is independent of \( n \), indicating that regardless of the real dataset $\mathcal{D}_0$ used at the beginning of training, the probability of obtaining a better model through recursive training remains unaffected by the real sample size \( n \).
    \item[(2)] An increasing sample scheme $\{c_t\}_{t=1}^{\infty}$ allows for higher probabilities of achieving a better model within a $T$-step recursive training. This is because when the $c_t$ values are large, the sum $\sum_{t=1}^{T-1} c_t^{-1}$ becomes smaller, causing $\Vert \bm{X} \Vert_2$ to be more tightly concentrated around zero with higher probability, which in turn leads to larger values of \( P(T) \). However, it is also worth noting that \( P(T) \) is always bounded above by $1/2$, regardless of the sample scheme, since $\Phi(x) < 1/2$ for any $x < 0$.
    \item[(3)] If $c_t$ is chosen such that $c_t = t^{1+s}$ for some $s > 0$, then we have $\sum_{t=1}^{\infty} c_t^{-1} < \infty$. This indicates that \( P(T) \) will stabilize at a nonzero value as \( T \rightarrow \infty \). A direct application of this result is that, when considering a recursive training process with an adaptive increasing sample scheme \( c_t \gtrsim t^{1+s} \), the probability of obtaining a better model after 1000 steps and after 2000 steps of recursive training might be essentially the same.
\end{itemize}

\begin{corollary}
\label{Coro:Identity}
Let $v = \sum_{t=1}^{T-1}c_t^{-1}$. Under the setting in Theorem \ref{Thm:GaussianBetter} with $\bm{\Sigma}=\bm{I}_p$, it holds that
    \begin{align*}
     \Phi\left(-\frac{\sqrt{vp}}{2}\right) \leq    P(T) \leq \frac{\sqrt{\pi}\,
        \Gamma\!\bigl(\tfrac{p-1}{2}\bigr)}
       {2^{\,p/2-1/2}\,v^{p/2}\,\Gamma(p/2)}\;
  \Bigl(\tfrac{8v}{v+4}\Bigr)^{\!\frac{p-1}{2}}.
    \end{align*}
Particularly, if $c_t = t^{1+s}$, then $\lim\limits_{T \rightarrow \infty} P(T)$ is bounded away from zero.
\end{corollary}

In Corollary \ref{Coro:Identity}, we present a special case of Theorem \ref{Thm:GaussianBetter} where $\bm{\Sigma}$ is the identity matrix. In this example, we provide more explicit forms of the upper and lower bounds for $P(T)$, allowing us to better understand how the sample schedule $c_t$ and the data dimension affect $P(T)$. It is worth noting that if $c_t = t^{1+s}$, then $P(T)$ will stabilize at a fixed positive value, indicating that the probability of obtaining a better model remains approximately the same for sufficiently large $T$.

In the following, we aim to extend the result in (\ref{Prob_Better}) to more general estimation problems, going beyond the normality assumption and providing a broader perspective on \( P(T) \). The main challenge is that, in general estimation problems, the distribution of the estimators is unknown, making it highly difficult to exactly characterize \( P(T) \) in the general case. Nevertheless, if the estimators satisfy asymptotic normality, we can use (\ref{Prob_Better}) as an approximation. Therefore, we impose Assumption \ref{Ass:AssNorm} on the estimation procedure \( \mathcal{M}(\cdot) \).

\begin{assumption}[Asymptotic Normality of $\mathcal{M}(\cdot)$]
    \label{Ass:AssNorm}
    Consider a class of parametric models \( \mathcal{P} = \{\mathbb{P}_{\bm{\theta}} : \bm{\theta} \in \bm{\Theta}\} \), let \( \mathcal{D} = \{\bm{x}_i\}_{i=1}^n \) be a dataset generated from \( \mathbb{P}_{\bm{\theta}} \) for some \( \bm{\theta} \in \bm{\Theta} \). Suppose that \( \widehat{\bm{\theta}} = \mathcal{M}(\mathcal{D}) \) satisfies that $\sqrt{n}(\widehat{\bm \theta}_n - \bm{\theta}) \xrightarrow{d} N(\bm{0}, \bm{\Sigma}(\bm{\theta})),
    $ where \( \bm{\Sigma}(\bm{\theta}) \) is a covariance matrix depending on $\bm{\theta}$ and continuous with respect to \( \bm{\theta} \).
\end{assumption}

Assumption \ref{Ass:AssNorm} imposes a mild requirement on the asymptotic normality of the estimation procedure \( \mathcal{M}(\cdot) \), which additionally ensures \( \sqrt{n} \)-consistency. In practice, such conditions are often satisfied in parametric models with well-specified structures, such as maximum likelihood estimation (MLE) within exponential families, making Assumption \ref{Ass:AssNorm} a reasonable and broadly applicable foundation for subsequent analysis. In particular, if $\mathcal{M}(\cdot)$ denotes the MLE, then $\bm{\Sigma}(\bm{\theta})$ typically corresponds to the Fisher information matrix. Under Assumption \ref{Ass:AssNorm}, we can extend the result of Theorem \ref{Thm:GaussianBetter} to estimation procedures with asymptotic normality.

\begin{theorem}
    \label{Thm:General}
Consider a general parametric recursive training process: $\widehat{\bm{\theta}}_{t+1} = \mathcal{M}(\mathcal{D}_{t})$ with $\mathcal{D}_t \sim \mathbb{P}_{\widehat{\bm{\theta}}_t}$. If the estimation procedure satisfies Assumption \ref{Ass:AssNorm}, it then follows that
    \begin{align}
    \label{Approx}
        \lim_{n\rightarrow \infty}P(T) =  \mathbb{E}_{\bm{X}}\left[\Phi\left(-\frac{\|\bm{X}\|_2}{2\Vert \bm{\Sigma}^{\frac{1}{2}}(\bm{\theta}^\star) \widetilde{\bm{X}}_T \Vert_2}\right)\right],
    \end{align}
    where $\widetilde{\bm{X}}_T = \bm{X}/\Vert\bm{X}\Vert_2$, $\bm{X} \sim N(\bm{0},\sum_{t=1}^{T-1}c_t^{-1}\bm{\Sigma}(\bm{\theta}^\star))$.
\end{theorem}

In Theorem \ref{Thm:General}, we characterize the quantity \(P(T)\) for a general recursive training process, assuming that the underlying estimation procedure satisfies the asymptotic normality condition stated in Assumption \ref{Ass:AssNorm}. This result generalizes Theorem \ref{Thm:GaussianBetter} beyond the Gaussian setting, showing that, under Assumption \ref{Ass:AssNorm}, the probability of obtaining an improved model after \(T\) steps of recursive training can be asymptotically approximated by expression~(\ref{Approx}). This closed-form approximation renders the result broadly applicable to well-behaved parametric estimators, such as maximum likelihood estimation in exponential family models.

\section{Experiments}
\label{Sec:Exp}

\subsection{Simulations}
In this section, we conduct a series of simulation studies to validate our theoretical results.

\noindent
\textbf{Scenario 1 (Data Expansion Avoids Model Collapse).} In this scenario, we aim to demonstrate that the superlinear synthetic pattern \( c_t = t^{1+s} \) can effectively avoids model collapse occurs in a class of general parametric estimation problems. We investigate the recursive behavior of one-parameter maximum likelihood estimators (MLEs) under synthetic data regeneration. Specifically, we consider three distributions with known closed-form MLEs: (1) the Gamma distribution with a fixed shape parameter \( k = 2 \), where the unknown parameter is the scale \( \theta \); (2) the Exponential distribution with an unknown rate parameter \( \theta \); (3) the Normal distribution with unknown mean $\theta$.

In each experiment, we initialize with \( n=100 \) real samples drawn from the true distribution with parameter \( \theta^\star \), and compute the initial estimate \( \hat{\theta}_0 \) using the corresponding MLE. Then, for each iteration \( t = 1, \ldots, T \) with \( T = 2,000 \), we generate \( n_t = c_t \cdot n_0 \) synthetic samples using \( \widehat{\theta}_{t} \), compute a new estimate \( \hat{\theta}_t \), and repeat the process. We consider two synthetic data patterns: (1) \( c_t = 1 \) and (2) \( c_t = t^{1.1} \).

\begin{figure}[htbp]
  \centering
  \begin{subfigure}[b]{0.31\textwidth}
    \includegraphics[width=\textwidth]{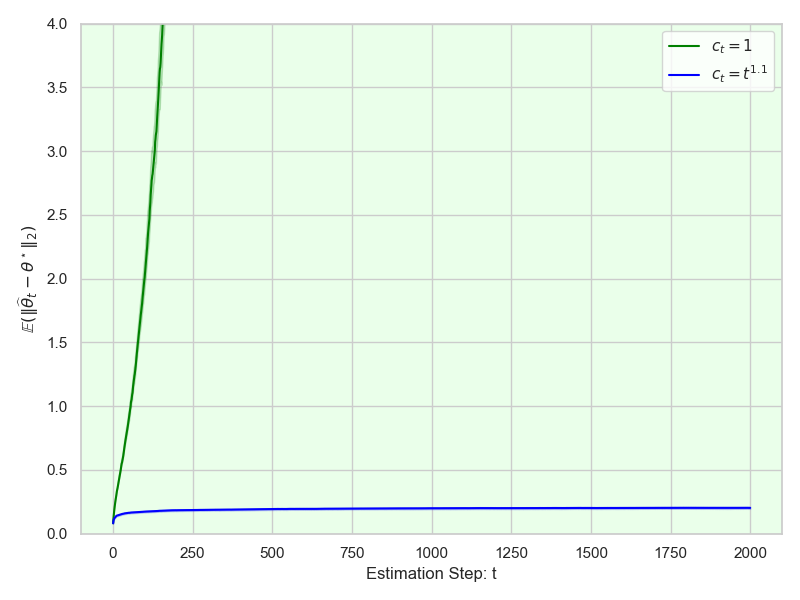}
  \end{subfigure}
  \begin{subfigure}[b]{0.31\textwidth}
    \includegraphics[width=\textwidth]{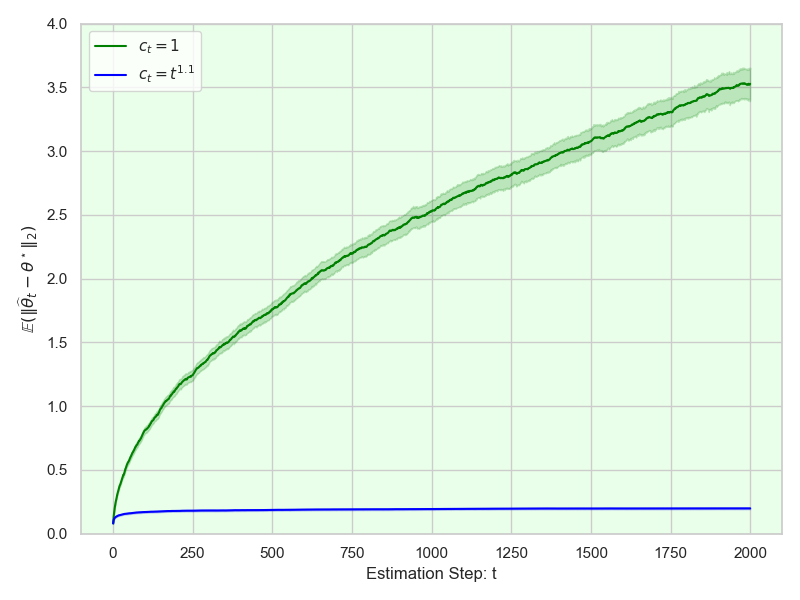}
  \end{subfigure}
  \begin{subfigure}[b]{0.31\textwidth}
    \includegraphics[width=\textwidth]{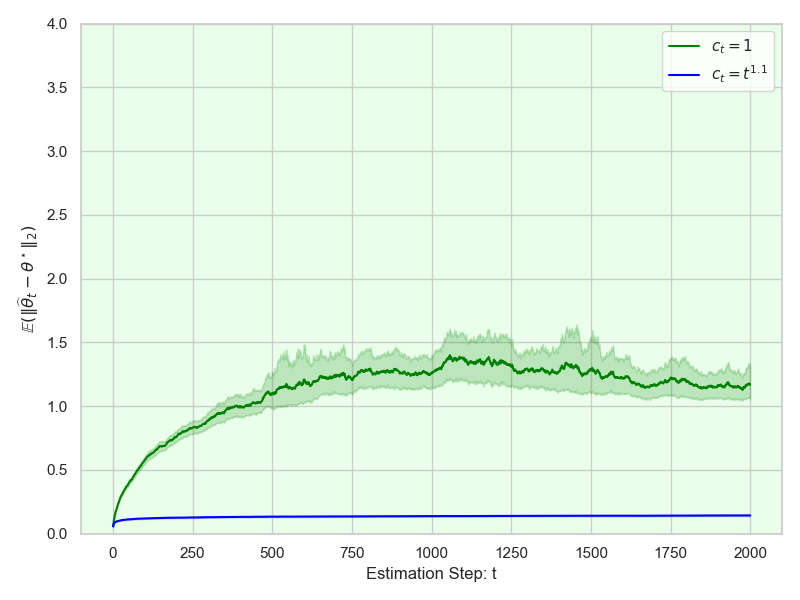}
  \end{subfigure}
  \begin{subfigure}[b]{0.31\textwidth}
    \includegraphics[width=\textwidth]{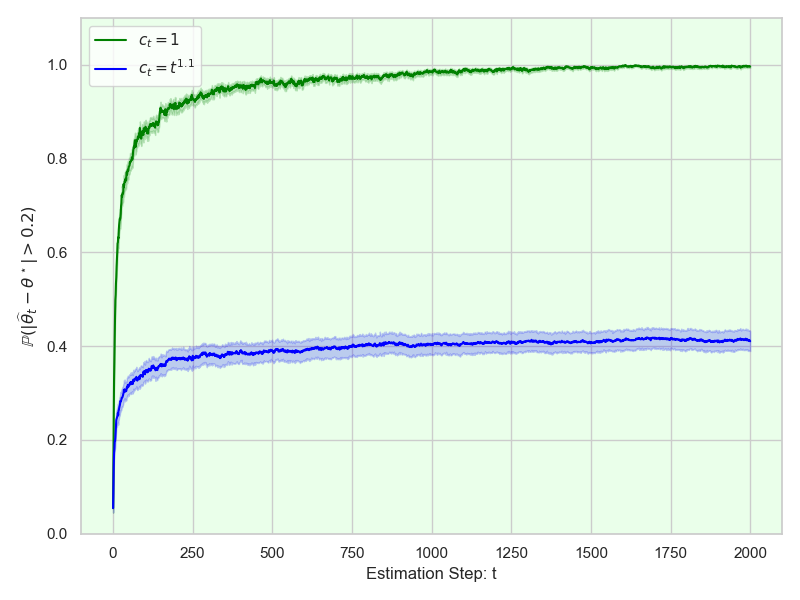}
    \caption{Exponential}
  \end{subfigure}
  \begin{subfigure}[b]{0.31\textwidth}
    \includegraphics[width=\textwidth]{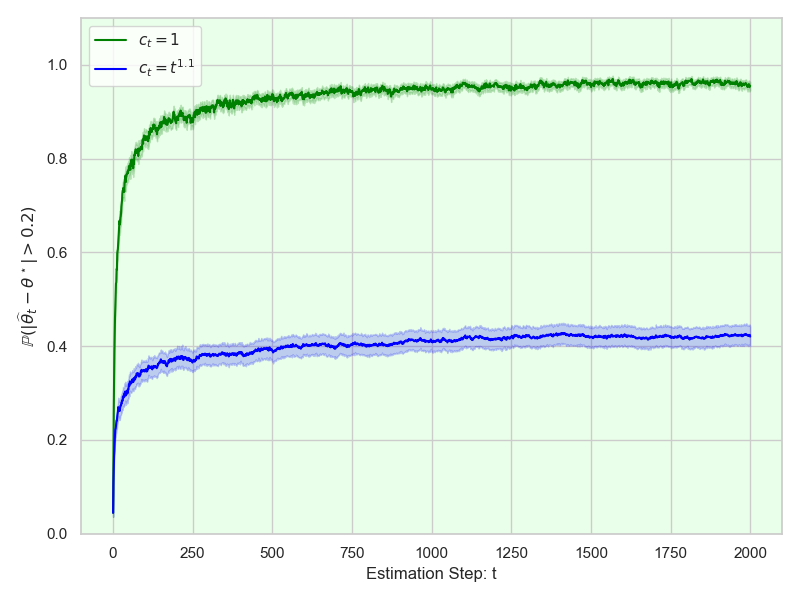}
   \caption{Normal}
  \end{subfigure}
  \begin{subfigure}[b]{0.31\textwidth}
    \includegraphics[width=\textwidth]{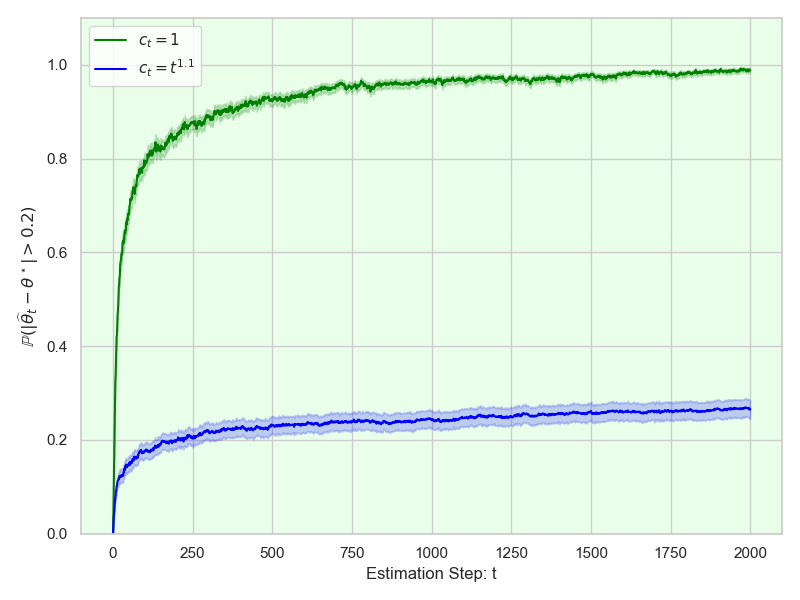}
    \caption{Gamma}
  \end{subfigure}

\caption{Experimental results for \textbf{Scenario 1} are presented, with 95\% confidence intervals plotted for each case.}
\label{Fig:Sce1}
\end{figure}

As shown in Figure~\ref{Fig:Sce1}, under the equal sample schedule \( c_t = 1 \), the mean squared errors of all three types of MLEs diverge, and the estimators escape from the true parameter \( \theta^\star \) with probability approaching one as \( t \) increases. In contrast, under the expansion scheme \( c_t = t^{1.1} \), the mean squared errors stabilize at fixed values, and the probability that the estimators remain close to \( \theta^\star \) converges to a value strictly less than one. This suggests that with an appropriate data expansion scheme, the estimators may still deviate from the true parameter during recursive training, but not almost surely.

\noindent
\textbf{Scenario 2 (Synthetic Data Produces Better Model).} In this setting, our goal is to validate the theoretical results stated in Theorems~\ref{Thm:GaussianBetter} and~\ref{Thm:General}. Specifically, we aim to show that the probability of obtaining a better model after \( T \)-step recursive training is characterized by Equation~\eqref{Prob_Better} in the Gaussian case, and by Equation~\eqref{Approx} in more general settings under asymptotic normality.

We begin by considering the Gaussian setting, where \( N(\bm{0}, \bm{I}) \) represents the real distribution. Specifically, we examine three initial sample sizes, \( n \in \{100, 200, 400\} \), and three synthetic data generation schemes: \( c_t \equiv 1 \), \( c_t = t \), and \( c_t = t^{1.5} \). As a baseline, we approximate the quantity $\mathbb{E}_{\bm{X}}\left[\Phi\left(-\frac{\|\bm{X}\|_2}{2 \|\bm{\Sigma}^{1/2} \widetilde{\bm{X}}_T \|_2}\right)\right]$ using Monte Carlo simulation. For each combination of \( (n, c_t) \), we perform 10-step recursive training and repeat the process \( 10^6 \) times. Specifically, for each pair \( (n, c_t) \), we obtain \( \bm{\theta}_T^{(i)} \) in the \( i \)-th replication, and estimate \( P(T) \) as follows:
\[
\widehat{P(T)} = \frac{1}{10^6} \sum_{i=1}^{10^6} I\left( \|\widehat{\bm{\theta}}_T^{(i)} - \bm{\theta}^\star \|_2 < \|\widehat{\bm{\theta}}_1^{(i)} - \bm{\theta}^\star \|_2 \right), \quad \text{for } 2 \leq T \leq 10.
\]

\begin{figure}[htbp]
  \centering
  \begin{subfigure}[b]{0.31\textwidth}
    \includegraphics[width=\textwidth]{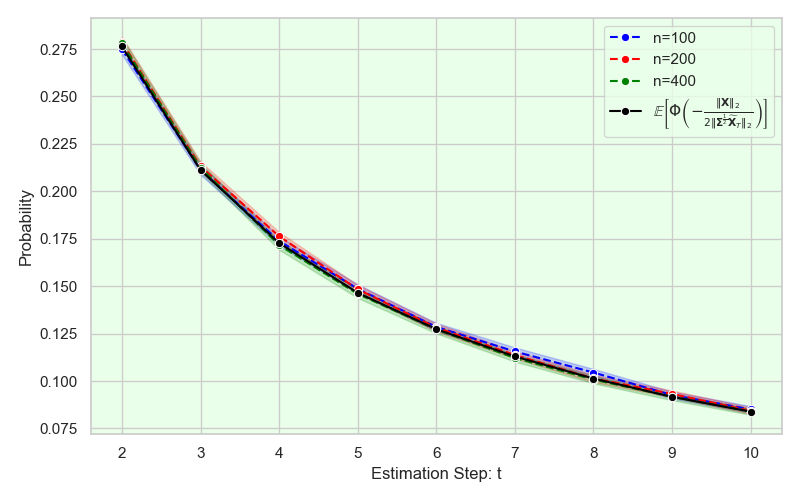}
   \caption{$c_t = 1$}
  \end{subfigure}
  \begin{subfigure}[b]{0.31\textwidth}
    \includegraphics[width=\textwidth]{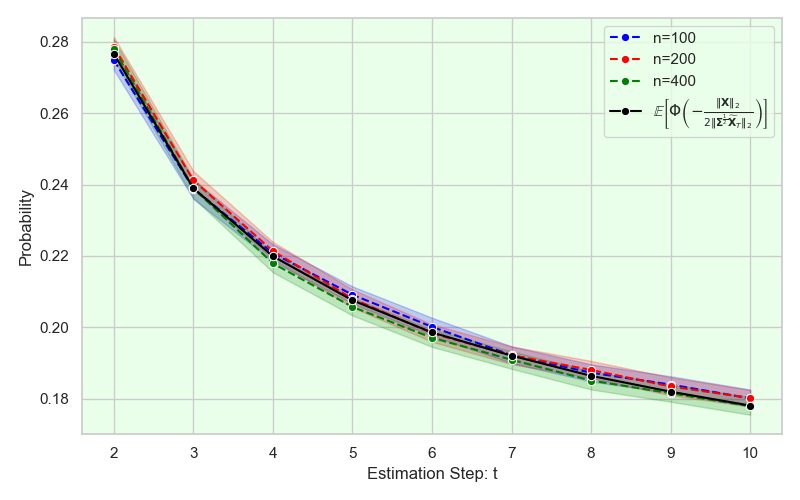}
    \caption{$c_t = t$}
  \end{subfigure}
  \begin{subfigure}[b]{0.31\textwidth}
    \includegraphics[width=\textwidth]{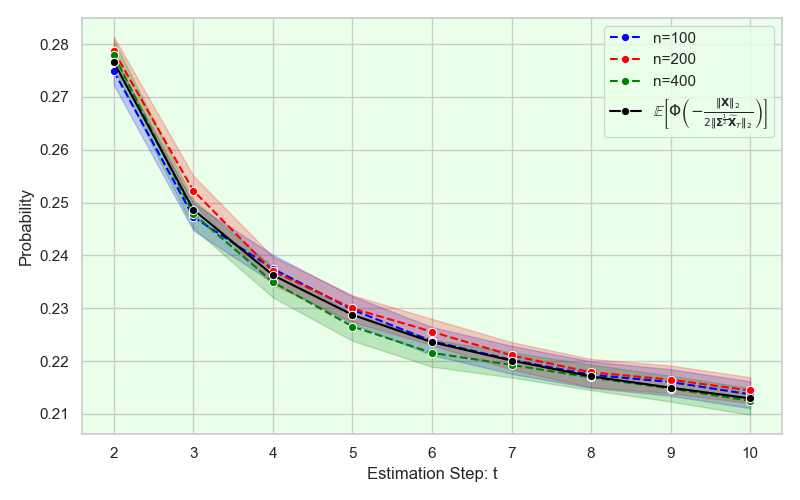}
    \caption{$c_t = t^{1.5}$}
  \end{subfigure}
\caption{Experimental results of Gaussian case for \textbf{Scenario 2} are presented, with 95\% confidence intervals plotted for each case.}
\label{Fig:Sce2}
\end{figure}

The results for \textbf{Scenario 2} are presented in Figure~\ref{Fig:Sce2}. The experimental findings are consistent with our theoretical analysis in two key aspects. First, Equation~(\ref{Prob_Better}) in Theorem~\ref{Thm:GaussianBetter} provides an exact expression for the probability of obtaining a better model during recursive training. As shown in Figure~\ref{Fig:Sce2}, the estimated values \( \widehat{P(T)} \) (dashed lines) align perfectly with the black solid line representing the true probability given by Equation~(\ref{Prob_Better}). Second, the probability \( P(T) \) is independent of the initial sample size \( n \), as evidenced by the near-complete overlap of the dashed lines corresponding to different values of \( n \).

\begin{figure}[ht!]
  \centering
  \begin{subfigure}[b]{0.31\textwidth}
    \includegraphics[width=\textwidth]{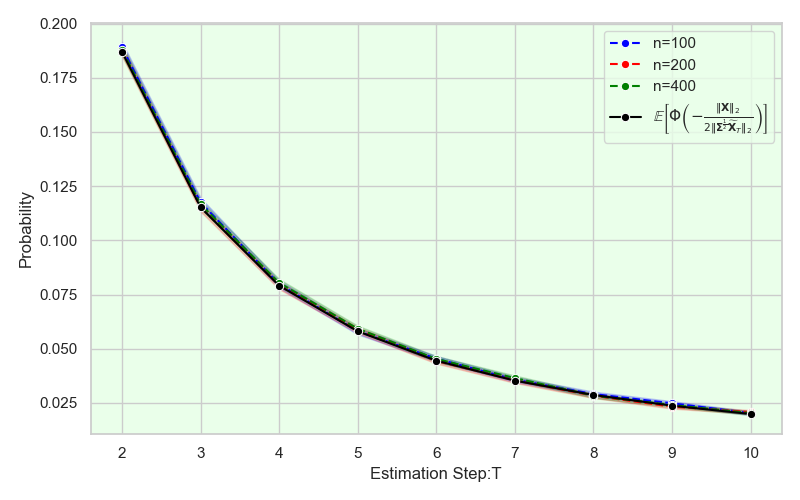}
   \caption{Exponential: $c_t = 1$}
  \end{subfigure}
  \begin{subfigure}[b]{0.31\textwidth}
    \includegraphics[width=\textwidth]{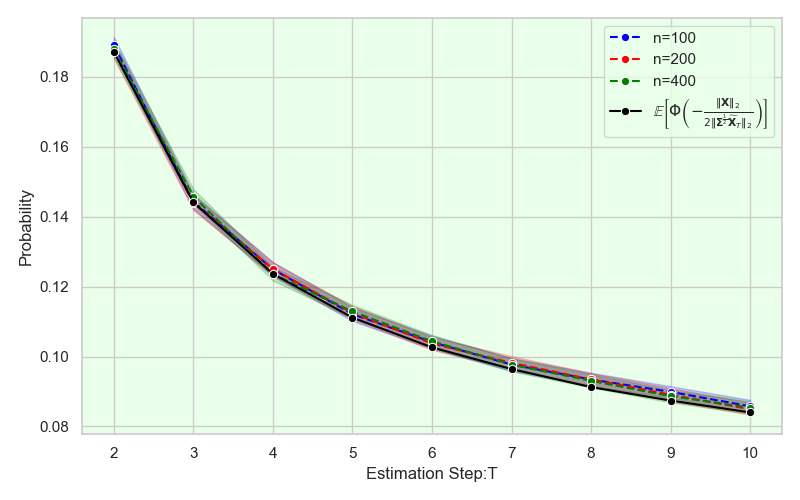}
    \caption{Exponential: $c_t = t$}
  \end{subfigure}
  \begin{subfigure}[b]{0.31\textwidth}
    \includegraphics[width=\textwidth]{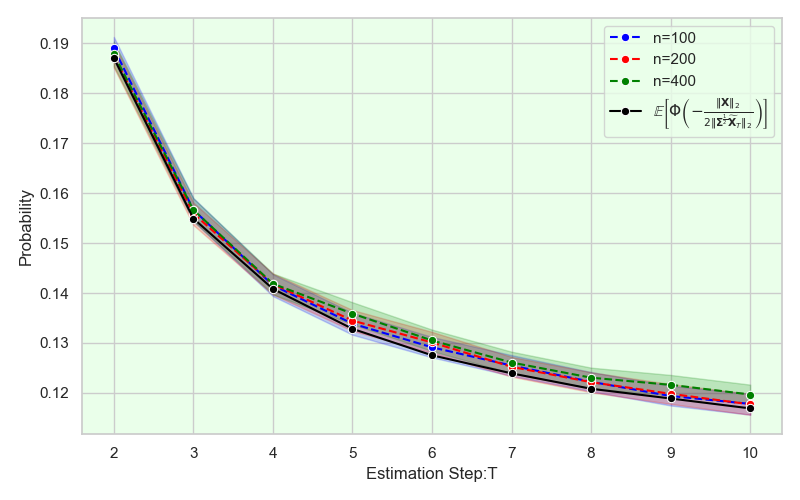}
    \caption{Exponential: $c_t = t^{1.5}$}
  \end{subfigure}
    \begin{subfigure}[b]{0.31\textwidth}
    \includegraphics[width=\textwidth]{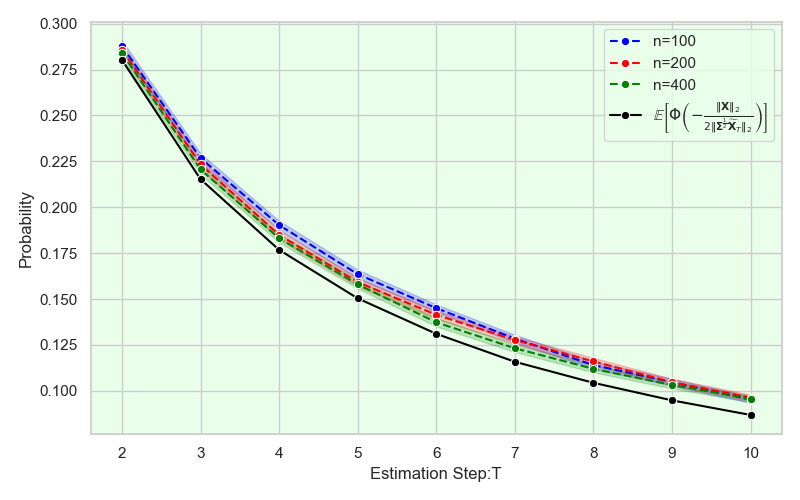}
   \caption{Logistic: $c_t = 1$}
  \end{subfigure}
  \begin{subfigure}[b]{0.31\textwidth}
    \includegraphics[width=\textwidth]{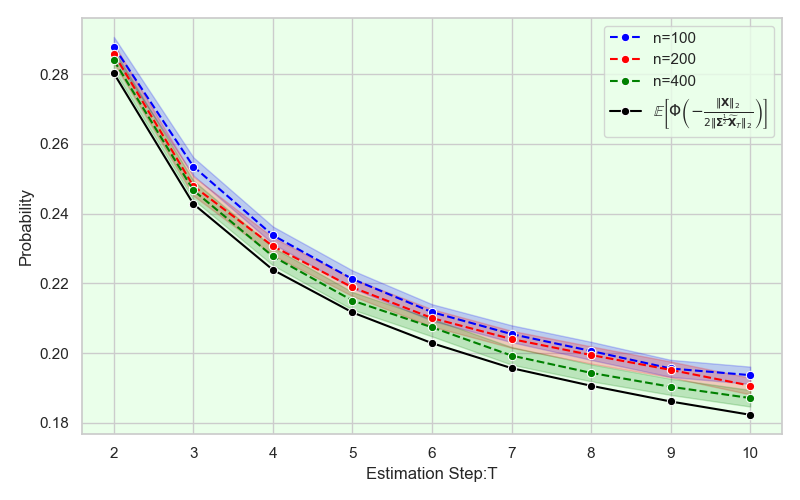}
    \caption{Logistic: $c_t = t$}
  \end{subfigure}
  \begin{subfigure}[b]{0.31\textwidth}
    \includegraphics[width=\textwidth]{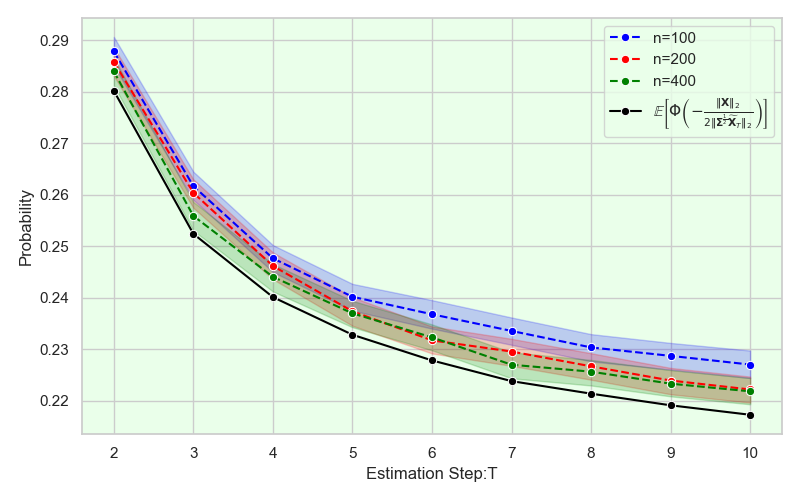}
    \caption{Logistic: $c_t = t^{1.5}$}
  \end{subfigure}
\caption{Experimental results for the non-Gaussian cases—Exponential distribution (top) and Logistic regression (bottom)—under \textbf{Scenario 2} are presented, with 95\% confidence intervals shown for each case.}
\label{Fig:Sce22}
\end{figure}

Next, we aim to demonstrate that Equation~(\ref{Prob_Better}) also provides a good approximation in non-Gaussian settings under Assumption~\ref{Ass:AssNorm}. In particular, we consider two representative examples: multivariate exponential distribution estimation and logistic regression. All other experimental configurations, including the initial sample generation and synthetic data expansion schemes, remain the same as in the Gaussian case. It is worth noting that in both cases, the calculation of Equation~(\ref{Prob_Better}) requires the computation of Fisher information matrices, which we approximate using Monte Carlo methods. The experimental results for these two cases are presented in Figure~\ref{Fig:Sce22}.

As shown in Figure~\ref{Fig:Sce22}, Equation~(\ref{Prob_Better}) also provides a reasonably accurate approximation of \( P(T) \) for non-Gaussian estimation procedures. These results support the validity of Theorem~\ref{Thm:General}. From these two examples, we observe that Equation~(\ref{Approx}) can be effectively used to analyze the probability of obtaining a better model in general recursive parametric training, provided that the assumption of asymptotic normality holds. In particular, for logistic regression, as the initial sample size \( n \) increases from 100 to 400, the estimated \( P(T) \) gradually approaches the black solid line, which represents the proposed estimate given by Equation~(\ref{Approx}).

\begin{figure}[ht!]
  \centering
  \begin{subfigure}[b]{0.31\textwidth}
    \includegraphics[width=\textwidth]{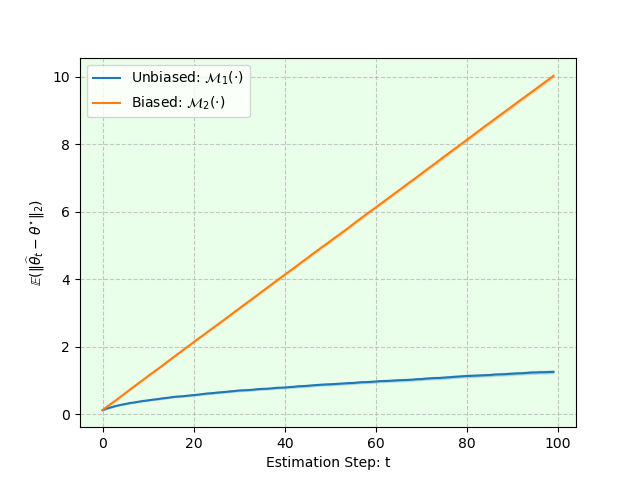}
  \end{subfigure}
  \begin{subfigure}[b]{0.31\textwidth}
    \includegraphics[width=\textwidth]{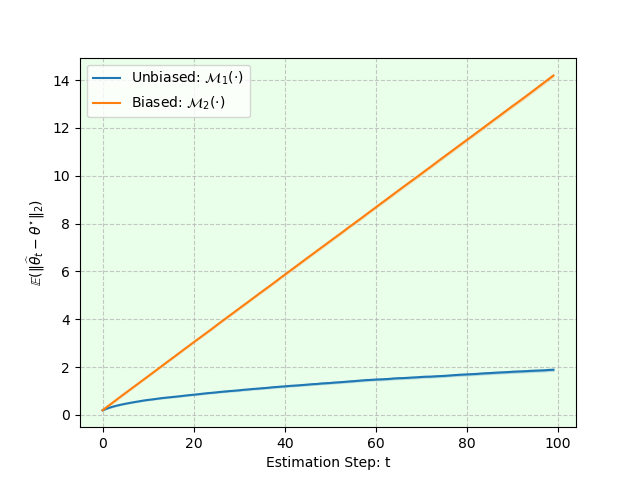}
  \end{subfigure}
  \begin{subfigure}[b]{0.31\textwidth}
    \includegraphics[width=\textwidth]{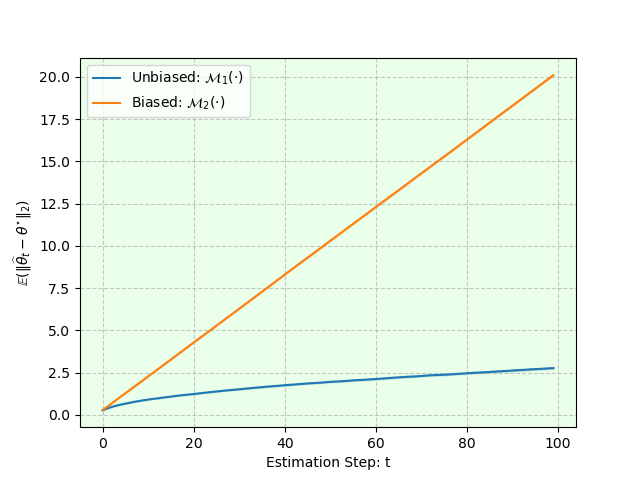}
  \end{subfigure}
  \begin{subfigure}[b]{0.31\textwidth}
    \includegraphics[width=\textwidth]{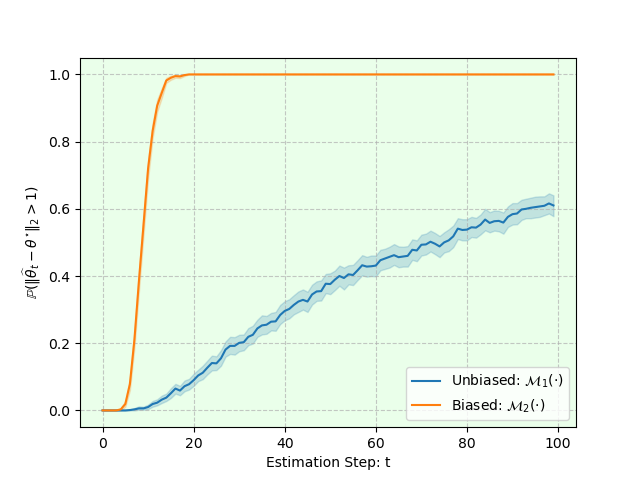}
  \end{subfigure}
  \begin{subfigure}[b]{0.31\textwidth}
    \includegraphics[width=\textwidth]{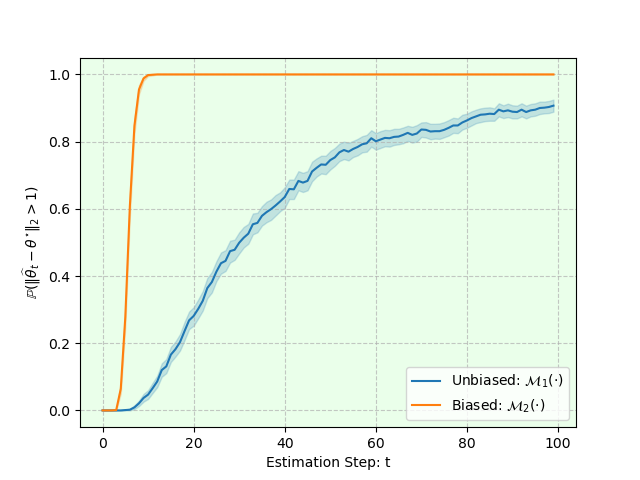}
  \end{subfigure}
  \begin{subfigure}[b]{0.31\textwidth}
    \includegraphics[width=\textwidth]{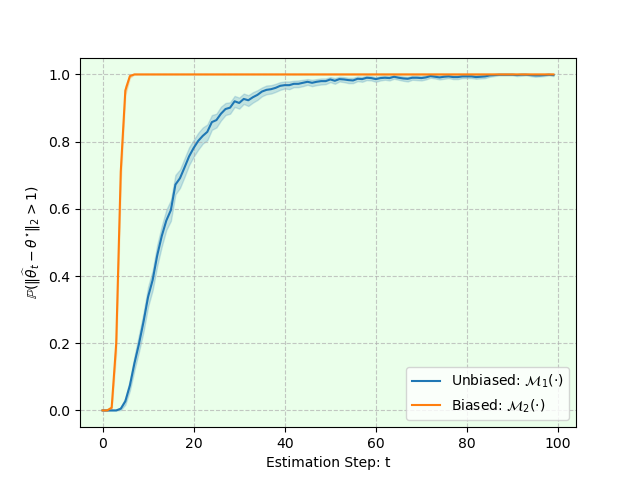}
  \end{subfigure}
  \begin{subfigure}[b]{0.31\textwidth}
    \includegraphics[width=\textwidth]{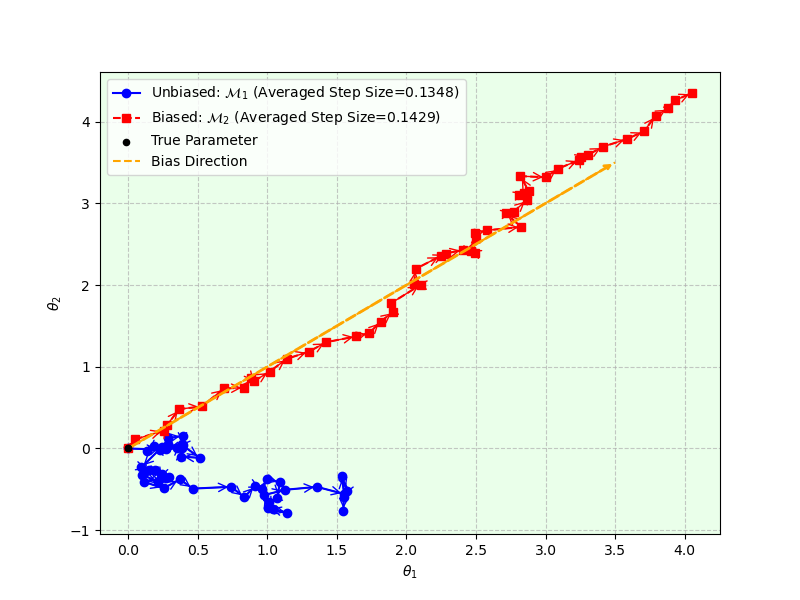}
  \end{subfigure}
  \begin{subfigure}[b]{0.31\textwidth}
    \includegraphics[width=\textwidth]{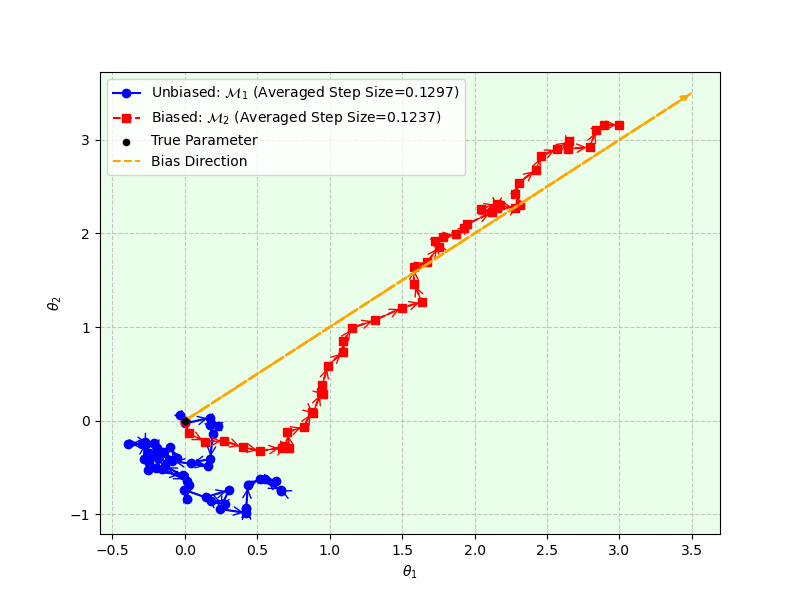}
  \end{subfigure}
  \begin{subfigure}[b]{0.31\textwidth}
    \includegraphics[width=\textwidth]{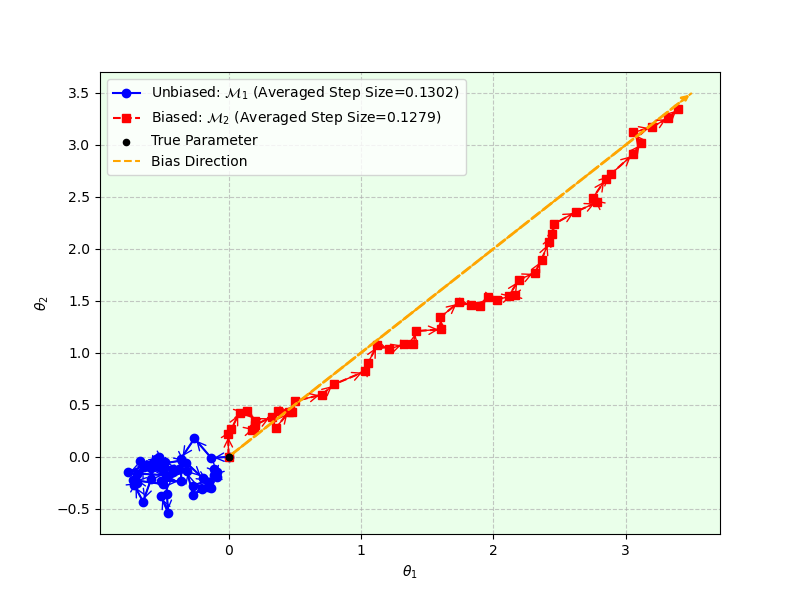}
  \end{subfigure}

\caption{The experimental results of \textbf{Scenario 3}: Displayed are the estimation error $\mathbb{E}(\Vert \widehat{\bm{\theta}}_t - \bm{\theta}^\star\Vert_2)$ (top row), the probability $\mathbb{P}(\Vert \widehat{\bm{\theta}}_t - \bm{\theta}^\star\Vert_2 \geq 1)$ (second row), and the parameter trajectories during recursive training for the unbiased estimator $\mathcal{M}_1$ and the biased estimators $\mathcal{M}_2$ (last row), under dimensionalities $p = 2$ (left), $p = 4$ (middle), and $p = 8$ (right).}
\label{Fig:Sce3}
\end{figure}

\noindent
\textbf{Scenario 3 (Large Bias Accelerates Model Collapse).} In this scenario, we intend to show that biased estimation leads to faster model collapse than unbiased estimation given that they have the same estimation error. Specifically, we consider a recursive Gaussian mean estimation with known covariance matrices. At the $t$-th step, we generate 200 samples from the last Gaussian $\mathcal{D}_{t} = \{\bm{x}_{t,i}\}_{i=1}^{200} \sim N(\widehat{\bm{\theta}}_{t}, \bm{I}_p)$. For the estimation of $\widehat{\bm{\theta}}_{t+1}$, we consider two estimation processes:
\begin{align*}
\widehat{\bm{\theta}}_{t+1} = \underbrace{\mathcal{M}_1(\mathcal{D}_t) = \frac{1}{100}\sum_{i=1}^{100}\bm{x}_{t,i}}_{\text{Unbiased Estimate }}\, \text{ and } \, \widehat{\bm{\theta}}_{t+1} = \underbrace{\mathcal{M}_2(\mathcal{D}_t) = \frac{1}{200}\sum_{i=1}^{200}\bm{x}_{t,i}+\frac{\bm{1}_p}{\sqrt{200}}}_{\text{Biased Estimate}}.
\end{align*}
Here, the unbiased estimate uses only the first 100 samples from $\mathcal{D}_t$ for estimation, while the biased estimate uses all available samples but introduces a bias term of $\frac{\bm{1}_p}{\sqrt{200}}$. It is worth noting that these two estimates have the same estimation errors at every step, that is
\begin{align*}
    \mathbb{E}\left[ \left(\mathcal{M}_1(\mathcal{D}_t)-\widehat{\bm{\theta}_t}\right)^2\Big|\widehat{\bm{\theta}_t}\right] = \frac{p}{100}  \mbox{ and } \mathbb{E}\left[ \left(\mathcal{M}_2(\mathcal{D}_t)-\widehat{\bm{\theta}_t}\right)^2\Big|\widehat{\bm{\theta}_t}\right] = \frac{p}{100}.
\end{align*}
We set $\bm{\theta}^\star = \bm{0}_p$ and consider cases where $p \in \{2, 4, 8\}$ with $T = 100$, replicating each case $10^4$ times. We report the average estimation error $\mathbb{E}(\Vert \widehat{\bm{\theta}}_t - \bm{\theta}^\star\Vert_2)$ and the probability $\mathbb{P}(\Vert \widehat{\bm{\theta}}_t - \bm{\theta}^\star\Vert_2 \geq 1)$ for each case in Figure \ref{Fig:Sce3}.

As illustrated in Figure~\ref{Fig:Sce3}, although the unbiased estimation procedure $\mathcal{M}_1(\cdot)$ and the biased procedure $\mathcal{M}_2(\cdot)$ yield similar estimation errors, the unbiased nature of $\mathcal{M}_1(\cdot)$ effectively mitigates model collapse in parameter estimation, in contrast to the biased counterpart. This observation is consistent with our theoretical findings in Theorem~\ref{Thm:positive_bias}, which show that biased estimation leads to a faster onset of model collapse. The underlying mechanism is further illustrated by the parameter trajectories during recursive training (last row of Figure~\ref{Fig:Sce3}). Clearly, under the biased estimation $\mathcal{M}_2$, the parameter trajectory tends to follow the direction from $(0,0)$ to $(1,1)$, causing $\widehat{\bm{\theta}}_T$ to drift away from $\bm{\theta}^\star$ more rapidly.

\subsection{Real Application: Recursive Generation of Tabular Data}
In this section, we validate our theoretical results using the House 16H dataset\footnote{Available at \url{https://www.openml.org/search?type=data&sort=runs&id=574}}. Constructed from the 1990 U.S. Census, this dataset contains aggregated demographic and housing statistics at the State-Place level across all U.S. states. It comprises 16 numerical covariates and one continuous response variable—the median house price—across 22,784 observations. First, we demonstrate that the synthetic data expansion scheme can effectively prevent model collapse by stabilizing the quality of synthetic data in tabular generative models. Second, we show that the theoretical results in Theorem \ref{Thm:General} are applicable to real-world datasets.

\textbf{Experiment 1.} In the first experiment, we adopt the Gaussian Copula method \citep{Gaussiancopula} as the generative model for tabular data, implemented via the Python package \textit{Synthetic Data Vault} \citep{patki2016synthetic}. To evaluate the quality of the synthetic data, we assess both distributional fidelity and covariance structure estimation. For fidelity, we compute the average Kolmogorov–Smirnov (KS) distance \citep{stephens1974edf} and the energy distance \citep{szekely2013energy} across all dimensions, quantifying the distributional differences between real and synthetic samples. For covariance estimation, we first fit a Gaussian copula model to the full real dataset, treating the resulting covariance matrix as the ground truth. At each round of synthetic data generation, we then compute the squared error between the covariance estimated from the synthetic data and the true covariance. We set the initial sample size to $n = 1000$ and consider two synthetic data generation schemes: $c_t = 1$ and $c_t = t^{1.1}$. Each setting is evaluated over $T = 25$ rounds and replicated 100 times.

\begin{figure}[htbp]
  \centering
  \begin{subfigure}[b]{0.31\textwidth}
    \includegraphics[width=\textwidth]{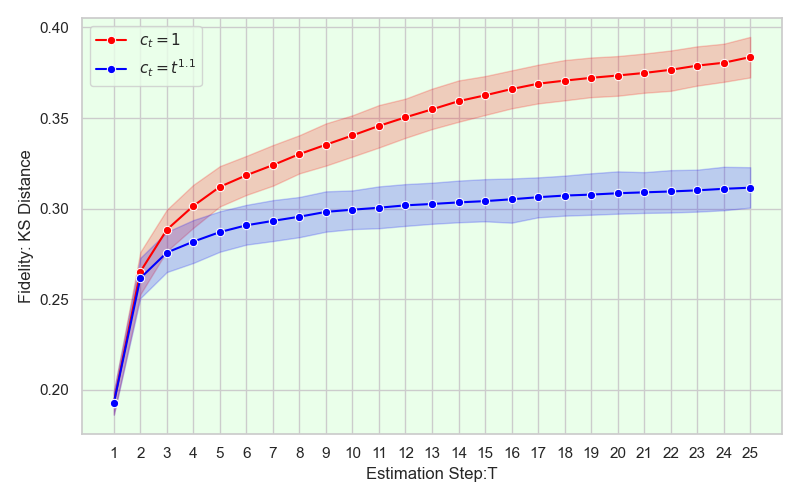}
   \caption{KS distance}
  \end{subfigure}
  \begin{subfigure}[b]{0.31\textwidth}
    \includegraphics[width=\textwidth]{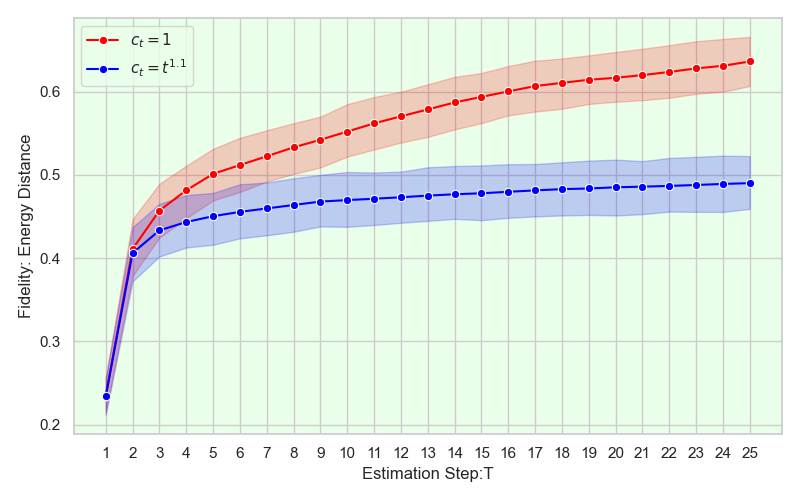}
    \caption{Energy Distance}
  \end{subfigure}
  \begin{subfigure}[b]{0.31\textwidth}
    \includegraphics[width=\textwidth]{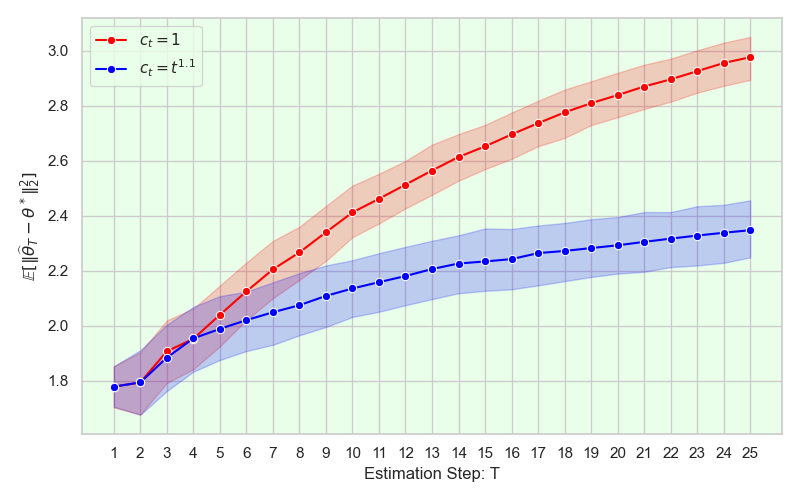}
    \caption{Parameter Estimation}
  \end{subfigure}
\caption{The results of \textbf{Experiment 1} evaluate the evolution of distributional differences throughout the recursive training process.
}
\label{Fig:Real}
\end{figure}

The results of \textbf{Experiment 1} are presented in Figure~\ref{Fig:Real}. Under the equal sample schedule ($c_t=1$), all fidelity metrics and parameter estimates degrade as $T$ increases. In contrast, with a suitable expansion scheme—specifically, the superlinear setting $c_t = t^{1.1}$—the metrics eventually stabilize, indicating that the degradation in synthetic data quality levels off and remains bounded.

\begin{figure}[htbp]
  \centering
  \begin{subfigure}[b]{0.31\textwidth}
    \includegraphics[width=\textwidth]{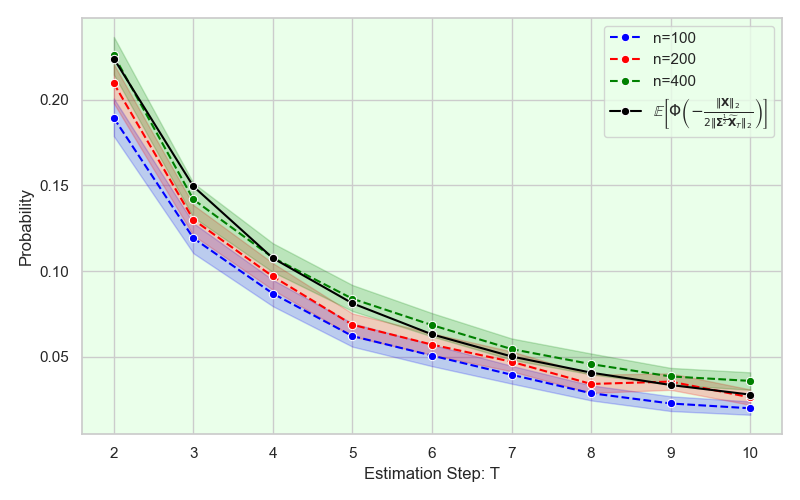}
   \caption{$c_t=1$}
  \end{subfigure}
  \begin{subfigure}[b]{0.31\textwidth}
    \includegraphics[width=\textwidth]{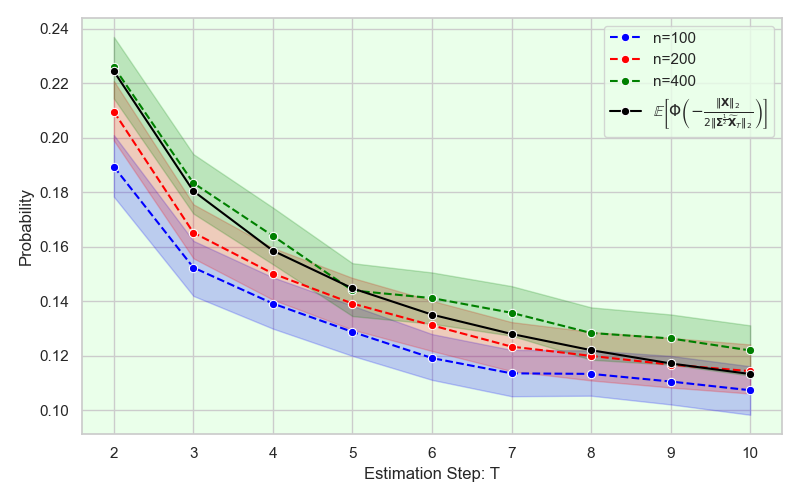}
    \caption{$c_t=t$}
  \end{subfigure}
  \begin{subfigure}[b]{0.31\textwidth}
    \includegraphics[width=\textwidth]{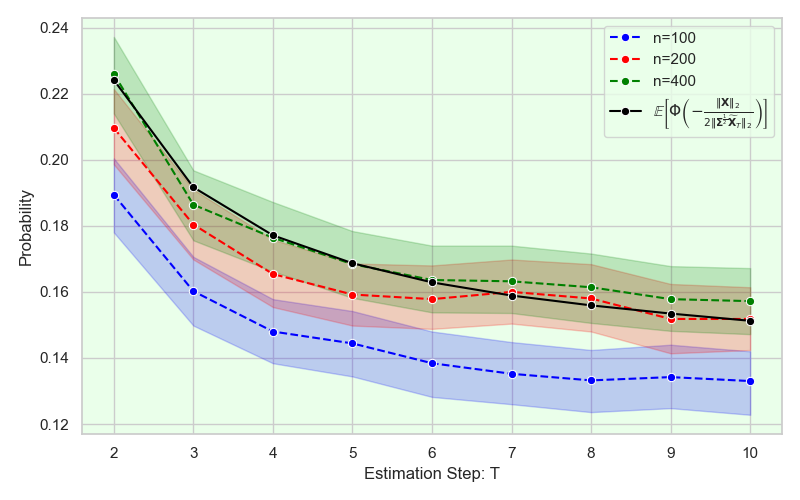}
    \caption{$c_t=t^{1.5}$}
  \end{subfigure}
\caption{The results of \textbf{Experiment 2} show how $P(T)$ evolves during the recursive training process under different synthetic data expansion schemes.}
\label{Fig:Real2}
\end{figure}

\textbf{Experiment 2.} In the second experiment, we aim to evaluate Theorem~\ref{Thm:General} on a real dataset under a regression framework. We consider the \textit{median house price} as the response variable and treat the remaining features as covariates. A linear regression model is fitted to the entire dataset, and the resulting coefficient vector is treated as \( \bm{\theta}^\star \). While the linear model may not correctly specify the underlying relationship between covariates and the response, it is well known that the ordinary least squares estimator remains asymptotically normal even in the presence of model misspecification~\citep{white1980heteroskedasticity}. Based on this, we estimate the asymptotic covariance matrix using nonparametric bootstrap. In each round of the recursive training process, we resample covariates from the real dataset and generate synthetic responses using the coefficient estimates from the previous round, adding independent standard normal noise. We repeat the 10-step recursive procedure 5{,}000 times for each of three synthetic data schemes and for initial sample sizes \( n \in \{100, 200, 400\} \). For each configuration, we estimate the probability \( P(T) \), and report the results in Figure~\ref{Fig:Real2}.

As shown in Figure~\ref{Fig:Real2}, the proposed approximation defined in (\ref{Approx}) (black solid line) closely matches the corresponding $P(T)$, particularly as the initial sample size $n$ increases from 100 to 400. This result demonstrates that the proposed method effectively characterizes the probability of obtaining a better model during the recursive training process.

\spacingset{1.5} 
\bibliography{ref}
\newpage

\appendix
\baselineskip=24pt
\setcounter{page}{1}
\setcounter{equation}{0}
\setcounter{section}{0}
\renewcommand{\thesection}{A.\arabic{section}}
\renewcommand{\thelemma}{A\arabic{lemma}}

\section*{\LARGE Appendix}
\addcontentsline{toc}{section}{Appendix} 
\spacingset{1.7} 

In this appendix, we present the proofs of all examples, lemmas, and theorems from the main text, along with several additional auxiliary lemmas. To facilitate the theoretical development, we first introduce the following notation:

\begin{enumerate}
\item We use the notation \( A_n \lesssim B_n \) to mean that there exists a constant \( C > 0 \) such that \( A_n \leq C B_n \), and \( A_n \gtrsim B_n \) to mean that there exists a constant \( C > 0 \) such that \( C A_n \geq B_n \).
\item We denote the Gamma function by $\Gamma(x) = \int_0^{\infty} t^{x-1} e^{-t}  dt$.
\item We denote the Riemann zeta function by $\zeta(s) = \sum_{n=1}^{\infty} \frac{1}{n^s}$ for $s > 1$.
\end{enumerate}

\section{Proof of Examples}
\noindent
\textbf{Proof of Example \ref{Exam:prop_recursive_variance}}: By the \textit{law of total variance}, we have  
\[
\mathrm{Var}(\widehat{\theta}_{t+1}) = \mathrm{Var}[\mathbb{E}(\widehat{\theta}_{t+1} \mid \widehat{\theta}_t)] + \mathbb{E}[\mathrm{Var}(\widehat{\theta}_{t+1} \mid \widehat{\theta}_t)].
\]
To evaluate the terms on the right-hand side, we note that, conditioned on \(\widehat{\theta}_t\), the estimator \(\widehat{\theta}_{t+1}\) is given by  
\[
\widehat{\theta}_{t+1} = \frac{1}{n_t} \sum_{i=1}^{n_t} x_{t,i},
\]
where \(\{x_{t,i}\}_{i=1}^{n_t}\) are independent normal random variables with mean \(\widehat{\theta}_t\) and variance 1. Thus, the conditional expectation follows as  
$
\mathbb{E}[\widehat{\theta}_{t+1} \mid \widehat{\theta}_t] = \widehat{\theta}_t,
$
which implies  
\[
\mathrm{Var}[\mathbb{E}(\widehat{\theta}_{t+1} \mid \widehat{\theta}_t)] = \mathrm{Var}(\widehat{\theta}_t).
\]
For the conditional variance, since each \(x_{t,i}\) is normally distributed with variance 1, it follows that $
\mathrm{Var}(\widehat{\theta}_{t+1} \mid \widehat{\theta}_t) = \frac{1}{n_t}$. Taking the expectation over \(\widehat{\theta}_t\), we obtain  
\[
\mathbb{E}[\mathrm{Var}(\widehat{\theta}_{t+1} \mid \widehat{\theta}_t)] = \frac{1}{n_t}.
\]
Substituting these results back into the law of total variance, we obtain  
\[
\mathrm{Var}(\widehat{\theta}_{t+1}) = \mathrm{Var}(\widehat{\theta}_t) + \frac{1}{n_t}.
\]
This completes the proof.  \hfill$\blacksquare$ \\

\noindent
\textbf{Proof of Example \ref{Exam:Gaus}:} Since $\bm{x}_i \sim N(\bm{\theta}, \bm{I})$ independently for $i \in [n]$, the sample mean is  
$$
\bar{\bm{x}} = \frac{1}{n} \sum_{i=1}^{n} \bm{x}_i.
$$
By the properties of the multivariate normal distribution, we have  
$$
\bar{\bm{x}} \sim N\left(\bm{\theta}, \frac{1}{n} \bm{I} \right).
$$
Multiplying both sides by $\sqrt{n}$, we obtain  
$$
\sqrt{n} (\bar{\bm{x}} - \bm{\theta}) \sim N(\bm{0}, \bm{I}).
$$

Next, consider the squared Euclidean norm:
$$
n \|\widehat{\bm{\theta}} - \bm{\theta} \|_2^2 = n \sum_{l=1}^{p} (\bar{x}_l - \theta_l)^2.
$$
Since each component $\sqrt{n} (\bar{x}_l - \theta_l)$ follows an independent standard normal distribution $N(0,1)$, the sum of their squares follows a chi-squared distribution with $p$ degrees of freedom:
$$
n \|\widehat{\bm{\theta}} - \bm{\theta} \|_2^2 \sim \chi^2(p).
$$
To obtain an upper bound for $\mathbb{P}\left(\Vert \widehat{\bm{\theta}} - \bm{\theta} \Vert_2 > \delta \right)$, it suffices to analyze the tail behavior of the chi-square distribution.
\begin{align*}
   \mathbb{P}\left(\Vert \widehat{\bm{\theta}} - \bm{\theta} \Vert_2 > \delta \right) =  
   \mathbb{P}\left(n\Vert \widehat{\bm{\theta}} - \bm{\theta} \Vert_2^2 > n\delta^2 \right) = 
   \mathbb{P}\left(\chi^2(p) > n\delta^2 \right).
\end{align*}
If $n\delta^2>p$, by Theorem 1 of \citet{ghosh2021exponential}, we have
\begin{align*}
    \mathbb{P}\left(\chi^2(p) > n\delta^2 \right) \leq  &
    \exp\left\{-\frac{p}{2}\left[\frac{n\delta^2}{p}-1-\log\left(\frac{n\delta^2}{p}\right)\right]\right\} \\ 
    \leq &
    \exp\left[-\frac{p}{2}\left(\frac{n\delta^2}{2p}-1\right)\right] = 
    \exp(p/2) \exp\left(-\frac{n\delta^2}{4}\right).
\end{align*}
where the second inequality follows from the fact that $x/2>\log(x)$ for any $x>0$. Note that when $n\delta^2 \leq p$, the above bound becomes $\mathbb{P}\left(\chi^2(p) > n\delta^2 \right) \leq \exp(p/4)$, which is still valid. Therefore, for any $\delta>0$, it holds that
\begin{align}
\label{Chi-Tail}
    \mathbb{P}\left(\chi^2(p) > n\delta^2 \right) \leq \exp(p/2) \exp\left(-\frac{n\delta^2}{4}\right).
\end{align}
Note that the above proof holds for any $\bm{\theta} \in \mathbb{R}^p$. Consequently, we obtain
\begin{align*}
 \sup_{\bm{\theta} \in \mathbb{R}^p}
      \mathbb{P}\left(\Vert \widehat{\bm{\theta}}-\bm{\theta}\Vert_2>\delta\right)  \leq \exp(p/2) \exp\left(-\frac{n\delta^2}{4}\right).
\end{align*}
This completes the proof.\hfill$\blacksquare$ \\

\noindent
\textbf{Proof of Example \ref{Exam:Gauss_subopt}}: Let \( x_1, x_2, \dots, x_n \) be independent random variables with mean \( \mathbb{E}[x_i] = \theta \) and variance \( \text{Var}(x_i) = 1 \). We define the estimator:
\[
\widehat{\theta} = \sum_{i=1}^{n} x_i w_i,
\]
where the weights are given by $ w_i = \frac{1/i}{H_n}$ and $H_n = \sum_{i=1}^n \frac{1}{i}$. We check that \( \widehat{\theta} \) is an unbiased estimator of \( \mu \) by computing its expectation
\[
\mathbb{E}[\widehat{\theta}] = \mathbb{E}\left[ \sum_{i=1}^{n} x_i w_i \right] = \sum_{i=1}^{n} \mathbb{E}[x_i] w_i.
\]
Since \( \mathbb{E}[x_i] = \theta \), we obtain $\mathbb{E}[\widehat{\theta}] = \mu \sum_{i=1}^{n} w_i = \theta \sum_{i=1}^{n} \frac{1/i}{H_n} = \theta$.

Since the \( x_i \) are independent, the variance of \( \widehat{\theta} \) is given by:
\[
\text{Var}(\widehat{\theta}) = \text{Var}\left( \sum_{i=1}^{n} x_i w_i \right) = \sum_{i=1}^{n} w_i^2 \text{Var}(x_i).
\]
Using \( \text{Var}(x_i) = 1 \), we obtain:
\[
\text{Var}(\widehat{\theta}) =  \sum_{i=1}^{n} w_i^2 =  \sum_{i=1}^{n} \left( \frac{1/i}{H_n} \right)^2 \leq \frac{\pi^2}{6 H_n^2}.
\]
where the last inequality follows from the fact that the series \( \sum_{i=1}^{n} \frac{1}{i^2} \) converges to \( \frac{\pi^2}{6} \) as $n$ approaches infinity. Since the harmonic number satisfies \( H_n \geq \log n  \) for any \( n \geq 1 \). Therefore, we have
\[
\text{Var}(\widehat{\theta}) \leq  \frac{1}{(\log n)^2} \cdot \frac{\pi^2}{6}.
\]
Using the sub-Gaussian tail bound, we have
\begin{align*}
    \mathbb{P}\left(
|\widehat{\theta}-\theta|>\delta
    \right) \leq 2\exp\left(-\frac{3 (\log n)^2\delta^2}{\pi^2}\right).
\end{align*}
The above result is independent of $\theta$, it then holds true that
$$
\sup_{\theta \in \mathbb{R}}\mathbb{P}\left(
|\widehat{\theta}-\theta|>\delta
    \right) \leq 2\exp\left(-\frac{3 (\log n)^2\delta^2}{\pi^2}\right).
$$
This completes the proof. 
\hfill$\blacksquare$ \\

\noindent
\textbf{Proof of Example \ref{Exam:Unif}}: We begin by noting that the estimate for \( \theta \) is given by \( \widehat{\theta} = \max_{i \in [n]} x_i \), the maximum of the sample values. 

First, observe that if \( \delta > M \), then for any \( \theta \in [1, M] \), we have
$$
\mathbb{P}\left(|\widehat{\theta} - \theta| > \delta\right) = 0.
$$
Thus, it suffices to consider \( \delta \in [0, M] \). Next, suppose that \( |\widehat{\theta} - \theta| > \delta \). This implies that \( \widehat{\theta} \) must be at most \( \theta - \delta \), i.e., 
$$
\max_{i \in [n]} x_i \leq \theta - \delta.
$$
The probability of this event can be computed as follows. The probability that each individual sample \( x_i \) is less than or equal to \( \theta - \delta \) is \( \frac{\theta - \delta}{\theta} \). Therefore, the probability that all \( n \) samples are less than or equal to \( \theta - \delta \) is given by
$$
\mathbb{P}\left( \max_{i \in [n]} x_i \leq \theta - \delta \right) = \left( \frac{\theta - \delta}{\theta} \right)^n.
$$
This can be simplified to
$$
\mathbb{P}\left( |\widehat{\theta} - \theta| > \delta \right) = \left( 1 - \frac{\delta}{\theta} \right)^n.
$$
Next, we bound this expression. Since \( \delta \leq M \), we have
$$
\left( 1 - \frac{\delta}{M} \right)^n = \left[ \left( 1 - \frac{\delta}{M} \right)^M \right]^{\frac{n}{M}}.
$$
Using the fact that \( \left( 1 - \frac{\delta}{M} \right)^M \leq \exp\left( -\delta \right) \) for any $\delta \in [0,M]$, we obtain
$$
\left( 1 - \frac{\delta}{M} \right)^n \leq \exp\left( -\frac{n\delta}{M} \right),
$$
Thus, we have
$$
\sup_{\theta \in [1, M]} \mathbb{P}\left( |\widehat{\theta} - \theta| > \delta \right) \leq \exp\left( -\frac{n\delta}{M} \right).
$$
This completes the proof. \hfill$\blacksquare$ \\

\noindent
\textbf{Proof of Example \ref{Exam:Gaussian}.} First, Example \ref{Exam:Gaus} shows that the Gaussian mean estimation process satisfies Assumption \ref{Ass:Uniform} with $r(n)=n$ and $\gamma=2$.

Next, we proceed to prove the main result. In Example \ref{Exam:Gaussian}, the recursive Gaussian estimation process can be formulated as
\begin{align*}
    \widehat{\bm{\theta}}_T = \widehat{\bm{\theta}}_{T-1} + 
    \frac{1}{n_{T-1}}\sum_{i=1}^{n_{T-1}}z_{T-1,i} = 
\bm{\theta}^\star + \sum_{t=0}^{T-1}\frac{1}{n_{t}}\sum_{i=1}^{n_{t}}z_{t,i},
\end{align*}
where $z_{t,i}$'s are independent standard normal samples. Clearly, we have
\begin{align*}
    \widehat{\bm{\theta}}_T - \bm{\theta}^\star \sim N(0,\bm V),
\end{align*}
where $V$ is computed as
\begin{align*}
   \bm V = \sum_{t=0}^{T-1} \frac{1}{n_{t}} \cdot \bm{I}_p = \frac{1}{n}\sum_{t=0}^{T-1} \frac{1}{c_t} \cdot \bm{I}_p \triangleq v_T \cdot \bm{I}_p.
\end{align*}
Clearly, the underlying distribution of $\Vert \widehat{\bm{\theta}}_T - \bm{\theta}^\star \Vert_2^2$ is
\begin{align*}
    \frac{1}{v_T}\Vert \widehat{\bm{\theta}}_T - \bm{\theta}^\star \Vert_2^2 \sim \chi^2(p),
\end{align*}
where $\chi^2(p)$ denotes the chi-squared distribution with $p$ degrees of freedom.

Applying the same argument as in (\ref{Chi-Tail}), we have
\begin{align*}
    &\mathbb{P}\left(
    \Vert \widehat{\bm{\theta}}_T - \bm{\theta}^\star \Vert_2 \geq \delta
    \right) =  \mathbb{P}\left(v_T^{-1}
    \Vert \widehat{\bm{\theta}}_T - \bm{\theta}^\star \Vert_2^2 \geq v_T^{-1}\delta^2
    \right) \\
    = &\mathbb{P}\left(\chi^2(p) \geq v_T^{-1}\delta^2
    \right) 
    \leq \exp(p/2)\exp\left(-\frac{n\delta^2}{4\sum_{t=0}^{T-1}\frac{1}{c_t}}\right).
\end{align*}
Clearly, as long as $\sum_{t=1}^{\infty}\frac{1}{c_t}<\infty$, we have
\begin{align*}
    \lim_{n\rightarrow \infty}\lim_{T\rightarrow \infty}
    \mathbb{P}\left(
    \Vert \widehat{\bm{\theta}}_T - \bm{\theta}^\star \Vert_2 \geq \delta
    \right)  = 0.
\end{align*}
To meet $\sum_{t=1}^{\infty}\frac{1}{c_t}<\infty$, $c_t$ can be chosen as $c_t \asymp t^{1+s}$ for any $s>0$. This completes the proof. \hfill$\blacksquare$ \\

\noindent
\textbf{Proof of Example \ref{Exam:Gaussian_unbound}.} We first show that this biased version of Gaussian mean estimation satisfies Assumption \ref{Ass:Uniform} with $r(n)=n$ and $\delta^\gamma = \delta^2$. For any $t \geq 2$, conditional on $\widehat{\bm{\theta}}_{t-1}$, it holds true that
\begin{align*}
   \sqrt{c_{t-1} \cdot n}\left( \widehat{\bm{\theta}}_t - \widehat{\bm{\theta}}_{t-1} \right)\sim N\left( \bm{1}_p, \bm{I}_p \right).
\end{align*}
Suppose $c_t \geq 1$. Conditional on $\widehat{\bm{\theta}}_{t-1}$, we have
\begin{align*}
    n_{t-1}\Vert \widehat{\bm{\theta}}_t - \widehat{\bm{\theta}}_{t-1}\Vert_2^2 \sim 
    \chi^2_p(p),
\end{align*}
where $\chi^2_{p_1}(p_2)$ denotes the non-central chi-squared distribution with $p_1$ degrees of freedom and a noncentrality parameter of $p_2$ \citep{patnaik1949non}.
\begin{align*}
   & \mathbb{P}\left(\Vert \widehat{\bm{\theta}}_t - \widehat{\bm{\theta}}_{t-1} \Vert_2 \geq \delta\Big|\widehat{\bm{\theta}}_{t-1}\right) = 
        \mathbb{P}\left(n_{t-1}\Vert \widehat{\bm{\theta}}_t - \widehat{\bm{\theta}}_{t-1} \Vert_2^2 \geq n_{t-1}\delta^2\Big|\widehat{\bm{\theta}}_{t-1}\right)  \\
        = &
        \mathbb{P}\left(n_{t-1}\Vert \widehat{\bm{\theta}}_t - \widehat{\bm{\theta}}_{t-1} \Vert_2^2-2p \geq n_{t-1}\delta^2-2p\Big|\widehat{\bm{\theta}}_{t-1}\right).
\end{align*}
Applying Theorem 7 of \citet{zhang2020non} yields that 
\begin{itemize}
    \item[(1)] For $2p<n_{t-1}\delta^2 \leq 5p$,
\begin{align*}
  \mathbb{P}\left(\Vert \widehat{\bm{\theta}}_t - \widehat{\bm{\theta}}_{t-1} \Vert_2 \geq \delta\Big|\widehat{\bm{\theta}}_{t-1}\right) \leq 
  \exp\left(
-\frac{C[n_t\delta^2-2p]^2}{3p}
  \right) ,
\end{align*}
for some positive constant $C>0$.
\item[(2)] For $n_t\delta^2 > 5p$,
\begin{align*}
  \mathbb{P}\left(\Vert \widehat{\bm{\theta}}_t - \widehat{\bm{\theta}}_{t-1} \Vert_2 \geq \delta\Big|\widehat{\bm{\theta}}_{t-1}\right) \leq 
  \exp(-C(n_t\delta^2-2p)) = \exp(2Cp)\cdot\exp(-Cn_t\delta^2).
\end{align*}
\end{itemize}

Clearly, as $n_t$ increases, $n_t\delta^2 >5 p$ is more likely to hold. Therefore, there exists a large integer $N_0$ such that (2) holds for any $n_t \geq N_0$. Hence, we can find a large constant $C_0$ such that 
\begin{align*}
  \mathbb{P}\left(\Vert \widehat{\bm{\theta}}_t - \widehat{\bm{\theta}}_{t-1} \Vert_2 \geq \delta\Big|\widehat{\bm{\theta}}_{t-1}\right) \leq 
  \exp(-C(n_t\delta^2-2p)) = C_0\cdot\exp(-Cn_t\delta^2),
\end{align*}
for any $n_t \geq 1$. Note that the above upper bound holds regardless of the value of $\widehat{\bm{\theta}}_{t-1}$. Therefore, this setting satisfies the assumption of Corollary~\ref{corollary_main_thm} with $(\kappa, \gamma) = (1, 2)$. 

\textbf{Proof of Main Result.} Next, we intend to show that $c_t = t^{2 + s}$ for some $s > 0$ is a sufficient condition for ensuring
$$
\lim_{n\rightarrow \infty}
\lim_{T\rightarrow \infty}
    \mathbb{P}\left(
    \Vert \widehat{\bm{\theta}}_T-\bm{\theta}^\star\Vert_2 \geq \delta 
    \right) = 0.
$$
However, when $c_t = t^{1 + s}$ for some $s \in (0,1)$, we have
$$
\lim_{n\rightarrow \infty}
\lim_{T\rightarrow \infty}
    \mathbb{P}\left(
    \Vert \widehat{\bm{\theta}}_T-\bm{\theta}^\star\Vert_2 \geq \delta 
    \right) = 1.
$$
Using the same technique as in Example \ref{Exam:Gaussian}, we have
\begin{align*}
    \widehat{\bm{\theta}}_T =\bm{\theta}^\star+ \sum_{t=0}^{T-1}\frac{1}{\sqrt{n_t}} + 
    \sum_{t=0}^{T-1}\frac{1}{c_t \cdot n} \sum_{i=1}^{n_t}\bm{z}_{t,i},
\end{align*}
where $c_0=1$ and $\bm{z}_{t,i}\sim N(\bm{0}_p,\bm{I}_p)$. Clearly, $\text{Var}(\widehat{\bm{\theta}}_T) = \bm{I}_p \cdot \sum_{t=0}^{T-1}\frac{1}{c_t \cdot n} =\bm{I}_p \cdot v_T$ with $v_T = \sum_{t=0}^{T-1}\frac{1}{c_t \cdot n}$. Therefore, we have
\begin{align*}
    v_T^{-\frac{1}{2}}(\widehat{\bm{\theta}}_T-\bm{\theta}^\star) \sim N(\bm{1}_p \cdot r_T,\bm{I}_p) \text{ with } r_T \triangleq  
    \frac{\sum_{t=0}^{T-1}\frac{1}{\sqrt{c_t}}}{\sqrt{\sum_{t=0}^{T-1}\frac{1}{c_t}}}=\frac{1+\sum_{t=1}^{T-1}\frac{1}{\sqrt{c_t}}}{\sqrt{1+\sum_{t=1}^{T-1}\frac{1}{c_t}}}.
\end{align*}
Next, we analyze the behavior of $r_T$ under different patterns of $c_t$. Suppose $c_t = t^{a}$ for $a>1$,
\begin{align*}
    \lim_{T\rightarrow \infty}r_T = 
    \begin{cases}
        \frac{1+\zeta(a/2)}{\sqrt{1+\zeta(a)}}<\infty,&\text{ if } a>2, \\
        \infty, &\text{ if } 1<a \leq 2.
    \end{cases}
\end{align*}
where $\zeta(a)=\sum_{t=1}^{\infty} t^{-a}$ is the Riemann zeta function.

Next, we can reformulate $\mathbb{P}\left(
    \Vert \widehat{\bm{\theta}}_T-\bm{\theta}^\star\Vert_2 \geq \delta 
    \right)$ as
\begin{align*}
    \mathbb{P}\left(
    \Vert \widehat{\bm{\theta}}_T-\bm{\theta}^\star\Vert_2 \geq \delta 
    \right) = 
    \mathbb{P}\left(
    v_T^{-1}\Vert \widehat{\bm{\theta}}_T-\bm{\theta}^\star\Vert_2^2 - p(1+r_T^2) \geq \delta^2 v_T^{-1} - p(1+r_T^2)
    \right).
\end{align*}
Applying Theorem 7 of \citet{zhang2020non} yields the following results.
\begin{itemize}
    \item[(1)] If $p(1+r_T^2)v_T \leq \delta^2  \leq p(2+3r_T^2)v_T$, we have
    $$
    \mathbb{P}\left(
    \Vert \widehat{\bm{\theta}}_T-\bm{\theta}^\star\Vert_2 \geq \delta 
    \right) \leq  \exp\left(-\frac{C(\delta^2 v_T^{-1} - p(1+r_T^2))^2}{p(1+2r_T^2)}\right).
    $$
    \item[(2)] If $p(2+3r_T^2)v_T \leq \delta^2$, we have
    $$
    \mathbb{P}\left(
    \Vert \widehat{\bm{\theta}}_T-\bm{\theta}^\star\Vert_2 \geq \delta 
    \right) \leq  \exp\left(-C\delta^2 v_T^{-1}\right).
    $$
\end{itemize}

\noindent
\textbf{Case 1: When $a>2$.} Clearly, if $c_t = t^a$ with $a>2$, we have $\lim_{T\rightarrow \infty} \sum_{t=0}^{T-1} c_T^{-1} <\infty$, which then implies that
\begin{align*}
    \lim_{T\rightarrow \infty} p(1+r_T^2)v_T<\infty \mbox{ and }
    \lim_{n\rightarrow \infty}\lim_{T\rightarrow \infty} p(1+r_T^2)v_T = 0.
\end{align*}
This implies that for any $\delta_0>0$, there exists a large $n_0$ such that $ p(1+r_T^2)v_T<\delta^2$ for any $n \geq n_0$. Therefore, for any $\delta>0$, we have
\begin{align*}
\lim_{n\rightarrow \infty}
\lim_{T\rightarrow \infty}
    \mathbb{P}\left(
    \Vert \widehat{\bm{\theta}}_T-\bm{\theta}^\star\Vert_2 \geq \delta 
    \right) = 0,
\end{align*}
under the pattern $c_t = t^{2+s}$ for any $s>0$.

\noindent
\textbf{Case 2: When $1<a \leq 2$.} If $c_t = t^{a}$ with $1<a \leq 2$, we have
\begin{align*}
    \lim_{T\rightarrow \infty}\sum_{t=1}^{\infty} \frac{1}{\sqrt{c_t}} = \infty \mbox{ and }\lim_{T\rightarrow \infty}\sum_{t=1}^{\infty} \frac{1}{c_t}<\infty.
\end{align*}
With this, $\mathbb{P}\left(
    \Vert \widehat{\bm{\theta}}_T-\bm{\theta}^\star\Vert_2 \geq \delta 
    \right)$ can be lower bounded as
\begin{align*}
    \mathbb{P}\left(
    \Vert \widehat{\bm{\theta}}_T-\bm{\theta}^\star\Vert_2 \geq \delta 
    \right) \geq\mathbb{P}\left(
    |\widehat{\theta}_{T,i}-\theta^{\star}_i|\geq \delta 
    \right) \geq \mathbb{P}\left(
    \widehat{\theta}_{T,i}-\theta^{\star}_i\geq \delta 
    \right),
\end{align*}
where $\widehat{\theta}_{T,i}$ represents the $i$-th element of $\widehat{\bm{\theta}}_T$. Note that the mean is $\mathbb{E}(\widehat{\theta}_{t,i}|\widehat{\theta}_{t-1}) =\theta^{\star}_i + \sum_{t=0}^{T-1}\frac{1}{\sqrt{nc_t}}$, which diverges to infinity as $T\rightarrow \infty$. Further, the variance of $\widehat{\theta}_{T,i}$ is $\sum_{t=0}^{T-1}\frac{1}{n_t}<\infty$ even if $T\rightarrow \infty$. Therefore, we have
\begin{align*}
   \mathbb{P}\left(
    \Vert \widehat{\bm{\theta}}_T-\bm{\theta}^\star\Vert_2 \geq \delta 
    \right) \geq 1 - \Phi\left(\frac{\delta- \sum_{t=0}^{T-1}\frac{1}{\sqrt{n_t}}}{ \sqrt{\sum_{t=0}^{T-1}\frac{1}{n_t }}}\right).
\end{align*}
Note that for $c_t = t^a$ with $1<a \leq 2$,
$$
\lim_{T\rightarrow \infty}\frac{\delta- \sum_{t=0}^{T-1}\frac{1}{\sqrt{n_t}}}{ \sqrt{\sum_{t=0}^{T-1}\frac{1}{n_t }}} = -\infty.
$$
Therefore, if $c_t = t^a$ with $1<a\leq 2$, for any $n$ and $\delta>0$, we have
\begin{align*}
    \lim_{T\rightarrow \infty} \mathbb{P}\left(
    \Vert \widehat{\bm{\theta}}_T-\bm{\theta}^\star\Vert_2 \geq \delta 
    \right) = 1.
\end{align*}
This completes the proof. \hfill$\blacksquare$ \\

\noindent\textbf{Proof of Example \ref{Exam:Exponential}.} Let $X_1, X_2, \dots, X_n$ be i.i.d. random variables from an exponential distribution with rate parameter $\theta > 0$. The likelihood function is:
\[
L(\theta) = \prod_{i=1}^n \theta e^{-\theta X_i} = \theta^n \exp\left(-\theta \sum_{i=1}^n X_i \right)
\]
The log-likelihood function is
\[
\ell(\theta) = n \log \theta - \theta \sum_{i=1}^n X_i.
\]
Taking the derivative with respect to $\theta$ and setting it to zero
\[
\frac{d\ell}{d\theta} = \frac{n}{\theta} - \sum_{i=1}^n X_i = 0
\quad \Rightarrow \quad
\hat{\theta}_{\text{MLE}} = \frac{n}{\sum_{i=1}^n X_i} = \frac{1}{\bar{X}}.
\]
We compute the expectation of $\hat{\theta}_{\text{MLE}} = 1 / \bar{X}$. Since $\bar{X} = \frac{1}{n} \sum_{i=1}^n X_i$, and $\sum_{i=1}^n X_i$ follows a Gamma distribution $\sum_{i=1}^n X_i \sim \text{Gamma}(n, \theta)
$, where the Gamma distribution is parameterized by shape $n$ and rate $\theta$.

Let $Y = \sum_{i=1}^n X_i$, then:
\[
\hat{\theta}_{\text{MLE}} = \frac{n}{Y}
\quad \Rightarrow \quad
\mathbb{E}[\hat{\theta}_{\text{MLE}}] = n \cdot \mathbb{E}\left[ \frac{1}{Y} \right].
\]
It is known that for $Y \sim \text{Gamma}(n, \theta)$ with $n > 1$, the expectation of $1/Y$ is:
\[
\mathbb{E}\left[\frac{1}{Y}\right] = \frac{\theta}{n - 1}
\]
Therefore, $\mathbb{E}[\hat{\theta}_{\text{MLE}}] = n \cdot \frac{\theta}{n - 1} = \frac{n}{n - 1} \theta$. The bias is defined as
\[
\text{Bias}(\hat{\theta}_{\text{MLE}}) = \mathbb{E}[\hat{\theta}_{\text{MLE}}] - \theta = \left( \frac{n}{n - 1} - 1 \right) \theta = \frac{\theta}{n - 1}.
\]
Since $\theta \in [1, M]$, we have
\[
\frac{\theta}{n-1} \leq \frac{M}{n-1} \asymp \frac{2M}{n}, \quad \text{for } n \geq 2,
\]
which implies that the bias scales as $O(1/n)$ and thus satisfies Assumption~\ref{Ass:biased} with $\rho = 1$. \hfill$\blacksquare$

\section{Proof of Lemmas}

\noindent
\textbf{Proof of Lemma \ref{Lemma:SpeedBound}.} First, by Lemma \ref{lemma:varBound}, we have
    \begin{align*}
        \mathbb{E}[|\widehat{\theta}_i - \theta_i|^2] \leq \frac{2 C_1}{\gamma} \cdot \Gamma\left( \frac{2}{\gamma} \right) \cdot C_2^{-2/\gamma} n^{-\frac{2\kappa}{\gamma}},
    \end{align*}
    for any $i \in [p]$, where $\widehat{\theta}_i$ denotes the $i$-th element of $\widehat{\bm{\theta}}$. Then we can decompose $\mathbb{E}[|\widehat{\theta}_i - \theta_i|^2]$ as
    \begin{align*}
        \mathbb{E}[|\widehat{\theta}_i - \theta_i|^2] = 
        \mathbb{E}[|\widehat{\theta}_i -\mathbb{E}(\widehat{\theta}_i)+\mathbb{E}(\widehat{\theta}_i)- \theta_i|^2] \geq 
        \left(\mathbb{E}(\widehat{\theta}_i)- \theta_i\right)^2.
    \end{align*}
    This then implies that
    \begin{align*}
       \left(\mathbb{E}(\widehat{\theta}_i)- \theta_i\right)^2 \leq  
       \frac{2 C_1}{\gamma} \cdot \Gamma\left( \frac{2}{\gamma} \right) \cdot C_2^{-2/\gamma} n^{-\frac{2\kappa}{\gamma}}.
    \end{align*}
    By Assumption \ref{Ass:biased}, we have
    \begin{align*}
        \left(\mathbb{E}(\widehat{\theta}_i)- \theta_i\right)^2 \asymp
        \frac{1}{n^{2\rho}} \lesssim \frac{2 C_1}{\gamma} \cdot \Gamma\left( \frac{2}{\gamma} \right) \cdot C_2^{-2/\gamma} n^{-\frac{2\kappa}{\gamma}}.
    \end{align*}
    This then implies $\rho \geq \kappa/\gamma$. This completes the proof. \hfill $\blacksquare$

\begin{lemma}
\label{Lemma:Gamma}
    Suppose $X$ follows the Gamma$(\alpha,\beta)$ with $\alpha>0$ and $\beta>0$ being shape and rate parameters, respectively. Then we have
    \begin{align*}
      &  \mathbb{E}(\log X) =  \psi(\alpha) + \log \beta, \\
      &  \mathbb{E}(\log^2 X) = (\log \beta)^2 + 2 \log \beta \psi(\alpha) + \psi_1(\alpha) + (\psi(\alpha))^2, \\
      & \mathrm{Var}(\log X) = \psi_1(\alpha).
    \end{align*}
    where $\psi(\alpha)$ and $\psi_1(\alpha)$ are defined as follows:
    \begin{align*}
        \psi(\alpha) = \frac{d}{d\alpha} \log \Gamma(\alpha)=\frac{\Gamma'(\alpha)}{\Gamma(\alpha)} \text{ and } 
        \psi_1(\alpha) = \frac{d}{d\alpha}\psi(\alpha).
    \end{align*}
    where $\Gamma(\alpha) = \int_0^\infty x^{\alpha - 1} e^{-x}  dx$
\end{lemma}

\noindent
\textbf{Proof of Lemma \ref{Lemma:Gamma}.} If $X$ follows the Gamma$(\alpha,\beta)$, then density function of $X$ is given as
\[
f_X(x) = \frac{x^{\alpha-1} e^{-x/\beta}}{\beta^{\alpha} \Gamma(\alpha)}\text{ for } x > 0.
\]
Then, we get
\begin{align}
\label{Eqn1Lemma1}
 \mathbb{E}[\log X] = \int_0^\infty \log(x) \frac{x^{\alpha-1} e^{-x/\beta}}{\beta^\alpha \Gamma(\alpha)}  dx= \frac{1}{\beta^\alpha \Gamma(\alpha)} \int_0^\infty \log(x) x^{\alpha-1} e^{-x/\beta}  dx.   
\end{align}
Perform the substitution \( u = \frac{x}{\beta} \), so that \( x = \beta u \) and \( dx = \beta du \). Substituting these into (\ref{Eqn1Lemma1}) gives
\begin{align}
\label{Expect_logX}
  \mathbb{E}[\log X] = \frac{1}{\beta^{\alpha-1} \Gamma(\alpha)}
\int_0^\infty \log(\beta u) (\beta u)^{\alpha-1} e^{-u} du \triangleq 
 \frac{1}{\beta^{\alpha-1} \Gamma(\alpha)} \cdot I_1.  
\end{align}
Using the fact that \( \log(\beta u) = \log(\beta) + \log(u) \), $I_1$ becomes
\begin{align}
\label{I_1_Bound}
I_1 = &
\int_0^\infty \log(\beta) (\beta u)^{\alpha-1} e^{-u}  du + \int_0^\infty \log(u) (\beta u)^{\alpha-1} e^{-u}  du  \notag \\
   = & \log(\beta) \beta^{\alpha-1} \Gamma(\alpha)+\beta^{\alpha-1} \int_0^\infty \log(u) u^{\alpha-1} e^{-u} du \notag \\ 
   =&\log(\beta) \beta^{\alpha-1} \Gamma(\alpha)+
   \beta^{\alpha-1} \Gamma(\alpha) \psi(\alpha),
\end{align}
where the last equality follows from the fact that
\[
\frac{d}{d\alpha} \Gamma(\alpha) = \frac{d}{d\alpha} \left( \int_0^\infty x^{\alpha-1} e^{-x}  dx \right) = \int_0^\infty \frac{\partial}{\partial \alpha} \left( x^{\alpha-1} e^{-x} \right) dx = \int_0^\infty  x^{\alpha-1} e^{-x} \log(x)dx.
\]
Plugging (\ref{I_1_Bound}) into (\ref{Expect_logX}), we have
\begin{align*}
    \mathbb{E}[\log X] = \log (\beta) + \psi(\alpha).
\end{align*}
This completes the proof for $\mathbb{E}[\log X]$.

Next, we proceed to compute \( \mathbb{E}(\log^2 X) \). Substituting the PDF of the Gamma distribution into the expectation yields that
\[
\mathbb{E}(\log^2 X) = \int_0^\infty \log^2(x) \frac{x^{\alpha - 1} e^{-x/\beta}}{\Gamma(\alpha) \beta^\alpha} dx.
\]
Using the same substitution \( y = \frac{x}{\beta} \), so that \( x = \beta y \) and \( dx = \beta \, dy \), we get
\[
\mathbb{E}(\log^2 X) = \frac{1}{\Gamma(\alpha) \beta^\alpha} \int_0^\infty \log^2(\beta y) (\beta y)^{\alpha - 1} e^{-y} \beta  dy.
\]
Simplifying the powers of \( \beta \) and the logarithm term gives
\[
\mathbb{E}(\log^2 X) = \frac{1}{\Gamma(\alpha)} \int_0^\infty \left[ (\log \beta)^2 + 2 \log \beta \log y + (\log y)^2 \right] y^{\alpha - 1} e^{-y}  dy.
\]
Now split the integral into three parts:
\[
\mathbb{E}(\log^2 X) = \frac{1}{\Gamma(\alpha)} \left[ (\log \beta)^2 \int_0^\infty y^{\alpha - 1} e^{-y} \, dy + 2 \log \beta \int_0^\infty \log y \, y^{\alpha - 1} e^{-y} \, dy + \int_0^\infty (\log y)^2 y^{\alpha - 1} e^{-y}  dy \right].
\]
Using the same arguments as above, we have
\[
\mathbb{E}(\log^2 X) = (\log \beta)^2 + 2 \log \beta \psi(\alpha) + \psi_1(\alpha) + (\psi(\alpha))^2,
\]
where $\psi_1(\alpha)=\psi'(\alpha)$ is the derivative of $\psi(\alpha)$. This completes the proof. \hfill$\blacksquare$ \\

\begin{lemma}
\label{Lemma:Improve}
Let $\bm{Z}_1, \bm{Z}_2, \dots, \bm{Z}_T$ be independent \(p\)-dimensional normal vectors with $\bm{Z}_i \sim N(\bm{0},\bm{\Sigma}_i)$. We first define $\bm{S}_T = \bm{Z}_1+\bm{Z}_2+\cdots+\bm{Z}_T$ and $\bm{V}_T = \bm{Z}_2+\bm{Z}_3+\cdots+\bm{Z}_T$. Then, it holds true that
\[
\mathbb{P}\Bigl(\|\bm{Z}_1\|_2> \|\bm{S}_T\|_2\Bigr)=\mathbb{E}_{\bm{V}_T}\left[\Phi\left(-\frac{\|\bm{V}_T\|_2}{2\Vert \bm{\Sigma}_1^{\frac{1}{2}} \widetilde{\bm{V}}_T \Vert_2}\right)\right],
\]
where $\widetilde{\bm{V}}_T =\bm{V}_T/\Vert \bm{V}_T\Vert_2$ and $\Phi(x) = \int_{-\infty}^x \frac{1}{\sqrt{2\pi}} e^{-\frac{t^2}{2}}  dt$ denotes the cumulative distribution function (CDF) of the standard normal distribution.
\end{lemma}

\noindent
\textbf{Proof of Lemma \ref{Lemma:Improve}.} 
Since $\bm{S}_T = \bm{Z}_1 + \bm{V}_T$, we have
\[
\|\bm{S}_T\|_2^2 = \|\bm{Z}_1 + \bm{V}_T\|_2^2 = \|\bm{Z}_1\|_2^2 + 2\langle \bm{Z}_1, \bm{V}_T\rangle + \|\bm{V}_T\|_2^2.
\]
The event $\Bigl\{\|\bm{Z}_1\|_2> \|\bm{S}_T\|_2\Bigr\}$ is equivalent to
\[
\|\bm{Z}_1\|_2^2 > \|\bm{Z}_1\|_2^2+ 2\langle \bm{Z}_1, \bm{V}_T\rangle + \|\bm{V}_T\|_2^2,
\]
which reduces to 
$$
2\langle \bm{Z}_1, \bm{V}_T\rangle + \|\bm{V}_T\|_2^2 < 0
\Leftrightarrow 2\langle \bm{\Sigma}_1^{-\frac{1}{2}}\bm{Z}_1, \bm{\Sigma}_1^{\frac{1}{2}}\bm{V}_T\rangle + \|\bm{V}_T\|_2^2 < 0.
$$.Note that $\bm{V}_T \sim N(\bm{0},\bm{\Sigma}_{2:T})$, where $\bm{\Sigma}_{2:T} = \sum_{i=2}^T \bm{\Sigma}_i$. For a fixed \(\bm{V}_T\), we write
\[
\bm{\Sigma}_1^{\frac{1}{2}}\bm{V}_T = \| \bm{\Sigma}_1^{\frac{1}{2}}\bm{V}_T\|_2 \cdot \bm{u},
\]
where \(\bm{u}\) is a unit vector in \(\mathbb{R}^p\). Then the inner product $\langle \bm{\Sigma}_1^{-\frac{1}{2}} \bm{Z}_1, \bm{u}\rangle$ is distributed as \(N(0,1)\). Denote
\[
X \triangleq \langle \bm{\Sigma}_1^{-\frac{1}{2}}\bm{Z}_1, \bm{u}\rangle \sim N(0,1).
\]
Then, it follows that
\[
\langle \bm{\Sigma}_1^{-\frac{1}{2}}\bm{Z}_1, \bm{\Sigma}_1^{\frac{1}{2}}\bm{V}_T\rangle = \langle \bm{\Sigma}_1^{-\frac{1}{2}}\bm{Z}_1, \|\bm{\Sigma}_1^{\frac{1}{2}}\bm{V}_T\|_2 \cdot \bm{u} \rangle = \| \bm{\Sigma}_1^{\frac{1}{2}}\bm{V}_T\|_2 \cdot \langle \bm{\Sigma}_1^{-\frac{1}{2}}\bm{Z}_1, \bm{u}\rangle = \|\bm{\Sigma}_1^{\frac{1}{2}}\bm{V}_T\|_2 \cdot X.
\] Hence, the inequality becomes
\[
2\|\bm{\Sigma}_1^{\frac{1}{2}}\bm{V}_T\|_2 \cdot  X+\|\bm{V}_T\|_2^2 < 0 \quad \Longleftrightarrow \quad X < -\frac{\|\bm{V}_T\|_2}{2\Vert \bm{\Sigma}_1^{\frac{1}{2}} \bm{V}_T/\|\bm{V}_T\|_2 \Vert_2}.
\]
Thus, conditional on \(\bm{V}_T\), we have
\[
\mathbb{P}\Bigl(\|\bm{Z}_1\|_2> \|\bm{S}_T\|_2\mid \bm{V}_T\Bigr) = \mathbb{P}\left(X< -\frac{\|\bm{V}_T\|_2}{2\Vert \bm{\Sigma}_1^{\frac{1}{2}} \bm{V}_T/\|\bm{V}_T\|_2 \Vert_2}\right)
=\Phi\left(-\frac{\|\bm{V}_T\|_2}{2\Vert \bm{\Sigma}_1^{\frac{1}{2}} \bm{V}_T/\|\bm{V}_T\|_2 \Vert_2}\right).
\]
Taking expectation with respect to \(\bm{V}_T\) yields
\[
\mathbb{P}\Bigl(\|\bm{Z}_1\|_2> \|\bm{S}_T\|_2\Bigr) = \mathbb{E}_{\bm{V}_T}\left[\Phi\left(-\frac{\|\bm{V}_T\|_2}{2\Vert \bm{\Sigma}_1^{\frac{1}{2}} \widetilde{\bm{V}}_T \Vert_2}\right)\right],
\]
where $\widetilde{\bm{V}}_T =\bm{V}_T/\Vert \bm{V}_T\Vert_2$. This completes the proof. \hfill$\blacksquare$ \\

\begin{lemma}
\label{lemma:varBound}
    Suppose that \( \widehat{\bm{\theta}} = \mathcal{M}(\mathcal{D}) \) is an estimate of \( \bm{\theta} \) with $\mathcal{D}=\{\bm{x}_i\}_{i=1}^n \sim \mathbb{P}_{\bm{\theta}}$ satisfying Assumption \ref{Ass:Uniform} with $r(n)=n^\kappa$, it then follows that
    \begin{align*}
        \mathbb{E}[|\widehat{\theta}_i - \theta_i|^p] \leq \frac{p C_1}{\gamma} \cdot \Gamma\left( \frac{p}{\gamma} \right) \cdot C_2^{-p/\gamma} n^{-\frac{p\kappa}{\gamma}},
    \end{align*}
    for any $i \in [p]$, where $\widehat{\theta}_i$ denotes the $i$-th element of $\widehat{\bm{\theta}}$.
\end{lemma}

\noindent
\textbf{Proof of Lemma \ref{lemma:varBound}.}  First, under Assumption \ref{Ass:Uniform} and the assumption that $r(n)=n^{\kappa}$, we have
\[
\mathbb{P}\left( | \widehat{\theta}_i - \theta_i| \geq \delta \right) \leq 
\mathbb{P}\left( \| \widehat{\bm{\theta}} - \bm{\theta}\|_2 \geq \delta \right) \leq C_1 \exp(-C_2 n^{\kappa} \delta^\gamma).
\]

\paragraph{Step 1: Use the tail-sum formula.}
We consider the $p$-th central moment:
\[
\mathbb{E}[|\widehat{\theta}_i - \theta_i|^p] 
= \int_0^\infty \mathbb{P}\left(|\widehat{\theta}_i - \theta_i|^p > t\right) dt 
= \int_0^\infty \mathbb{P}\left(|\widehat{\theta}_i - \theta_i| > t^{1/p}\right) dt.
\]

\paragraph{Step 2: Apply the tail bound.}
Using the given tail inequality, we get:
\[
\mathbb{E}[|\widehat{\theta}_i - \theta_i|^p] \leq \int_0^\infty C_1 \exp(-C_2 n^{\kappa} t^{\gamma/p}) dt.
\]

\paragraph{Step 3: Change of variables.}
Let \( \delta = C_2 n^{\kappa} t^{\gamma/p} \), then \( t = (C_2 n^{\kappa})^{-p/\gamma} \delta^{p/\gamma} \), and
\[
dt = \frac{p}{\gamma} (C_2 n^{\kappa})^{-p/\gamma} \delta^{\frac{p}{\gamma} - 1} d\delta.
\]
Hence,
\[
\mathbb{E}[|\widehat{\theta}_i - \theta_i|^p] \leq \frac{p C_1}{\gamma} (C_2 n^{\kappa})^{-p/\gamma} \int_0^\infty \delta^{\frac{p}{\gamma} - 1} e^{-\delta} d\delta.
\]

\paragraph{Final bound.}
This yields the moment bound:
\[
\mathbb{E}[|\widehat{\theta}_i - \theta_i|^p] \leq C'(p) \cdot n^{-\kappa \cdot \frac{p}{\gamma}},
\]
where 
\[
C'(p) = \frac{p C_1}{\gamma} \cdot \Gamma\left( \frac{p}{\gamma} \right) \cdot C_2^{-p/\gamma}.
\]
This completes the proof. \hfill$\blacksquare$ \\

\section{Proof of Theorems}

\noindent
\textbf{Proof of Theorem \ref{Exam:Probvari}:} This proof consists of two parts: (1) Proof of Diverging Population Risk and (2) Proof of Vanishing Diversity.

\textbf{1 (Diverging Population Risk).} We first prove that the population risk of $\widehat{\sigma}_t^2$ diverges as $T$ increases. Note that $\mathbb{E}\left[(\widehat{\sigma}_t^2-\sigma^2)^2\right]$ can be rewritten as
\begin{align}
\label{Eqn_ini}
    \mathbb{E}\left[(\widehat{\sigma}_t^2-\sigma^2)^2\right] = 
    \mathbb{E}\left[(\widehat{\sigma}_t^2-\widehat{\sigma}_{t-1}^2)^2\right]+2\mathbb{E}\left[(\widehat{\sigma}_t^2-\widehat{\sigma}_{t-1}^2)(\widehat{\sigma}_{t-1}^2-\sigma^2)\right]+
    \mathbb{E}\left[(\widehat{\sigma}_{t-1}^2-\sigma^2)^2\right].
\end{align}
Let $\mathcal{F}_{t-1}$ denote the $\sigma$-algebra generated by all events associated with datasets up to the $(t-1)$-th training step. Then, we have
\begin{align}
\label{Eqn1}
    &\mathbb{E}\left[(\widehat{\sigma}_t^2-\widehat{\sigma}_{t-1}^2)^2\right] =  \mathbb{E} \left\{\mathbb{E}\left[(\widehat{\sigma}_t^2-\widehat{\sigma}_{t-1}^2)^2 \mid \mathcal{F}_{t-1}\right] \right\} \notag \\
    = & \mathbb{E} \left\{\mathbb{E}\left[\widehat{\sigma}_{t-1}^4\left(\frac{\widehat{\sigma}_t^2}{\widehat{\sigma}_{t-1}^2}-1\right)^2 \Big| \mathcal{F}_{t-1}\right] \right\} = \mathbb{E} \left\{\frac{\widehat{\sigma}_{t-1}^4}{n^2}\mathbb{E}\left[\left(\frac{n\widehat{\sigma}_t^2}{\widehat{\sigma}_{t-1}^2}-n\right)^2 \Big| \mathcal{F}_{t-1}\right] \right\}.
\end{align}
Since $\frac{n\widehat{\sigma}_t^2}{\widehat{\sigma}_{t-1}^2}$ follows a $\chi^2_n$ distribution conditional on $\mathcal{F}_{t-1}$, (\ref{Eqn1}) can be rewritten as
\begin{align}
    \label{Eqn_First}
    \mathbb{E}\left[(\widehat{\sigma}_t^2-\widehat{\sigma}_{t-1}^2)^2\right] = 
\mathbb{E} \left[\frac{2\widehat{\sigma}_{t-1}^4}{n} \right].
\end{align}
For the second term of (\ref{Eqn_ini}), we have 
\begin{align}
\label{Eqn_2}
    \mathbb{E}\left[(\widehat{\sigma}_t^2-\sigma_{t-1}^2)(\widehat{\sigma}_{t-1}^2-\sigma^2)\right] = \mathbb{E}\left\{\mathbb{E}\left[(\widehat{\sigma}_{t-1}^2-\sigma^2)
    (\widehat{\sigma}_t^2-\widehat{\sigma}_{t-1}^2)\mid \mathcal{F}_{t-1}\right] \right\}=0.
\end{align}
Then, plugging (\ref{Eqn_First}) and (\ref{Eqn_2}) into (\ref{Eqn_ini}), we have
\begin{align}
\label{Temp_Eq}
    \mathbb{E}\left[(\widehat{\sigma}_t^2-\sigma^2)^2\right] = \mathbb{E} \left[\frac{2\widehat{\sigma}_{t-1}^4}{n} \right] + \mathbb{E}\left[(\widehat{\sigma}_{t-1}^2-\sigma^2)^2\right],
\end{align}
Note that for any $t \geq 1$, we have $\mathbb{E}(\widehat{\sigma}_t^2)=\mathbb{E}\{\mathbb{E}[\widehat{\sigma}_{t}^2|\mathcal{F}_{t-1}]\} = \mathbb{E}(\widehat{\sigma}_{t-1}^2)$. Therefore, it follows that
\begin{align*}
\mathbb{E}(\widehat{\sigma}_t^2)=\sigma^2, \text{ for any } t \geq 1.
\end{align*}
With this, we can show that 
\begin{align}
\label{Temp_Eq2}
\mathbb{E}(\widehat{\sigma}_{t-1}^4)=\mathbb{E}\left[(\sigma_{t-1}^2-\sigma^2)^2\right]+\sigma^4.
\end{align}
Plugging (\ref{Temp_Eq2}) into \eqref{Temp_Eq} yields that
\begin{align}
\mathbb{E}\left[(\widehat{\sigma}_t^2-\sigma^2)^2\right]=\left(1+\frac{2}{n}\right)\mathbb{E}\left[(\sigma_{t-1}^2-\sigma^2)^2\right]+\frac{2}{n}\sigma^4,
\end{align}
which further implies that
\begin{align*}
\frac{\mathbb{E}[(\widehat{\sigma}_t^2-\sigma^2)^2]+\sigma^4}{\mathbb{E}[(\sigma_{t-1}^2-\sigma^2)^2]+\sigma^4}=1+\frac{2}{n}.
\end{align*}
Note that $\mathbb{E}[(\sigma_{1}^2-\sigma^2)^2] = \frac{2\sigma^2}{n}$. It then follows that
\begin{align*}
\mathbb{E}\left[(\widehat{\sigma}_T^2-\sigma^2)^2\right]=\left[\left(1+\frac{2}{n}\right)^{T}-1\right]\sigma^4.
\end{align*}
Finally, for any $n \geq 2$, it holds that 
\begin{align*}
\lim_{T \rightarrow \infty} \mathbb{E}[(\widehat{\sigma}_T^2-\sigma^2)^2] = \lim_{T \rightarrow \infty}
    \left[\left(1+\frac{2}{n}\right)^{T}-1\right]\sigma^4 =\infty.
\end{align*}
This completes the proof of first part. \\

\textbf{2 (Vanishing Diversity).} Next, we aim to show that $\widehat{\sigma}_{T}$ converges to 1 in probability as $T$ approaches infinity. Note that $\widehat{\sigma}_t^2$ can be written as
\begin{align*}
    \widehat{\sigma}_t^2 = \frac{1}{n}\sum_{i=1}^n (x_{t-1,i}-\mu)^2 =
    \underbrace{\frac{1}{n}\sum_{i=1}^n \frac{(x_{t-1,i}-\mu)^2}{\widehat{\sigma}_{t-1}^2}}_{\text{Normalized } \chi^2(n)} \cdot \widehat{\sigma}_{t-1}^2.
\end{align*}
Note that regardless of the value of \( \widehat{\sigma}_{t-1}^2 \), the quantity \( \frac{1}{n} \sum_{i=1}^{n} \frac{(x_{t-1,i} - \mu)^2}{\widehat{\sigma}_{t-1}^2} \) follows the same distribution. Consequently, \( \frac{1}{n} \sum_{i=1}^{n} \frac{(x_{t-1,i} - \mu)^2}{\widehat{\sigma}_{t-1}^2} \) is independent of \( \widehat{\sigma}_{t-1}^2 \). Therefore, we can express $\widehat{\sigma}_t^2$ as
\begin{align*}
    \widehat{\sigma}_t^2 = \prod_{l=0}^{t-1} \frac{1}{n}\sum_{i=1}^n \frac{(x_{t-1,i}-\mu)^2}{\widehat{\sigma}_{t-1}^2} = \prod_{l=1}^{t}C_l,
\end{align*}
where \( C_l = \frac{1}{n} \sum_{i=1}^n \frac{(x_{l-1,i} - \mu)^2}{\widehat{\sigma}_{l-1}^2} \), and clearly, the \( C_l \)'s are independent, normalized \( \chi^2 \) random variables with \( n \) degrees of freedom.

Note that $\mathbb{E}(C_l)=1$ for any $l \geq 1$. Therefore, for $T \geq 1$, 
\begin{align}
\label{Main_Equ}
    \mathbb{P}(\widehat{\sigma}_T^2>\epsilon) =  &
    \mathbb{P}(\log(\widehat{\sigma}_T^2)>\log(\epsilon))=
    \mathbb{P}\left(\sum_{l=1}^T \log C_l>\log(\epsilon)\right) \notag \\
    = &
    \mathbb{P}\left(\sum_{l=1}^T \log C_l - T\mathbb{E}(\log C_l)>\log(\epsilon)-T\mathbb{E}(\log C_l)\right) \notag \\
    = &
    \mathbb{P}\left(\frac{1}{T}\sum_{l=1}^T \log C_l - \mathbb{E}(\log C_l)>\frac{\log(\epsilon)}{T}-\mathbb{E}(\log C_l)\right).
\end{align}
where \( \mathbb{E}(\log C_l) \) is negative, as derived from Jensen's inequality $\mathbb{E}(\log C_l) < \log(\mathbb{E}(C_l))=0$. 

Next, we aim to compute \( \mathbb{E}(\log C_l) \).  
By utilizing the relationship between the Gamma and Chi-square distributions, we deduce that \( C_l \) follows a Gamma distribution with shape parameter \( \frac{n}{2} \) and scale parameter \( \frac{2}{n} \), i.e.,  
$$
C_l \sim \text{Gamma}\left( \frac{n}{2}, \frac{2}{n} \right), \text{for } 1 \leq l \leq T
$$  
where \( \frac{n}{2} \) is the shape parameter and \( \frac{2}{n} \) is the scale parameter. By Lemma \ref{Lemma:Gamma}, $-\mathbb{E}(\log C_l)$ can be calculated as
\begin{align*}
   - \mathbb{E}[\log C_l] = -\psi\left(\frac{n}{2}\right)-\log\left(\frac{2}{n}\right) =-\psi\left(\frac{n}{2}\right)+\log\left(\frac{n}{2}\right),
\end{align*}
where \( \psi(\alpha) \) is the digamma function, which has the following polynomial representation \citep{bernardo1976psi}
\begin{align*}
    \psi(\alpha) = \frac{\int_{0}^{+\infty} \log(x) x^{\alpha-1} e^{-x}dx}{\int_{0}^{+\infty}  x^{\alpha-1} e^{-x}dx} = \log \alpha -\frac{1}{2\alpha}-\frac{1}{12\alpha^2}+\frac{1}{120\alpha^4}-
    O\left(\frac{1}{\alpha^6}\right).
\end{align*}
It can be verified that when $n \geq 1$, we have
\begin{align*}
    -\psi\left(\frac{n}{2}\right)+\log\left(\frac{n}{2}\right) = \frac{1}{n}+\frac{1}{3n^2}-\frac{2}{15n^4}-O\left(\frac{1}{n^6}\right) \geq \frac{1}{3n}.
\end{align*}
With this, (\ref{Main_Equ}) can be further bounded as
\begin{align*}
    \mathbb{P}(\widehat{\sigma}_T^2>\epsilon)=
    &\mathbb{P}\left(\frac{1}{T}\sum_{l=1}^T \log C_l - \mathbb{E}(\log C_l)>\frac{\log(\epsilon)}{T}-\mathbb{E}(\log C_l)\right) \\
    \leq & 
    \mathbb{P}\left(\frac{1}{T}\sum_{l=1}^T \log C_l - \mathbb{E}(\log C_l)>\frac{\log(\epsilon)}{T} + \frac{1}{3n}\right).
\end{align*}
If \( \frac{\log(\epsilon)}{T} + \frac{1}{3n} > 0 \), i.e., \( \epsilon > e^{-\frac{T}{3n}} \), then by Chebyshev's inequality, we have
\begin{align*}
  \mathbb{P}(\widehat{\sigma}_T^2>\epsilon)  \leq \frac{\text{Var}[\log C_l]}{T\left(\frac{\log(\epsilon)}{T} + \frac{1}{3n}\right)^2} = \frac{\psi_1(\frac{n}{2})}{T\left(\frac{\log(\epsilon)}{T} + \frac{1}{3n}\right)^2} \leq 
    \frac{6}{nT\left(\frac{\log(\epsilon)}{T} + \frac{1}{3n}\right)^2}.
\end{align*}
where the last equality follows from Lemma \ref{Lemma:Gamma}. Here $\psi_1(x)$ is the derivative of $\psi(\alpha)$ and the last inequality follows from the fact that 
\begin{align*}
    \psi_1\left(\frac{n}{2}\right) = \frac{2}{n} + \frac{2}{n^2} +\frac{4}{3n^3} -O\left(\frac{1}{n^5}\right)  \leq \frac{6}{n}.
\end{align*}
To sum up, for any \( n \geq 2 \) and \( T \geq 1 \), it holds for any \( \epsilon > e^{-\frac{T}{3n}} \) that
\begin{align*}
    \mathbb{P}(\widehat{\sigma}_T^2 \leq \epsilon) \geq 1 - \frac{6}{nT\left(\frac{\log(\epsilon)}{T} + \frac{1}{3n}\right)^2}.
\end{align*}

Note that for any $n \geq 2$ and $\epsilon>0$, there exists some large $T_0$ such that $\epsilon>e^{-\frac{T}{3n}}$ for any $T \geq T_0$. Therefore, for any $\epsilon>0$ and $n \geq 2$, we have
\begin{align*}
    \lim_{T\rightarrow \infty}\mathbb{P}(\widehat{\sigma}_T^2 \leq \epsilon) = 1.
\end{align*}
This completes the proof of Theorem \ref{Exam:Probvari}. \hfill$\blacksquare$ \\

\noindent 
\textbf{Proof of Theorem \ref{Thm:Main}:} Denote the event $\mathcal{E}_T(\delta)$ as
\begin{align*}
   \mathcal{E}_T(\delta) =  \{\Vert \widehat{\bm{\theta}}_{T} - \bm{\theta}^\star\Vert_2 \leq \delta \}
\end{align*}
Due to the triangle inequality, we have
\begin{align*}
    \Vert \widehat{\bm{\theta}}_{T} - \bm{\theta}^\star\Vert_2
    \leq \sum_{t=1}^{T}
    \Vert \widehat{\bm{\theta}}_{t} - \widehat{\bm{\theta}}_{t-1}\Vert_2
\end{align*}
where $\bm{\theta}_0 = \bm{\theta}^\star$. 
Let $\{\delta_t\}_{t=1}^{T}$ be a sequence such that $\sum_{t=1}^{T}\delta_t \leq \delta$. Then, we have
\begin{align*}
\mathbb{P}\left( \bigcap_{t=1}^{T}\{\Vert \widehat{\bm{\theta}}_{t} - \widehat{\bm{\theta}}_{t-1}\Vert_2 \leq \delta_t\} \right) \leq 
    \mathbb{P}\left(\Vert \widehat{\bm{\theta}}_{T} - \bm{\theta}^\star\Vert_2 \leq \delta \right).
\end{align*}
To obtain a lower bound for $\mathbb{P}\left(\Vert \widehat{\bm{\theta}}_{T} - \bm{\theta}^\star\Vert_2 \leq \delta \right)$, it suffices to provide a lower bound for $\mathbb{P}\left( \bigcap_{t=1}^{T}\{\Vert \widehat{\bm{\theta}}_{t} - \widehat{\bm{\theta}}_{t-1}\Vert_2 \leq \delta_t\} \right)$. Using the De Morgan's law, we have
\begin{align*}
    \mathbb{P}\left( \bigcap_{t=1}^{T}\{\Vert \widehat{\bm{\theta}}_{t} - \widehat{\bm{\theta}}_{t-1}\Vert_2 \leq \delta_t\} \right) =  &
    1 - \mathbb{P}\left( \bigcup_{t=1}^{\infty}\{\Vert \widehat{\bm{\theta}}_{t} - \widehat{\bm{\theta}}_{t-1}\Vert_2 > \delta_t\} \right) \\
    \geq & 1- \sum_{t=1}^{\infty}
     \mathbb{P}\left( \Vert \widehat{\bm{\theta}}_{t} - \widehat{\bm{\theta}}_{t-1}\Vert_2 > \delta_t \right) \\
     \geq  & 1 - C_1\sum_{t=1}^{T} \exp(-C_2 r(c_t n)\delta_t^{\gamma}),
\end{align*}
where the last inequality follows from Assumption \ref{Ass:Uniform}. Therefore, we have
\begin{align}
\label{Ineq1:Thm1}
     \mathbb{P}\left(\Vert \widehat{\bm{\theta}}_{T} - \bm{\theta}^\star\Vert_2 \leq \delta \right) \geq 1 - C_1\sum_{t=1}^{T} \exp(-C_2 r(c_t n)\delta_t^{\gamma}).
\end{align}
Rearranging the inequality in (\ref{Ineq1:Thm1}) gives
\color{black}
\begin{align*}
    \mathbb{P}\left(\Vert \widehat{\bm{\theta}}_{T} - \bm{\theta}^\star\Vert_2 > \delta \right) \leq C_1\sum_{t=1}^{T} \exp(-C_2 r(c_t n)\delta_t^{\gamma}).
\end{align*}
\color{black}
The proof for 
\color{black}
$\mathbb{P}\left(\Vert \widehat{\bm{\theta}}_{\infty} - \bm{\theta}^\star\Vert_2 > \delta \right)$
\color{black}
follows the same reasoning as above, and this completes the proof. \hfill$\blacksquare$ \\

\noindent
\textbf{Proof of Theorem \ref{Thm:Martingale}.} We first derive the following upper bound for $\mathbb{P}\left(
\Vert \widehat{\bm{\theta}}_T - \bm{\theta}^\star \Vert_2
\geq \delta \right)$.
\begin{align*}
    \mathbb{P}\left(
\Vert \widehat{\bm{\theta}}_T - \bm{\theta}^\star \Vert_2
\geq \delta \right) = &
\mathbb{P}\left(
\sum_{i=1}^p \left(\sum_{t=1}^T \xi_{t,i}\right)^2
\geq \delta^2 \right) \leq 
\mathbb{P}\left(
\bigcup_{i=1}^p \left\{\left(\sum_{t=1}^T \xi_{t,i}\right)^2
\geq \frac{\delta^2}{p}\right\} \right)  \\
\leq &\sum_{i=1}^p
\mathbb{P}\left(
\left|\sum_{t=1}^T \xi_{t,i}\right|
\geq \frac{\delta}{\sqrt{p}} \right) \triangleq \sum_{i=1}^p Q_i(T).
\end{align*}
Next, we aim to establish a universal bound for \( Q_i(T) \). To begin with, specify the following definitions in this proof.
\begin{align*}
  c_t = t^{1+s}, \,
    \bm{\xi}_t = \widehat{\bm{\theta}}_{t}-\widehat{\bm{\theta}}_{t-1} \, \text{ and } \,
  &\bm{\xi}_1 = \widehat{\bm{\theta}}_{1}-\bm{\theta}^\star,
\end{align*}
where $s>0$ and $t \geq 2$. Additionally, we highlight the following fact, which follows from the martingale property:
\[
\mathbb{E}(\xi_{t,i} \mid \bm{\xi}_1, \ldots, \bm{\xi}_{t-1}) =
\mathbb{E}\big[\xi_{t,i} \cdot I(|\xi_{t,i}| < M) \mid \bm{\xi}_1, \ldots, \bm{\xi}_{t-1}\big] +
\mathbb{E}\big[\xi_{t,i} \cdot I(|\xi_{t,i}| \geq M) \mid \bm{\xi}_1, \ldots, \bm{\xi}_{t-1}\big] = 0,
\]
for any $t \in [T], i \in [p]$, and $M>0$. This implies that
\begin{align}
\label{Afact}
\mathbb{E}\big[\xi_{t,i} \cdot I(|\xi_{t,i}| < M) \mid \bm{\xi}_1, \ldots, \bm{\xi}_{t-1}\big] 
 =    -
\mathbb{E}\big[\xi_{t,i} \cdot I(|\xi_{t,i}| \geq M) \mid \bm{\xi}_1, \ldots, \bm{\xi}_{t-1}\big] 
.
\end{align}

\noindent
\textbf{Step 1: Bounding $Q_i(T)$.} We can decompose $\xi_{t,i}$ as follows:
\begin{align*}
    \xi_{t,i} = & \xi_{t,i}I(|\xi_{t,i}|< M_t) + 
    \xi_{t,i}I(|\xi_{t,i}| \geq M_t).
\end{align*}
With this, $Q_i(T)$ can be bounded as
\begin{align*}
  &  Q_i(T) = \mathbb{P}\left(
\left|\sum_{t=1}^T \xi_{t,i}\right|
\geq \frac{\delta}{\sqrt{p}} \right) \\ \leq  &
\mathbb{P}\left(
\left|\sum_{t=1}^T \xi_{t,i}I(|\xi_{t,i}|< M_t) \right|
\geq \frac{\delta}{2\sqrt{p}} \right)+
\mathbb{P}\left(
\left|\sum_{t=1}^T 
    \xi_{t,i}I(|\xi_{t,i}|\geq M_t)\right|
\geq \frac{\delta}{2\sqrt{p}} \right) \\
\leq &
\mathbb{P}\left(
\left|\sum_{t=1}^T \xi_{t,i}I(|\xi_{t,i}|< M_t) \right|
\geq \frac{\delta}{2\sqrt{p}} \right)+
\mathbb{P}\left(
\sum_{t=1}^T 
    \left|\xi_{t,i}\right|I(|\xi_{t,i}|\geq M_t)
\geq \frac{\delta}{2\sqrt{p}} \right) 
\triangleq & Q_i^{(1)}(T) + Q_i^{(2)}(T).
\end{align*}
It remains to bound $Q_i^{(1)}(T)$ and $Q_i^{(2)}(T)$ separately.

\noindent
\textbf{Step 2: Bounding $Q_i^{(1)}(T)$.} We first define $\Omega_{t,i}$ as
\begin{align*}
    \Omega_{t,i} = \mathbb{E}
    \Big(
\xi_{t,i}I(|\xi_{t,i}|< M_t)\Big| \bm{\xi}_1,\ldots,\bm{\xi}_{t-1}
    \Big) \text{ for } t \in [T].
\end{align*}
With this, we can further bound $Q_i^{(1)}(T)$ as 
\begin{align*}
Q_i^{(1)}(T) = & \mathbb{P}\left(
\left|\sum_{t=1}^T \xi_{t,i}I(|\xi_{t,i}|< M_t) \right|
\geq \frac{\delta}{2\sqrt{p}} \right) \\ 
\leq &
\mathbb{P}\left(
\left|\sum_{t=1}^T \xi_{t,i}I(|\xi_{t,i}|< M_t) - \sum_{t=1}^T\Omega_{t,i} \right|
\geq \frac{\delta}{2\sqrt{p}} -\Big|\sum_{t=1}^T\Omega_{t,i}\Big| \right) \\
= &
\mathbb{P}\left(
\left|\sum_{t=1}^T \overline{\xi}_{t,i} \right|
\geq \frac{\delta}{2\sqrt{p}} -\Big|\sum_{t=1}^T\Omega_{t,i}\Big| \right),
\end{align*}
where $\overline{\xi}_{t,i} = I(|\xi_{t,i}|< M_t) - \Omega_{t,i}$. 

Note that $\{\overline{\xi}_{t,i}\}_{t=1}^{T}$ is a discrete-time martingale and $|\overline{\xi}_{t,i}| \leq 2M_t$ for any $t \in [T]$ and $i \in [p]$. If $\Big|\sum_{t=1}^T\Omega_{t,i}\Big| \leq  \frac{\delta}{4\sqrt{p}}$, applying the Azuma–Hoeffding inequality \citep{hoeffding1994probability} yields that
\begin{align*}
Q_i^{(1)}(T) \leq
2\exp\left(
-\frac{\delta^2}{64p\sum_{t=1}^TM_t^2}
\right).
\end{align*}
Therefore, we can construct the bound for $Q_i^{(1)}(T)$ as
\begin{align}
\label{Q1_bound}
    Q_i^{(1)}(T) \leq \max\left\{
I\left(\Big|\sum_{t=1}^T\Omega_{t,i}\Big| > \frac{\delta}{4\sqrt{p}}\right), 2\exp\left(
-\frac{\delta^2}{64p\sum_{t=1}^TM_t^2}
\right)
    \right\}.
\end{align}
Using $M_t = \sqrt{\frac{\log n}{n \cdot c_t^L}}$ with $L = \frac{1+s/2}{1+s}$, the upper bound in (\ref{Q1_bound}) becomes
\begin{align}
\label{Q1_Bound_Final}
 Q_i^{(1)}(T) \leq 
 \max\left\{
 I\left(\Big|\sum_{t=1}^T\Omega_{t,i}\Big| > \frac{\delta}{4\sqrt{p}}\right),
2\exp\left(
-\frac{n\delta^2}{64p\sum_{t=1}^T \frac{\log n}{t^{1+s/2}}}
\right) \right\}.
\end{align}
Note that (\ref{Q1_Bound_Final}) is valid only when $\Big|\sum_{t=1}^T\Omega_{t,i}\Big| \leq \frac{\delta}{4\sqrt{p}}$, making $ I\left(\Big|\sum_{t=1}^T\Omega_{t,i}\Big| > \frac{\delta}{4\sqrt{p}}\right)=0$. Next, we analyze the behavior of $\sum_{t=1}^{T}\Omega_{t,i}$. Based on (\ref{Afact}), we have
\begin{align}
\label{Q1_overall_b}
   \left| \sum_{t=1}^{T}\Omega_{t,i} \right| = &   \left|\sum_{t=1}^{T} \mathbb{E}
    \Big(
\xi_{t,i}I(|\xi_{t,i}|< M_t)\Big| \bm{\xi}_1,\ldots,\bm{\xi}_{t-1}
    \Big)\right| 
    = \left|\sum_{t=1}^{T} \mathbb{E}
    \Big(
\xi_{t,i}I(|\xi_{t,i}| \geq M_t)\Big| \bm{\xi}_1,\ldots,\bm{\xi}_{t-1}
    \Big) \right|\notag  \\
    \leq &
    \sum_{t=1}^{T} \mathbb{E}
    \Big(
\left|\xi_{t,i}\right|I(|\xi_{t,i}| \geq M_t)\Big| \bm{\xi}_1,\ldots,\bm{\xi}_{t-1}
    \Big) \notag \\
    \leq & \sum_{t=1}^{T}  \sqrt{\mathbb{E}(\xi_{t,i}^2| \bm{\xi}_1,\ldots,\bm{\xi}_{t-1})} \cdot \sqrt{\mathbb{P}(|\xi_{t,i}| \geq M_t| \bm{\xi}_1,\ldots,\bm{\xi}_{t-1})}\notag  \\
        \leq & \sum_{t=1}^{T}  \sqrt{\mathbb{E}(\xi_{t,i}^2| \bm{\xi}_1,\ldots,\bm{\xi}_{t-1})} \cdot \sqrt{\mathbb{P}(\Vert \bm{\xi}_{t}\Vert_2 \geq M_t| \bm{\xi}_1,\ldots,\bm{\xi}_{t-1})}\notag  \\
    \leq & \left\{
    \sum_{t=1}^{T} \left[\mathbb{E}(\xi_{t,i}^2| \bm{\xi}_1,\ldots,\bm{\xi}_{t-1})\right]^{\frac{a}{2}} \right\}^{\frac{1}{a}} \cdot \left\{\sum_{t=1}^{T}\left[\mathbb{P}(\Vert \bm{\xi}_{t}\Vert_2 \geq M_t| \bm{\xi}_1,\ldots,\bm{\xi}_{t-1})\right]^{\frac{b}{2}}\right\}^{\frac{1}{b}}  \notag \\
    \triangleq & [A_1(T,i,a)]^{\frac{1}{a}} \cdot [A_2(T,i,b)]^{\frac{1}{b}},
\end{align}
where the second inequality follows from the Cauchy–Schwarz inequality and and the last inequality follows from the H\"older inequality with $1/a+1/b=1$. Here the choices of $a$ and $b$ are flexible and will be specified as follows.

\noindent
\textbf{Step 2.1: Bounding $A_1(T,i,a)$.} Note that $\mathbb{E}(\xi_{t,i})=0$ and $\bm{\xi}_t$ satisfies Assumption \ref{Ass:Uniform} with $r(n)=n^{\kappa}$. Using Lemma \ref{lemma:varBound}, we have
\begin{align}
\label{A1_bound}
    A_1(T,i,a) = & \sum_{t=1}^{T} \left[\mathbb{E}(\xi_{t,i}^2| \bm{\xi}_1,\ldots,\bm{\xi}_{t-1})\right]^{\frac{a}{2}} \leq \left[\frac{2 C_1}{\gamma} \cdot \Gamma\left(\frac{2}{\gamma}\right) \cdot C_2^{-\frac{2}{\gamma}} \right]^{\frac{a}{2}}\sum_{t=0}^{T-1}n_t^{-\frac{a\kappa}{\gamma}} \notag \\
    = &
    \left[\frac{2 C_1}{\gamma} \cdot \Gamma\left(\frac{2}{\gamma}\right) \cdot C_2^{-\frac{2}{\gamma}} \right]^{\frac{a}{2}} \frac{1}{n^{\frac{a\kappa}{\gamma}}}\sum_{t=0}^{T-1} t ^{-\frac{a\kappa(1+s)}{\gamma}}.
\end{align}
Clearly, given that $\kappa \geq \gamma/2$ and $a \geq 2$, we have
\begin{align*}
    \lim_{T\rightarrow \infty}A_1(T,i,a) =\lim_{T\rightarrow \infty}
    \left[\frac{2 C_1}{\gamma} \cdot \Gamma\left(\frac{2}{\gamma}\right) \cdot C_2^{-\frac{2}{\gamma}} \right]^{\frac{a}{2}} \frac{1}{n^{\frac{a\kappa}{\gamma}}}\sum_{t=0}^{T-1} t ^{-\frac{a\kappa(1+s)}{\gamma}}<\infty.
\end{align*}
This then implies that
\begin{align*}
    \lim_{n\rightarrow \infty}\lim_{T\rightarrow \infty}A_1(T,i,a) = 0.
\end{align*}

\noindent
\textbf{Step 2.2: Bounding $A_2(T,i,a)$.} Under Assumption \ref{Ass:Uniform}, $A_2(T,i,b)$ can be bounded as
\begin{align*}
A_2(T,i,b) =  \sum_{t=1}^T [\mathbb{P}\left(\Vert \bm{\xi}_t\Vert_2 \geq M_t |\bm{\xi}_1,\ldots,\bm{\xi}_{t-1} 
    \right)]^{\frac{b}{2}} \leq  C_1^{\frac{b}{2}}\sum_{t=1}^T \exp(-bC_2 r(n_t) M_t^\gamma/2).
\end{align*}
Under the assumption that $r(n)=n^{\kappa}$ for some $\kappa>0$ and $M_t = \sqrt{\frac{\log n}{n \cdot c_t^L}}$, $A_2(T,i,b)$ can be written as
\begin{align*}
    A_2(T,i,b) \leq C_1^{\frac{b}{2}}\sum_{t=1}^T \exp(-C_2' n_t^{\kappa} M_t^\gamma) =   C_1^{\frac{b}{2}}\sum_{t=1}^T \exp(-C_2' c_t^{\kappa-L\gamma/2}n^{\kappa-\gamma/2} (\log n)^{\gamma/2}),
\end{align*}
where $C_2' = bC_2/2$.

Under the assumption that $c_t = t^{1+s}$ for some $s>0$ and $\kappa \geq \gamma/2$, we can verify that $\exp(-C_2' t^{(1+s)(\kappa-L\gamma/2)}n^{\kappa-\gamma/2} (\log n)^{\gamma/2})$ is a decreasing function with $t$ given that $\kappa-L\gamma/2>0$. Therefore, we have
\begin{align*}
A_2(T,i,b) \leq & C_1^{\frac{b}{2}}\int_1^{T}  \exp(-C_2' t^{(1+s)(\kappa-L\gamma/2)}n^{\kappa-\gamma/2} (\log n)^{\gamma/2}) dt \\
    \leq & 
    C_1^{\frac
    {b}{2}} \int_1^{\infty}\exp(-C_2' t^{(1+s)(\kappa-L\gamma/2)}n^{\kappa-\gamma/2} (\log n)^{\gamma/2}) dt.
\end{align*}
For ease of notation, we denote $E(n) = C_2' n^{\kappa-\gamma/2} (\log n)^{\gamma/2}$ and $\tau =(1+s)(\kappa-L\gamma/2)$. Then we have
\begin{align*}
   &C_1^{\frac{b}{2}}\int_1^{\infty}\exp(-C_2' t^{(1+s)(\kappa-L\gamma/2)}n^{\kappa-\gamma/2} (\log n)^{\gamma/2}) dt =  C_1^{\frac{b}{2}}\int_1^{\infty} \exp(-t^{\tau}E(n))dt \\
   = & \frac{C_1^{\frac{b}{2}}}{\tau}\int_1^{\infty} u^{\frac{1}{\tau}-1} \exp(-uE(n))du =
   \frac{C_1^{\frac{b}{2}}}{\tau E(n)}\int_{E(n)}^{\infty} \left(\frac{v}{E(n)}\right)^{\frac{1}{\tau}-1} \exp(-v)dv  \\ 
   = &\frac{C_1^{\frac{b}{2}} }{\tau [E(n)]^{\frac{1}{\tau}}}\int_{E(n)}^{\infty} v^{\frac{1}{\tau}-1} \exp(-v)dv \leq \frac{BC_1^{\frac{b}{2}} }{\tau }
    [E(n)]^{-1} \exp(-E(n)),
\end{align*}
for some positive constant $B$, where the last inequality follows from the Taylor series approximation of the incomplete Gamma function \citep{natalini2000inequalities}.

To sum up, for any $s>0$, it holds that
\begin{align}
\label{A2_bound}
    A_2(T,i,b) \leq \frac{BC_1}{\tau }
    [E(n)]^{-1} \exp(-E(n)).
\end{align}
Here, the upper bound in (\ref{A2_bound}) holds true for any $T \geq 1$.

Combining (\ref{A1_bound}) and (\ref{A2_bound}) yields that
\begin{align}
\label{Omega_Bound}
  & \left| \sum_{t=1}^{T}\Omega_{t,i} \right| 
   \leq  [A_1(T,i,a)]^{\frac{1}{a}} \cdot [A_2(T,i,b)]^{\frac{1}{b}}  \notag \\
   \leq &
   \left[\frac{2 C_1}{\gamma} \cdot \Gamma\left(\frac{2}{\gamma}\right) \cdot C_2^{-\frac{2}{\gamma}} \right]^{\frac{1}{2}} \frac{1}{n^{\frac{\kappa}{\gamma}}} \left[\sum_{t=0}^{T-1} t ^{-\frac{a\kappa(1+s)}{\gamma}}\right]^{\frac{1}{a}}
   \cdot
   \left(\frac{BC_1}{\tau }\right)^{\frac{1}{b}}
    [E(n)]^{-1/b} \exp(-E(n)/b)
\end{align}
For ease of notation, we denote
$$
\mathcal{K}(T,n) = \left[\frac{2 C_1}{\gamma} \cdot \Gamma\left(\frac{2}{\gamma}\right) \cdot C_2^{-\frac{2}{\gamma}} \right]^{\frac{1}{2}} \frac{1}{n^{\frac{\kappa}{\gamma}}} \left[\sum_{t=0}^{T-1} t ^{-\frac{a\kappa(1+s)}{\gamma}}\right]^{\frac{1}{a}}
   \cdot
   \left(\frac{BC_1}{\tau }\right)^{\frac{1}{b}}
    [E(n)]^{-1/b} \exp(-E(n)/b).
$$
Clearly, taking $a=b=2$, we have
\begin{align}
\label{K_limit}
&\lim_{T\rightarrow \infty}
    \left| \sum_{t=1}^{T}\Omega_{t,i} \right|
    \leq \lim_{T\rightarrow \infty} \mathcal{K}(T,n)<\infty, \notag \\
  &  \lim_{n\rightarrow \infty}
    \lim_{T\rightarrow \infty}
    \left| \sum_{t=1}^{T}\Omega_{t,i} \right| =      \lim_{n\rightarrow \infty}
    \lim_{T\rightarrow \infty}\mathcal{K}(T,n) =0.
\end{align}

\noindent
\textbf{Step 3: Bounding $Q_i^{(2)}(T)$.} Using Markov's inequality, we have
\begin{align*}
  Q_i^{(2)}(T) =   \mathbb{P}\left(
\sum_{t=1}^T 
    \left|\xi_{t,i}\right|I(|\xi_{t,i}|\geq M_t)
\geq \frac{\delta}{2\sqrt{p}} \right)  \leq 
\frac{2\sqrt{p}\mathbb{E}\left[\sum_{t=1}^T 
    \left|\xi_{t,i}\right|I(|\xi_{t,i}|\geq M_t)\right]}{\delta}.
\end{align*}
Using the bound in (\ref{Q1_overall_b}), we have
\begin{align}
\label{Q2_Bound_Final}
    Q_i^{(2)}(T) \leq &\frac{2\sqrt{p}}{\delta} \cdot
    [A_1(T,i,a)]^{\frac{1}{a}} \cdot [A_2(T,i,b)]^{\frac{1}{b}} \notag \\
    \leq  &
    \frac{2\sqrt{p}}{\delta}  \cdot
    \left[\frac{2 C_1}{\gamma} \cdot \Gamma\left(\frac{2}{\gamma}\right) \cdot C_2^{-\frac{2}{\gamma}} \right]^{\frac{1}{2}} \frac{1}{n^{\frac{\kappa}{\gamma}}} \left[\sum_{t=0}^{T-1} t ^{-\frac{a\kappa(1+s)}{\gamma}}\right]^{\frac{1}{a}}
   \cdot
   \left(\frac{BC_1}{\tau }\right)^{\frac{1}{b}}
    [E(n)]^{-1/b} \exp(-E(n)/b)\notag \\
    =& \frac{2\sqrt{p}}{\delta}  \mathcal{K}(T,n).
\end{align}

\noindent
\textbf{Step 4: Bounding $Q_i(T)$.} To sum up, we have the following bound for $Q_i(T)$. Combining (\ref{Q1_Bound_Final}) and (\ref{Q2_Bound_Final}) yields that
\begin{align*}
Q_i(T) \leq & 
 \max\left\{
 I\left(\Big|\sum_{t=1}^T\Omega_{t,i}\Big| > \frac{\delta}{4\sqrt{p}}\right),
2\exp\left(
-\frac{n\delta^2}{2p\sum_{t=1}^T \frac{\log n}{t^{1+s/2}}}
\right) \right\}+ \frac{2\sqrt{p}}{\delta}\mathcal{K}(T,n) \\
\leq &
 \max\left\{
 I\left(\mathcal{K}(T,n) > \frac{\delta}{4\sqrt{p}}\right),
2\exp\left(
-\frac{n\delta^2}{2p\sum_{t=1}^T \frac{\log n}{t^{1+s/2}}}
\right) \right\}+ \frac{2\sqrt{p}}{\delta}\mathcal{K}(T,n),
\end{align*}
for any $i \in [p]$, where the last inequality follows from (\ref{Omega_Bound}).
Finally, we have
\begin{align*}
     &   \mathbb{P}\left(
\Vert \widehat{\bm{\theta}}_T - \bm{\theta}^\star \Vert_2 
\geq \delta \right) \leq \sum_{i=1}^p Q_i(T)  \\
\leq & p\max\left\{
 I\left(\mathcal{K}(T,n) > \frac{\delta}{4\sqrt{p}}\right),
2\exp\left(
-\frac{n\delta^2}{2p\sum_{t=1}^T \frac{\log n}{t^{1+s/2}}}
\right) \right\}+ \frac{2p\sqrt{p}}{\delta}\mathcal{K}(T,n)
\end{align*}
Based on (\ref{K_limit}), we have
\begin{align*}
   \lim_{n\rightarrow \infty}   \lim_{T\rightarrow \infty} \mathbb{P}\left(
\Vert \widehat{\bm{\theta}}_T - \bm{\theta}^\star \Vert_2
\geq \delta \right) = 0,
\end{align*}
for any $\delta>0$. This completes the proof. \hfill$\blacksquare$ \\

\noindent 
\textbf{Proof of Theorem \ref{Thm:positive_bias}}: In this proof, there are two cases $\rho \geq 1$ and $\frac{\kappa}{\gamma} \leq \rho <1$ if $1 \leq 2\kappa/\gamma \leq 2$.

\noindent
\textbf{Case 1: when $\rho \geq 1$.}
Define $\overline{\bm{\xi}}_t = \widehat{\bm{\theta}}_{t}-\mathbb{E}(\widehat{\bm{\theta}}_{t}|\widehat{\bm{\theta}}_{t-1}) \text{ and } 
  \bm{b}_t = \mathbb{E}(\widehat{\bm{\theta}}_{t}|\widehat{\bm{\theta}}_{t-1}) -  \widehat{\bm{\theta}}_{t-1}$. Then we have the following decomposition:
\begin{align}
\label{bias_ori_bound}
\mathbb{E}[\Vert \widehat{\bm{\theta}}_T-\bm{\theta}^\star\Vert_2^2]=&\sum_{t=1}^{T}\mathbb{E}\big[\Vert \overline{\bm{\xi}}_t\Vert_2^2\big]+2\sum_{1\leq i<j\leq T}\mathbb{E}\langle \overline{\bm{\xi}}_i,\overline{\bm{\xi}}_j \rangle+\mathbb{E}\Big\Vert \sum_{t=1}^{T}\bm{b}_t\Big\Vert_2^2 +
2\mathbb{E}\langle \sum_{t=1}^{T}\overline{\bm{\xi}}_t, \sum_{t=1}^{T}\bm{b}_t\rangle
\notag \\
\lesssim &\sum_{t=1}^{T}\mathbb{E}\big[\Vert \overline{\bm{\xi}}_t\Vert_2^2\big]+\mathbb{E}\Big\Vert \sum_{t=1}^{T}\bm{b}_t\Big\Vert_2^2,
\end{align}
where the inequality follows from the Cauchy–Schwarz inequality. By Lemma \ref{lemma:varBound} and the fact that variance is upper bounded by the second moment, we have
 \begin{align*}
        \mathbb{E}[\Vert \overline{\bm{\xi}}_t \Vert_2^2] \leq \mathbb{E}[\Vert \widehat{\bm{\theta}}_t-\widehat{\bm{\theta}}_{t-1} \Vert_2^2]=
        \mathbb{E}\left\{
        \mathbb{E}[\Vert \widehat{\bm{\theta}}_t-\widehat{\bm{\theta}}_{t-1} \Vert_2^2|\widehat{\bm{\theta}}_{t-1}]\right\} \leq 
        \frac{2 C_1}{\gamma} \cdot \Gamma\left( \frac{2}{\gamma} \right) \cdot C_2^{-2/\gamma} (c_tn)^{-\frac{2\kappa}{\gamma}},
    \end{align*}
then similar to the proof of Theorem \ref{Thm:Martingale}, \eqref{bias_ori_bound} can be further bounded as 
\begin{align*}
\mathbb{E}[\Vert \widehat{\bm{\theta}}_T-\bm{\theta}^\star\Vert_2^2] &\lesssim (n)^{-\frac{2\kappa}{\gamma}}\frac{2 C_1}{\gamma} \cdot \Gamma\left( \frac{2}{\gamma} \right) \cdot C_2^{-2/\gamma} \zeta(2(1+s)\kappa/\gamma)+C\left(\sum_{t=1}^{T}\frac{1}{n_{t}^{\rho}}\right)^2\Vert \bm{v} \Vert_2^2 \\
& \lesssim (n)^{-\frac{2\kappa}{\gamma}}\frac{2 C_1}{\gamma} \cdot \Gamma\left( \frac{2}{\gamma} \right) \cdot C_2^{-2/\gamma} \zeta(2(1+s)\kappa/\gamma)+C(n)^{-2}\left(\sum_{t=1}^{T}\frac{1}{c_t^{\rho}}\right)^2 \Vert \bm{v} \Vert_2^2 \\ 
& \lesssim (n)^{-\frac{2\kappa}{\gamma}}\frac{2 C_1}{\gamma} \cdot \Gamma\left( \frac{2}{\gamma} \right) \cdot C_2^{-2/\gamma} \zeta(2(1+s)\kappa/\gamma)+C(n)^{-2}\left(\sum_{t=1}^{\infty}\frac{1}{c_t^{\rho}}\right)^2 \Vert \bm{v} \Vert_2^2 \\ 
& \lesssim (n)^{-\frac{2\kappa}{\gamma}}\frac{2 C_1}{\gamma} \cdot \Gamma\left( \frac{2}{\gamma} \right) \cdot C_2^{-2/\gamma} \zeta(2(1+s)\kappa/\gamma)+C(n)^{-2}[\zeta(\rho(1+s)]^2,
\end{align*}
for some positive constant $C$, where $C_1$ and $C_2$ are as defined in Assumption \ref{Ass:Uniform}, and $\bm{v}=(v_1,\ldots,v_p)$ are as defined in Assumption \ref{Ass:biased}.

If $\rho \geq 1$ and $\kappa \geq \gamma/2$, then by Chebyshev Inequality,
\begin{align*}
\lim_{n\rightarrow \infty}\lim_{T\rightarrow \infty}
\mathbb{P}\left( \|\widehat{\bm{\theta}}_T - \bm{\theta}^\star\|_2 \geq \delta \right)\leq
\lim_{n\rightarrow \infty}\lim_{T\rightarrow \infty}
\frac{1}{\delta^2}\mathbb{E}\Vert\widehat{\bm{\theta}}_T-\bm{\theta}^\star\Vert_2^2 =0,
\end{align*}
for any $\delta>0$ and $c_t=t^{1+s}$ with any $s>0$.

\noindent
\textbf{Case 2: when $\frac{\kappa}{\gamma} \leq \rho <1$.} Next, we proceed to show that 
\begin{align*}
\lim_{n\rightarrow \infty}\lim_{T\rightarrow \infty}
\mathbb{P}\left( \|\widehat{\bm{\theta}}_T - \bm{\theta}^\star\|_2 \geq \delta \right)=1,
\end{align*}
for $ \frac{\kappa}{\gamma} \leq \rho <1$. First, for any $T \geq 1$, we have
\begin{align}
\label{LB_1}
    &\mathbb{P}\left( \|\widehat{\bm{\theta}}_T - \bm{\theta}^\star\|_2 \geq \delta \right) \geq 
    \mathbb{P}\left( |\widehat{\theta}_{T,i} - \theta_i^\star|\geq \delta \right) \notag \\
    = & 
    \mathbb{P}\left( \Big|\sum_{t=1}^T\left(\widehat{\theta}_{t,i}-\mathbb{E}(\widehat{\theta}_{t,i}|\widehat{\theta}_{t-1})+\mathbb{E}(\widehat{\theta}_{t,i}|\widehat{\theta}_{t-1})-\widehat{\theta}_{t-1,i}\right) \Big|\geq \delta \right) \notag \\
    \geq &
    \mathbb{P}\left( \sum_{t=1}^T\left(\widehat{\theta}_{t,i}-\mathbb{E}(\widehat{\theta}_{t,i}|\widehat{\theta}_{t-1}) \right) \geq \delta - \sum_{t=1}^T \left(\mathbb{E}(\widehat{\theta}_{t,i}|\widehat{\theta}_{t-1})-\widehat{\theta}_{t-1,i}\right) \right),
\end{align}
where $\widehat{\theta}_{0,i}=\theta^\star_{i}$ for ease of notation.
By Assumption \ref{Ass:biased}, we have \(\mathbb{E}(\widehat{\theta}_{t,i}|\widehat{\theta}_{t-1}) - \widehat{\theta}_{t-1,i} \asymp \frac{v_i}{n_t^\rho}\) for each \(t \geq 1\). Without loss of generality, we assume that \(v_i > 0\). Hence, inequality (\ref{LB_1}) can be further lower bounded as follows
\begin{align*}
\mathbb{P}\left( \|\widehat{\bm{\theta}}_T - \bm{\theta}^\star\|_2 \geq \delta \right) \geq &\mathbb{P}\left( \sum_{t=1}^T \left(\widehat{\theta}_{t,i}-\mathbb{E}(\widehat{\theta}_{t,i}|\widehat{\theta}_{t-1})\right)\geq \delta -C \sum_{t=1}^T \frac{v_i}{n_t^{\rho}}  \right) \\
=& 1 - \mathbb{P}\left( \sum_{t=1}^T \left(\widehat{\theta}_{t,i}-\mathbb{E}(\widehat{\theta}_{t,i}|\widehat{\theta}_{t-1})\right)<\delta -C \sum_{t=1}^T \frac{v_i}{n_t^{\rho}}  \right).
\end{align*}
Note that when $\kappa/\gamma \leq \rho<1$, there exists some small $s>0$ such that $(1+s)\rho<1$. Therefore, $\sum_{t=1}^T \frac{v_i}{n_t^{\rho}} \rightarrow \infty$ as $T\rightarrow \infty$. Without loss of generality, we suppose $C \sum_{t=1}^T \frac{v_i}{n_t^{\rho}} \gg \delta$. Therefore, we further have
\begin{align*}
    &\mathbb{P}\left( \sum_{t=1}^T \left(\widehat{\theta}_{t,i}-\mathbb{E}(\widehat{\theta}_{t,i}|\widehat{\theta}_{t-1})\right)<\delta -C \sum_{t=1}^T \frac{v_i}{n_t^{\rho}}  \right)  \\
    \leq  &
    \mathbb{P}\left( \Big|\sum_{t=1}^T \left(\widehat{\theta}_{t,i}-\mathbb{E}(\widehat{\theta}_{t,i}|\widehat{\theta}_{t-1})\right)\Big| \geq C \sum_{t=1}^T \frac{v_i}{n_t^{\rho}}-\delta  \right) \\
    \leq &
    \frac{\mathbb{E}\left[\sum_{t=1}^T \left(\widehat{\theta}_{t,i}-\mathbb{E}(\widehat{\theta}_{t,i}|\widehat{\theta}_{t-1})\right)\right]^2}{\left[C \sum_{t=1}^T \frac{v_i}{n_t^{\rho}}-\delta\right]^2} \leq 
    \frac{\sum_{t=1}^{\infty}\mathbb{E}\left[ \left(\widehat{\theta}_{t,i}-\mathbb{E}(\widehat{\theta}_{t,i}|\widehat{\theta}_{t-1})\right)^2\right]}{\left[C \sum_{t=1}^T \frac{v_i}{n_t^{\rho}}-\delta\right]^2} \\
    \leq & \frac{\sum_{t=1}^{\infty}\mathbb{E}[\Vert \overline{\bm{\xi}}_t \Vert_2^2]}{\left[C \sum_{t=1}^T \frac{v_i}{n_t^{\rho}}-\delta\right]^2}
    \leq \frac{\sum_{t=1}^{\infty}\frac{2 C_1}{\gamma} \cdot \Gamma\left( \frac{2}{\gamma} \right) \cdot C_2^{-2/\gamma} (c_tn)^{-\frac{2\kappa}{\gamma}}}{\left[C \sum_{t=1}^T \frac{v_i}{n_t^{\rho}}-\delta\right]^2},
\end{align*}
where the third inequality follows from the fact that $\mathbb{E}\left[\left(\widehat{\theta}_{t,i}-\mathbb{E}(\widehat{\theta}_{t,i}|\widehat{\theta}_{t-1})\right)\left(\widehat{\theta}_{t-l,i}-\mathbb{E}(\widehat{\theta}_{t-l,i}|\widehat{\theta}_{l-1})\right)\right]=0$ for any $t \geq 1$ and $l \leq t-1$. 

Given that $2\kappa/\gamma \geq 1$ and $\frac{\kappa}{\gamma} \leq \rho<1$, we have
\begin{align*}
    &\sum_{t=1}^{\infty}\frac{2 C_1}{\gamma} \cdot \Gamma\left( \frac{2}{\gamma} \right) \cdot C_2^{-2/\gamma} (c_tn)^{-\frac{2\kappa}{\gamma}}<\infty, \\
    &\lim_{T\rightarrow \infty} C \sum_{t=1}^T \frac{v_i}{n_t^{\rho}}-\delta = \infty.
\end{align*}
This implies that
\begin{align*}
\lim_{T\rightarrow \infty}
    \mathbb{P}\left( \|\widehat{\bm{\theta}}_T - \bm{\theta}^\star\|_2 \geq \delta \right) 
\geq& 1 - \lim_{T\rightarrow \infty}\mathbb{P}\left( \sum_{t=1}^T \left(\widehat{\theta}_{t,i}-\mathbb{E}(\widehat{\theta}_{t,i}|\widehat{\theta}_{t-1})\right)<\delta -C \sum_{t=1}^T \frac{v_i}{n_t^{\rho}}  \right) \\
\geq & 1 - \lim_{T\rightarrow \infty}
\frac{\sum_{t=1}^{\infty}\frac{2 C_1}{\gamma} \cdot \Gamma\left( \frac{2}{\gamma} \right) \cdot C_2^{-2/\gamma} (c_tn)^{-\frac{2\kappa}{\gamma}}}{\left[C \sum_{t=1}^T \frac{v_i}{n_t^{\rho}}-\delta\right]^2} = 1.
\end{align*}
Then the desired result immediately follows and this completes the proof. \hfill$\blacksquare$ \\

\noindent
\textbf{Proof of Theorem \ref{Thm:GaussianBetter}.} We first consider the following decomposition:
\begin{align*}
    \widehat{\bm{\theta}}_T -\bm{\theta}^\star = 
    \sum_{t=2}^{T} \left[\widehat{\bm{\theta}}_{t} -\widehat{\bm{\theta}}_{t-1}\right] +\widehat{\bm{\theta}}_{1} - \bm{\theta}^\star = \sum_{t=1}^{T}\bm{\xi}_{t},
\end{align*}
where $\bm{\xi}_{1} = \widehat{\bm{\theta}}_{1} - \bm{\theta}^\star $ and $\bm{\xi}_{t} = \widehat{\bm{\theta}}_{t} -\widehat{\bm{\theta}}_{t-1}$ for $t \geq 2$. Note that $\widehat{\bm{\theta}}_{t}$ is an unbiased estimate of $\widehat{\bm{\theta}}_{t-1}$ conditional on $\widehat{\bm{\theta}}_{t-1}$. Therefore, we have $ \mathbb{E}[\bm{\xi}_{t}\bm{\xi}_{l}] = 0 \text{ for any } t \neq l$. Note that $\bm{\xi}_t \sim N(\bm{0},n_{t-1}^{-1}\bm{\Sigma})$. Therefore, $\bm{\xi}_t$'s are independent multivariate Gaussian vectors. With this, $P(T)$ can be re-written as
\begin{align*}
    P(T) = \mathbb{P}\left(\Vert \sum_{t=1}^T \bm{\xi}_t \Vert_2 < \Vert \bm{\xi}_1 \Vert_2\right) = \mathbb{P}\left(\sqrt{n}\Vert \sum_{t=1}^T \bm{\xi}_t \Vert_2 < \sqrt{n}\Vert \bm{\xi}_1 \Vert_2\right) = 
    \mathbb{E}_{\bm{X}}\left[\Phi\left(-\frac{\|\bm{X}\|_2}{2\Vert \bm{\Sigma}^{\frac{1}{2}} \widetilde{\bm{X}}_T \Vert_2}\right)\right],
\end{align*}
where $\bm{X} = \sqrt{n}\sum_{t=2}^T \bm{\xi} \sim N(\bm{0},\sum_{t=1}^{T-1}c_{t}^{-1}\bm{\Sigma})$ and the last equation follows from Lemma \ref{Lemma:Improve}. This completes the proof. \hfill$\blacksquare$ \\

\noindent
\textbf{Proof of Theorem \ref{Thm:General}.} We first prove the asymptotic normality by induction under Assumption \ref{Ass:AssNorm} stating that for any \( T \geq 1 \), it holds that
\begin{align*}
    \sqrt{n}(\widehat{\bm{\theta}}_T-\bm{\theta}^\star) \xrightarrow{d} N(\bm{0},\bm{\Sigma}_T(\bm{\theta}^\star)),
\end{align*}
where $\bm{\Sigma}_T(\bm{\theta}^\star)=\sum_{t=0}^{T-1}c_t^{-1}\bm{\Sigma}(\bm{\theta}^\star)$ is a covariance matrix depending on $T$.

\noindent
\textbf{Case \(T=1\).} This case is directly derived based on Assumption~\ref{Ass:AssNorm}. Specifically,
\begin{align*}
   \sqrt{n}(\widehat{\bm{\theta}}_1-\bm{\theta}^\star) \xrightarrow{d} N(\bm{0},\bm{\Sigma}(\bm{\theta}^\star)).
\end{align*}

\noindent
\textbf{Case \(T=k\).} For $T=k$ with $k \geq 1$, we assume that
\begin{align*}
   \sqrt{n}(\widehat{\bm{\theta}}_k-\bm{\theta}^\star) \xrightarrow{d} N(\bm{0},\bm{\Sigma}_k(\bm{\theta}^\star)),
\end{align*}
where $\bm{\Sigma}_k(\bm{\theta}^\star) =\sum_{t=0}^{k-1}c_t^{-1}\bm{\Sigma}(\bm{\theta}^\star)$.

\noindent
\textbf{Induction Step: Case \(T=k+1\).} We want to show that
\[
\sqrt{n} (\widehat{\bm{\theta}}_{k+1} - \bm{\theta}^\star) \xrightarrow{d} N\left( \bm{0}, \bm{\Sigma}_{k+1}(\bm{\theta}^\star) \right),
\]
where \(\bm{\Sigma}_{k+1}(\bm{\theta}^\star) = \bm{\Sigma}_k(\bm{\theta}^\star) + c_k^{-1}\bm{\Sigma}(\bm{\theta}^\star)\).

We first decompose
\[
\sqrt{n} (\widehat{\bm{\theta}}_{k+1} - \bm{\theta}^\star) = \sqrt{n} (\widehat{\bm{\theta}}_{k+1} - \widehat{\bm{\theta}}_k) + \sqrt{n} (\widehat{\bm{\theta}}_k - \bm{\theta}^\star).
\]
Conditionally on \(\widehat{\bm{\theta}}_k\), we have
\[
\sqrt{n} (\widehat{\bm{\theta}}_{k+1} - \widehat{\bm{\theta}}_k) \mid \widehat{\bm{\theta}}_k \xrightarrow{d} N\left( \bm{0}, c_k^{-1}\bm{\Sigma}(\widehat{\bm{\theta}}_k) \right).
\]
As \( n \to \infty \), by assumption, we have $\widehat{\bm{\theta}}_k \overset{p}{\to} \bm{\theta}^\star$, and by the continuous mapping theorem, we further have $\bm{\Sigma}(\widehat{\bm{\theta}}_k) \overset{p}{\to} \bm{\Sigma}(\bm{\theta}^\star)$. Next, we proceed to show that
\begin{align*}
    \sqrt{n} (\widehat{\bm{\theta}}_{k+1} - \widehat{\bm{\theta}}_k)
    \xrightarrow{d} N\left( \bm{0}, c_{k}^{-1}\bm{\Sigma}(\bm{\theta}^\star) \right).
\end{align*}

For ease of notation, we denote that $\bm{Z}_n := \sqrt{n} (\widehat{\bm{\theta}}_{k+1} - \widehat{\bm{\theta}}_k)$. Since \(\bm{Z}_n \mid \widehat{\bm{\theta}}_k  \xrightarrow{d} N(\bm{0}, c_k^{-1} \bm{\Sigma}(\widehat{\bm{\theta}}_k))\), its conditional characteristic function is:
\[
\phi_n(\bm{t} \mid \widehat{\bm{\theta}}_k) \xrightarrow{n\rightarrow \infty}\exp\left(-\frac{1}{2} \bm{t}^\top c_k^{-1} \bm{\Sigma}(\widehat{\bm{\theta}}_k) \bm{t} \right) \text{ for any } \bm{t}.
\]
Now consider the unconditional characteristic function
\[
\phi_n(\bm t) = \mathbb{E}[e^{i \bm{t}^\top \bm{Z}_n}] = \mathbb{E}[\phi_n(\bm t \mid \widehat{\bm{\theta}}_k)].
\]
Since \(\widehat{\bm{\theta}}_k \overset{p}{\to} \bm{\theta}^\star\), and \(\bm{\Sigma}(\cdot)\) is continuous, we have $\bm{\Sigma}(\widehat{\bm{\theta}}_k) \overset{p}{\to} \bm{\Sigma}(\bm{\theta}^\star)$. Therefore, the conditional characteristic function converges:
\[
\exp\left(-\frac{1}{2} \bm t^\top c_k^{-1} \bm{\Sigma}(\widehat{\bm{\theta}}_k) \bm t \right) \overset{p}{\to} \exp\left(-\frac{1}{2} \bm t^\top c_k^{-1} \bm{\Sigma}(\bm{\theta}^\star)\bm t \right).
\]
Note that $\big|\exp\left(-\frac{1}{2} \bm t^\top c_k^{-1} \bm{\Sigma}(\widehat{\bm{\theta}}_k) \bm t \right)\big| \leq 1$ for any value of $\widehat{\bm{\theta}}_k$. By the Vitali convergence theorem, we have
\begin{align*}
    \mathbb{E}\left[\exp\left(-\frac{1}{2} \bm t^\top c_k^{-1} \bm{\Sigma}(\widehat{\bm{\theta}}_k) \bm t \right)\right]
    \xrightarrow{n\rightarrow\infty}  \exp\left(-\frac{1}{2} \bm t^\top c_k^{-1} \bm{\Sigma}(\bm{\theta}^\star)\bm t \right).
\end{align*}

By Assumption \ref{Ass:AssNorm}, we have
\begin{align*}
\phi_n(\bm t \mid \widehat{\bm{\theta}}_k) \xrightarrow{n\rightarrow\infty}
\exp\left(-\frac{1}{2} \bm t^\top c_k^{-1} \bm{\Sigma}(\widehat{\bm{\theta}}_k) \bm t \right),
\end{align*}
for any $\widehat{\bm{\theta}}_k$. Note that \( |\phi_n(\bm{t} \mid \widehat{\bm{\theta}}_k)| \leq 1 \) by the fundamental property of characteristic functions. Using the dominated convergence theorem, we have
\begin{align*}
    \phi_n(\bm{t}) = \mathbb{E}\left[\phi_n(\bm{t} \mid \widehat{\bm{\theta}}_k)\right]
    \xrightarrow{n\rightarrow\infty} \mathbb{E}\left[\exp\left(-\frac{1}{2} \bm t^\top c_k^{-1} \bm{\Sigma}(\widehat{\bm{\theta}}_k) \bm t \right)\right]\xrightarrow{n\rightarrow\infty}  \exp\left(-\frac{1}{2} \bm t^\top c_k^{-1} \bm{\Sigma}(\bm{\theta}^\star)\bm t \right).
\end{align*}
This then implies that 
\begin{align*}
    \sqrt{n} (\widehat{\bm{\theta}}_{k+1} - \widehat{\bm{\theta}}_k)
    \xrightarrow{d} N\left( \bm{0}, c_{k}^{-1}\bm{\Sigma}(\bm{\theta}^\star) \right).
\end{align*}
By the Slutsky's theorem, we have
$$
\sqrt{n} (\widehat{\bm{\theta}}_{k+1} - \bm{\theta}^\star) \xrightarrow{d} N\left( \bm{0}, \bm{\Sigma}_{k+1}(\bm{\theta}^\star) \right).
$$
To sum up, for any $T\geq 1$, we have
\begin{align*}
    \sqrt{n}(\widehat{\bm{\theta}}_T-\bm{\theta}^\star) \xrightarrow{d} N(\bm{0},\bm{\Sigma}_T(\bm{\theta}^\star)),
\end{align*}
where $\bm{\Sigma}_T(\bm{\theta}^\star) = \sum_{t=0}^{T-1}c_{t}^{-1}\bm{\Sigma}(\bm{\theta}^\star)$.

Based on the above results, we have
\begin{align*}
\begin{cases}
     &\sqrt{n}(\widehat{\bm{\theta}}_T-\bm{\theta}^\star) \xrightarrow{d} N(\bm{0},\bm{\Sigma}_T(\bm{\theta}^\star)), \\
     & \sqrt{n}(\widehat{\bm{\theta}}_1-\bm{\theta}^\star) \xrightarrow{d} N(\bm{0},\bm{\Sigma}(\bm{\theta}^\star)), \\
     & \lim_{n\rightarrow \infty}n\mathbb{E}(\widehat{\bm{\theta}}_T-\bm{\theta}^\star)
    (\widehat{\bm{\theta}}_1-\bm{\theta}^\star)  = \bm{\Sigma}(\bm{\theta^\star}).
\end{cases}
\end{align*}
Using the Cram\'er--Wold theorem, it follows that
\begin{align*}
    \big(\sqrt{n}(\widehat{\bm{\theta}}_T-\bm{\theta}^\star),
\sqrt{n}(\widehat{\bm{\theta}}_1-\bm{\theta}^\star)\big)\xrightarrow{d}
(\bm{A},\bm{B}),
\end{align*}
where $\bm{A} \sim N(\bm{0},\bm{\Sigma}_T(\bm{\theta}^\star))$ and $\bm{B}\sim N(\bm{0},\bm{\Sigma}(\bm{\theta}^\star))$ with $\text{Cov}(\bm{A},\bm{B}) = \bm{\Sigma}(\bm{\theta^\star})$. Therefore, we have
\begin{align*}
    \lim_{n\rightarrow \infty}\mathbb{P}\big(\Vert \widehat{\bm{\theta}}_T - \bm{\theta}^\star \Vert_2 < \Vert \widehat{\bm{\theta}}_1 - \bm{\theta}^\star \Vert_2\big) =
\mathbb{P}\big(\Vert \bm{A} \Vert_2 < \Vert \bm{B}\Vert_2\big).
\end{align*}
Note that we can decompose $\bm{A}$ as 
\begin{align*}
    \bm{A} = \bm{B} + \bm{A}_1 +\cdots \bm{A}_{T-1},
\end{align*}
where $\bm{A}_t \sim N(\bm{0},c_{t}^{-1}\bm{\Sigma}(\bm \theta^\star))$. Then, based on Lemma \ref{Lemma:Improve}, we have
    \begin{align*}
        \lim_{n\rightarrow \infty}P(T) =  \mathbb{E}_{\bm{X}}\left[\Phi\left(-\frac{\|\bm{X}\|_2}{2\Vert \bm{\Sigma}^{\frac{1}{2}}(\bm{\theta}^\star) \widetilde{\bm{X}}_T \Vert_2}\right)\right],
    \end{align*}
    where $\widetilde{\bm{X}}_T = \bm{X}/\Vert\bm{X}\Vert_2$ and $\bm{X} \sim N(\bm{0},\sum_{t=1}^{T-1}c_t^{-1}\bm{\Sigma}(\bm{\theta}^\star))$. This completes the proof. \hfill$\blacksquare$ \\

\section{Proof of Corollaries}

\noindent 
\textbf{Proof of Corollary \ref{corollary_main_thm}}: First, for any $s > 0$, we have  
\[
\sum_{t=1}^{\infty}\frac{\{\log t\}^{\frac{1}{\gamma}}}{t^{(1+s)}} = C(s) < \infty,
\]  
where $C(s)$ is a constant depending only on $s$. Define  
\[
\delta_t = \frac{\delta}{2C(s)} \frac{\{\log t\}^{\frac{1}{\gamma}}}{t^{(1+s)}}, \text{ for } t\geq 1
\]  
Then, it follows that $\sum_{t=1}^{\infty} \delta_t = \frac{\delta}{2} < \delta$. Clearly, 
\begin{align*}
D^\prime(\delta) = 
   \left\{\Delta= \left(\delta C^{-1}(s) \{\log t\}^{\frac{1}{\gamma}}t^{-(1+s)}/2
   \right)_{t=1}^{\infty} \Big | s >0\right\}
   \subset D(\delta).
\end{align*}

By the assumption that $r(n) = n^{\kappa}$, we obtain $r(c_t n) = c_t^{\kappa} n^{\kappa}$. Applying Theorem \ref{Thm:Main}, we derive  
\begin{align}
\label{Eqq}
\mathbb{P} \left( \|\widehat{\bm{\theta}}_{\infty} - \bm{\theta}^\star\|_2 > \delta \right)    \leq & \inf_{\Delta \in D(\delta)} C_1 \sum_{t=1}^{\infty} \exp(-C_2 c_t^{\kappa} n^{\kappa} \delta_t^\gamma) \notag  \\
 \leq & C_1 \sum_{t=1}^{\infty} \exp\left( -C_2 \frac{c_t^{\kappa} n^{\kappa} \delta^{\gamma} \log t}{[2C(s)]^{\gamma} t^{\gamma(1+s)}} \right),
\end{align}
for some $s>0$. If $c_t$ is chosen such that
\begin{align*}
    \frac{C_2  c_t^{\kappa} n^{\kappa} \delta^{\gamma} }{[2C(s)]^{\gamma} t^{\gamma(1+s)}} \geq 1+l(n)
    \Leftrightarrow c_t \geq \frac{[2C(s)]^{\frac{\gamma}{\kappa}}t^{\frac{\gamma(1+s)}{\kappa}}(1+l(n))^{\frac{1}{\kappa}}}{nC_2^{\frac{1}{\kappa}}\delta^{\frac{\gamma}{\kappa}}},
\end{align*}
where $l(n)$ is an arbitrary function satisfying $\lim_{n \rightarrow \infty} l(n) = \infty$, then (\ref{Eqq}) can be further bounded as
\begin{align*}
    \mathbb{P} \left( \|\widehat{\bm{\theta}}_{\infty} - \bm{\theta}^\star\|_2 > \delta \right) \leq C_1
    \sum_{t=1}^{\infty} t^{-(1+l(n))} = \zeta(1+l(n)),
\end{align*}
where $\zeta(x) = \sum_{t=1}^{\infty}t^{-x}$ is the Riemann zeta function. Using the property that $\lim_{x\rightarrow \infty}\zeta(x)=0$, we have
\begin{align*}
   \lim_{n\rightarrow \infty}
   \mathbb{P} \left( \|\widehat{\bm{\theta}}_{\infty} - \bm{\theta}^\star\|_2 > \delta \right) = \lim_{n\rightarrow \infty}\zeta(1+l(n))=0.
\end{align*}
Note that $l(n)$ can be any arbitrary diverging function, therefore, as long as $c_t \gtrsim \frac{t^{\frac{\gamma(1+s)}{\kappa}}}{C_2^{\frac{1}{\kappa}}\delta^{\frac{\gamma}{\kappa}}}$, we have
\begin{align*}
    \lim_{n\rightarrow \infty}
   \mathbb{P} \left( \|\widehat{\bm{\theta}}_{\infty} - \bm{\theta}^\star\|_2 > \delta \right) =0.
\end{align*}
This completes the proof. \hfill$\blacksquare$  \\

\noindent
\textbf{Proof of Corollary \ref{Coro:Identity}.} We first prove the lower bound. By Theorem \ref{Thm:GaussianBetter}, we have $\bm{X} \sim N(\bm{0},\sum_{t=1}^{T-1}c_t^{-1}\bm{I}_p)$. For ease of notation, let $v = \sum_{t=1}^{T-1}c_t^{-1}$. 
\begin{align*}
        P(T) = \mathbb{E}_{\bm{X}}\left[\Phi\left(-\frac{\|\bm{X}\|_2}{2}\right)\right] \geq 
        \Phi\left(-\frac{\mathbb{E}_{\bm{X}}\left[\|\bm{X}\|_2\right]}{2}\right) \geq 
        \Phi\left(-\frac{\sqrt{\mathbb{E}_{\bm{X}}\left[\|\bm{X}\|_2^2\right]}}{2}\right) = \Phi\left(-\frac{\sqrt{vp}}{2}\right),
\end{align*}
where the last and the second last inequality follows from the Jensen's inequality.

Next, we turn to present the upper bound. Clearly, $\Vert \bm{X}\Vert_2^2 \sim v \cdot \chi^2(p)$, and the density function of $\Vert \bm{X}\Vert_2^2/v$ is given as
\begin{align*}
   p(x) = \frac{1}{2^{p/2} \Gamma(p/2)} x^{p/2 - 1} e^{-x/2}.
\end{align*}
Using the transformation $y = \sqrt{vx}$, we have
\begin{align*}
    p(y) = \frac{1}{2^{p/2-1}v^{\frac{p}{2}} \Gamma(p/2)} y^{p - 1} e^{-\frac{y^2}{2v}}.
\end{align*}
Therefore, $\Vert \bm{X}\Vert_2$ follows the scaled chi distribution with $p(y)$ being its density function. Then, $P(T)$ can be calculated as
\begin{align}
\label{P(T)_ori}
    P(T) = \int_{0}^{\infty}
    \frac{\int_{-\infty}^{-y/2}\frac{1}{\sqrt{2\pi}}e^{-\frac{x^2}{2}}dx}{2^{p/2-1}v^{\frac{p}{2}} \Gamma(p/2)} y^{p - 1} e^{-\frac{y^2}{2v}}dy
\end{align}
Note that $\Phi(-y/2)$ can be bounded as follows \citep{choudhury2007approximating}:
\begin{align}
\label{bound_gaussian}
\Phi(-y/2)=
\int_{-\infty}^{-y/2}\frac{1}{\sqrt{2\pi}}e^{-\frac{x^2}{2}}dx < \frac{\sqrt{2\pi}}{y}e^{-\frac{y^2}{8}},
\end{align}
Plugging \eqref{bound_gaussian} into \eqref{P(T)_ori} yields that
\begin{align*}
P(T) 
&= \int_{0}^{\infty}
    \frac{\int_{-\infty}^{-y/2} \frac{1}{\sqrt{2\pi}} e^{-\frac{x^2}{2}} dx}{2^{p/2 - 1} v^{p/2} \Gamma(p/2)} \, y^{p - 1} e^{-\frac{y^2}{2v}} dy < \int_{0}^{\infty}
    \frac{\frac{\sqrt{2\pi}}{y} e^{-\frac{y^2}{8}}}{2^{p/2 - 1} v^{p/2} \Gamma(p/2)} \, y^{p - 1} e^{-\frac{y^2}{2v}} dy 
     \\
&= \frac{\sqrt{2\pi}}{2^{p/2 - 1} v^{p/2} \Gamma(p/2)} \int_{0}^{\infty} y^{p - 2} e^{ - \frac{y^2}{8} - \frac{y^2}{2v} } dy = \frac{\sqrt{2\pi}}{2^{p/2 - 1} v^{p/2} \Gamma(p/2)} \int_{0}^{\infty} y^{p - 2} e^{ - y^2 \left( \frac{1}{8} + \frac{1}{2v} \right) } dy \\
&= \frac{\sqrt{2\pi}}{2^{p/2 - 1} v^{p/2} \Gamma(p/2)} \int_{0}^{\infty} y^{p - 2} e^{ - \frac{y^2 (v + 4)}{8v} } dy = \frac{\sqrt{2\pi}}{2^{p/2 - 1} v^{p/2} \Gamma(p/2)} \cdot \frac{1}{2} \left( \frac{v + 4}{8v} \right)^{ -\frac{p - 1}{2} } \Gamma\!\left( \frac{p - 1}{2} \right) \\
&= \frac{\sqrt{\pi} \, \Gamma\!\left( \frac{p - 1}{2} \right)}{2^{p/2 - 1/2} v^{p/2} \Gamma(p/2)} \left( \frac{8v}{v + 4} \right)^{\frac{p - 1}{2}}.
\end{align*}
This completes the proof. \hfill$\blacksquare$ \\

\end{document}